# Deep Pathomic Learning Defines Prognostic Subtypes and Molecular Drivers in Colorectal Cancer

Zisong Wang[1,2#], Xuanyu Wang[1,3#], Hang Chen[1#], Haizhou Wang[4], Yuxin Chen[1], Yihang Xu[1], Yunhe Yuan[1], Lihuan Luo[1], Xitong Ling[5*], Xiaoping Liu[1*]

[1]Department of Pathology, Zhongnan Hospital of Wuhan University, Wuhan, 430071, China

[2]School & Hospital of Stomatology, Wuhan University, Wuhan, 430079, China

[3]Department of Urology, Zhongnan Hospital of Wuhan University, Wuhan, 430071, China

[4]Department of Gastroenterology, Zhongnan Hospital of Wuhan University, Wuhan, 430071, China

[5]Shenzhen International Graduate School, Tsinghua University, Shenzhen, 518071, China

# Zisong Wang, Xuanyu Wang and Hang Chen contributed equally to this work.

**Correspondence:**

Xiaoping Liu, Email: liuxiaoping@whu.edu.cn; 169 Donghu Road, Wuhan, Hubei Province, China, 430071.

Xitong Ling, Email: lingxt23@mails.tsinghua.edu.cn; Shenzhen International Graduate School, Tsinghua University, Shenzhen, China, 518055.

**Abstract**

Precise prognostic stratification of colorectal cancer (CRC) remains a major clinical challenge due to its high heterogeneity. The conventional TNM staging system is inadequate for personalized medicine. We aimed to develop and validate a novel multiple instance learning model TDAM-CRC using histopathological whole-slide images for accurate prognostic prediction and to uncover its underlying molecular mechanisms. We trained the model on the TCGA discovery cohort (n=581), validated it in an independent external cohort (n=1031), and further we integrated multi-omics data to improve model interpretability and identify novel prognostic biomarkers. The results demonstrated that the TDAM-CRC achieved robust risk stratification in both cohorts. Its predictive performance significantly outperformed the conventional clinical staging system and multiple state-of-the-art models. The TDAM-CRC risk score was confirmed as an independent prognostic factor in multivariable analysis. Multi-omics analysis revealed that the high-risk subtype is closely associated with metabolic reprogramming and an immunosuppressive tumor microenvironment. Through interaction network analysis, we identified and validated Mitochondrial Ribosomal Protein L37 (MRPL37) as a key hub gene linking deep pathomic features to clinical prognosis. We found that high expression of MRPL37, driven by promoter hypomethylation, serves as an independent biomarker of favorable prognosis. Finally, we constructed a nomogram incorporating the TDAM-CRC risk score and clinical factors to provide a precise and interpretable clinical decision-making tool for CRC patients. Our AI-driven pathological model TDAM-CRC provides a robust tool for improved CRC risk stratification, reveals new molecular targets, and facilitates personalized clinical decision-making.

**Keywords**

Digital pathology; Deep Learning; Colorectal Cancer; Prognosis; Multi-omics; Biomarker

## Introduction

Colorectal cancer (CRC) is one of the most prevalent and deadly malignancies worldwide. According to World Health Organization, CRC ranks third in incidence and second in mortality among all cancers. CRC poses a huge threat to the global public health with an estimated 1.93 million new cases and 0.90 million deaths in 2022(1). Despite advancements in diagnosis and treatment, the prognosis for CRC patients remains highly unpredictable. Early diagnosis and accurate prognostic assessment are crucial for guiding clinical decisions and formulating personalized treatment plans (2). However, CRC is a highly heterogeneous disease. Its progression is influenced by multiple genetic and environmental factors, which further complicates prognostic prediction(3).

Currently, the primary clinical practice for CRC prognosis includes the TNM staging system, histological grading, and molecular biomarkers such as microsatellite instability (MSI) and mismatch repair deficiency (dMMR) (4, 5). However, these methods have limitations in predicting individual outcomes. For example, the TNM staging fails to adequately capture the microscopic morphological features and tumor microenvironment (TME) characteristics. Studies have showed a significant survival paradox between stage IIB/IIC and IIIA CRC patients(6, 7). Meanwhile, molecular biomarkers mainly focus on specific genes or protein expression and it is difficult to comprehensively assess the complex tumor biological behavior(8). Furthermore, molecular tests rely on specific equipment and technologies resulting in high time and resource cost and are hard for widespread adoption. Consequently, there is a critical unmet need in CRC for more precise, comprehensive, economical, and scalable prognostic prediction methods.

In recent years, digital pathology and artificial intelligence have sparked revolutionary breakthroughs in cancer research. Deep learning-based computational pathology can automatically and quantitatively extract complex morphological features from hematoxylin and eosin (H&E) stained slides and many of them are even unperceivable to human eyes(9-11). The multiple instance learning (MIL) is particularly suited for whole-slide images (WSIs) analysis. It can process gigapixel-scale images efficiently under weak supervision with only slide-level labels. WSIs contains thousands of or even millions of patches but MIL can automatically identify key diagnostic regions relevant to diseases(12, 13). Although MIL models have demonstrated potential in cancer prognostic prediction, there still remains space for optimizing their ability to effectively capture long-range dependencies between patches and aggregate global information. It is particularly important for tumors with extensive morphological heterogeneity such as CRC, where model performance and interpretability are key challenge.

Moreover, most existing computational pathology studies have primarily focused on high-performance of models while insufficiently explored the biological mechanisms underlying model decisions. An ideal prognostic model should not only predict outcomes precisely but also show the underlying rationale. Integrating deep pathomic features with high-throughput sequencing data can bridge the gap between tissue morphology and molecular mechanisms(14). Such multi-omics integration not only enhance the model interpretation but also offer opportunities to discover novel therapeutic targets and biomarkers.

This study aims to integrate histopathological morphology and transcriptomic data from CRC patients to develop a novel, interpretable MIL model named TDAM-CRC (Transformer-Dynamic Agent-Mamba for CRC) for precise prognostic risk stratification. This architecture combines (1) Transformer to efficiently process long-sequence data and capture long-range dependencies(15); (2) dynamic agent attention with low computational complexity and high expressive power at the same time to accelerate inference and improve performance(16); and (3) sequence reordering Mamba (SRMamba) with state-space modeling and awareness of instance order and distribution(17, 18). Based on TDAM-CRC, we further integrated the model-defined risk subtypes with transcriptomic data for comprehensive downstream bioinformatics analyses. This approach was designed to reveal molecular differences between risk subtypes at the levels of signaling pathways, TME, and key driver genes. By constructing an interaction network of pathological features, deep learning features, and genes, we successfully identified novel prognostic biomarkers. Finally, we developed and validated a nomogram that integrates deep learning features with clinical factors to provide a precise and interpretable tool for personalized prognostic assessment. In conclusion, this study not only proposes a new prognostic model with superior performance but also offers new perspectives for a deeper understanding of the CRC heterogeneity and its underlying molecular mechanisms.

## Results

### TDAM-CRC: A Deep Learning Model for Prognostic Stratification of CRC

This study aimed to integrate computational pathology with multi-omics analysis to develop and validate a novel MIL model for precise prognostic prediction for CRC patients **(Figure 1 & Supplementary Figure 1)**. We constructed the TDAM-CRC, which innovatively integrated a transformer module, a dynamic agent attention module, and an SRMamba module **(Figure 2A)**. To determine the optimal pathomic feature encoder, we compared three different pretrained foundation models as feature extractors. The results showed that regardless of the feature extractor used, TDAM-CRC demonstrated outstanding predictive performance in 5-fold cross-validation in the TCGA-COADREAD discovery cohort (n=581). Its mean concordance index (C-index) outperforming that of seven state-of-the-art (SOTA) models. In particular, when CONCH was employed as the feature extractor, TDAM-CRC achieved the best performance, with a mean C-index of 0.747 ± 0.044 **(Table 1)**. Therefore, CONCH was selected as the feature extractor for subsequent analyses.

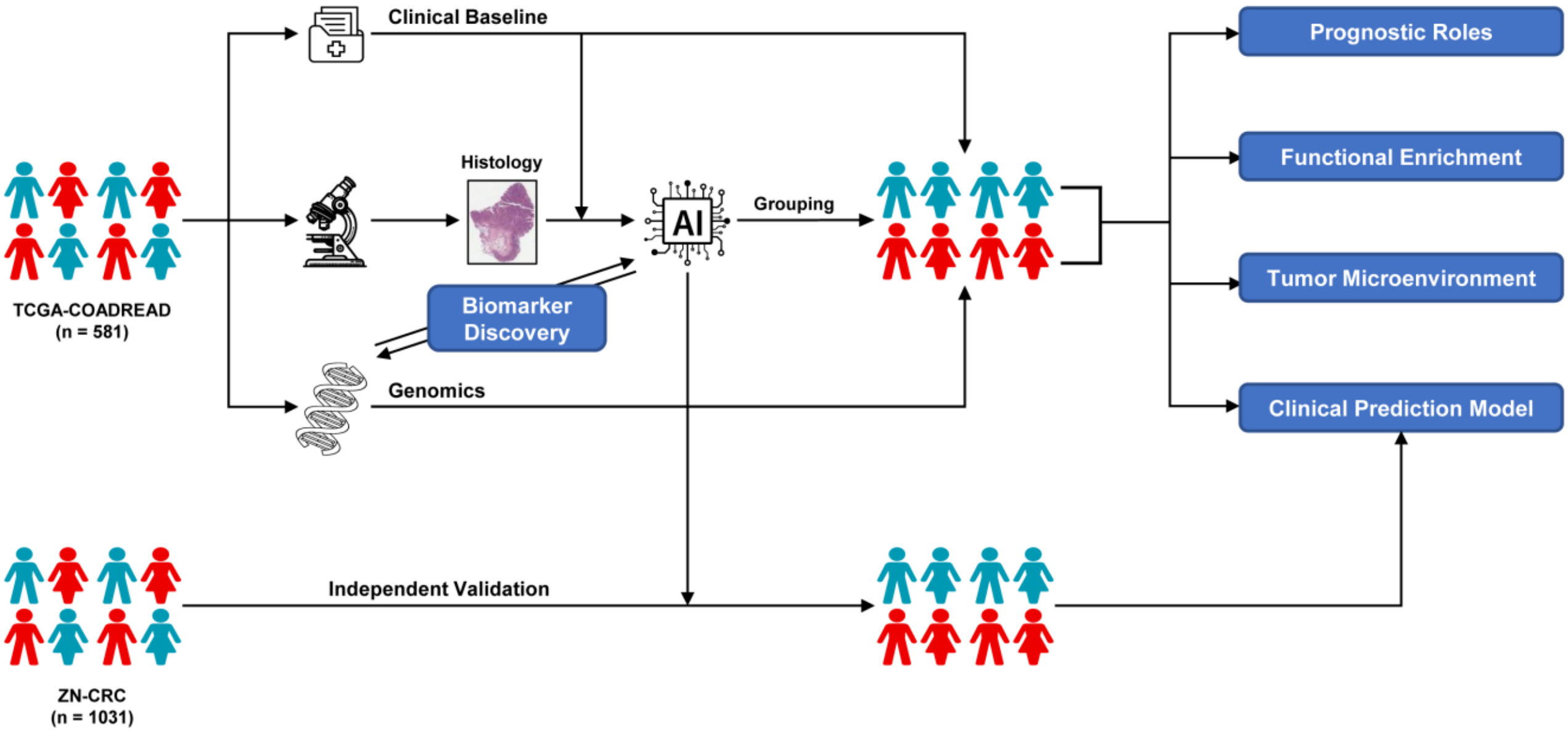

**Figure 1. Overall design and analysis workflow of this study.**

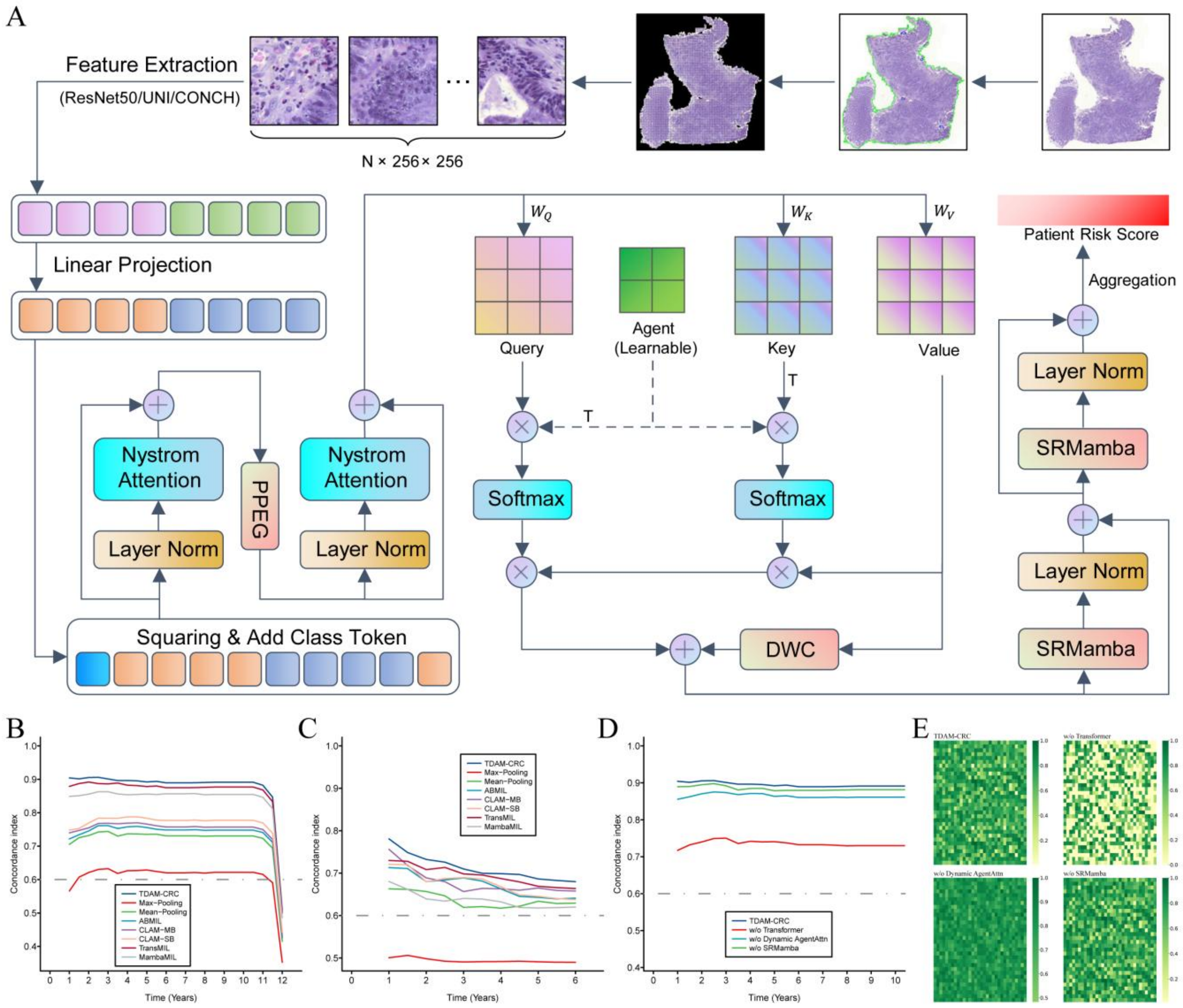

**Figure 2. Construction and performance validation of TDAM-CRC.** (A) Schematic of the TDAM-CRC architecture. (B-C) Time-dependent C-index curves for TDAM-CRC and seven SOTA model, evaluated using 5-fold cross-validation in the (B) TCGA-COADREAD and (C) ZN-CRC external validation cohorts. (D) Time-dependent C-index curves for different model variants in the ablation experiments. (E) Visualization of the ERF for the complete TDAM-CRC and its ablated variants.

| Model \ Extractor | ResNet-50 | UNI | CONCH |
|---|---|---|---|
| Max-Pooling | 0.609±0.090 | 0.651±0.048 | 0.596±0.087 |
| Mean-Pooling | 0.598±0.051 | 0.682±0.053 | 0.669±0.063 |
| ABMIL | 0.606±0.071 | 0.696±0.064 | 0.700±0.074 |
| CLAM-MB | 0.611±0.059 | 0.709±0.067 | 0.694±0.070 |
| CLAM-SB | 0.628±0.064 | 0.701±0.070 | 0.716±0.060 |
| TransMIL | 0.608±0.031 | 0.637±0.078 | 0.677±0.039 |
| MambaMIL | 0.680±0.025 | 0.699±0.053 | 0.729±0.066 |
| TDAM-CRC | **0.690±0.037** | **0.715±0.062** | **0.747±0.044** |

**Table 1. Comparison of 5-fold cross-validation performance of TDAM-CRC and SOTA models using different feature extractors in the TCGA-COADREAD cohort.**

To further assess the capability and generalizability of TDAM-CRC, we conducted a time-dependent C-index analysis and validated its robustness in an independent external cohort. In the TCGA-COADREAD cohort, the C-index curves of TDAM-CRC remained consistently higher than those of SOTA models over a follow-up time up to 10 years **(Figure 2B)**. The performance advantage was then confirmed in the independent, large-scale ZN-CRC external cohort (n=1031) **(Figure 2C)**. To elucidate the rationale of its performance, we conducted ablation experiments to verify the necessity of each module of TDAM-CRC. The results showed that removing any individual module, whether the Transformer (C-index: 0.687±0.063), the dynamic agent attention (C-index: 0.726±0.055), or the SRMamba (C-index: 0.739±0.048), led to reduction in the model's performance **(Figure 2D)**. To explore the contribution of each module from an information aggregation perspective, we performed effective receptive field (ERF) analysis. The results revealed that the complete TDAM-CRC model exhibited a broad and heterogeneous ERF. This indicated its ability to effectively capture and integrate key regions from fine-grained local features to global contextual information. In contrast, removing the Transformer module resulted in a significant decay of the ERF. Although the model without the dynamic agent attention showed a high-intensity overall ERF, it tended to be homogeneous and lacked the distinct focus characteristic of the whole model **(Figure 2E)**.

To investigate the pathological rationale of the model's predictions, we employed attention heatmaps to visualize its decision-making process and correlated them with histopathological features. The results showed that the pathological features in high-attention regions were closely related to risk stratification. High-attention regions in low- and medium-low-risk bins corresponded to lymphocyte aggregates, fibrotic stroma, or sparse cancer cells, whereas high- and medium-high-risk bins were associated with complex cellular morphology in tumor infiltration areas. These findings indicate that TDAM-CRC could automatedly learn and identify prognostically significant histomorphological characteristics **(Figure 3A-B)**.

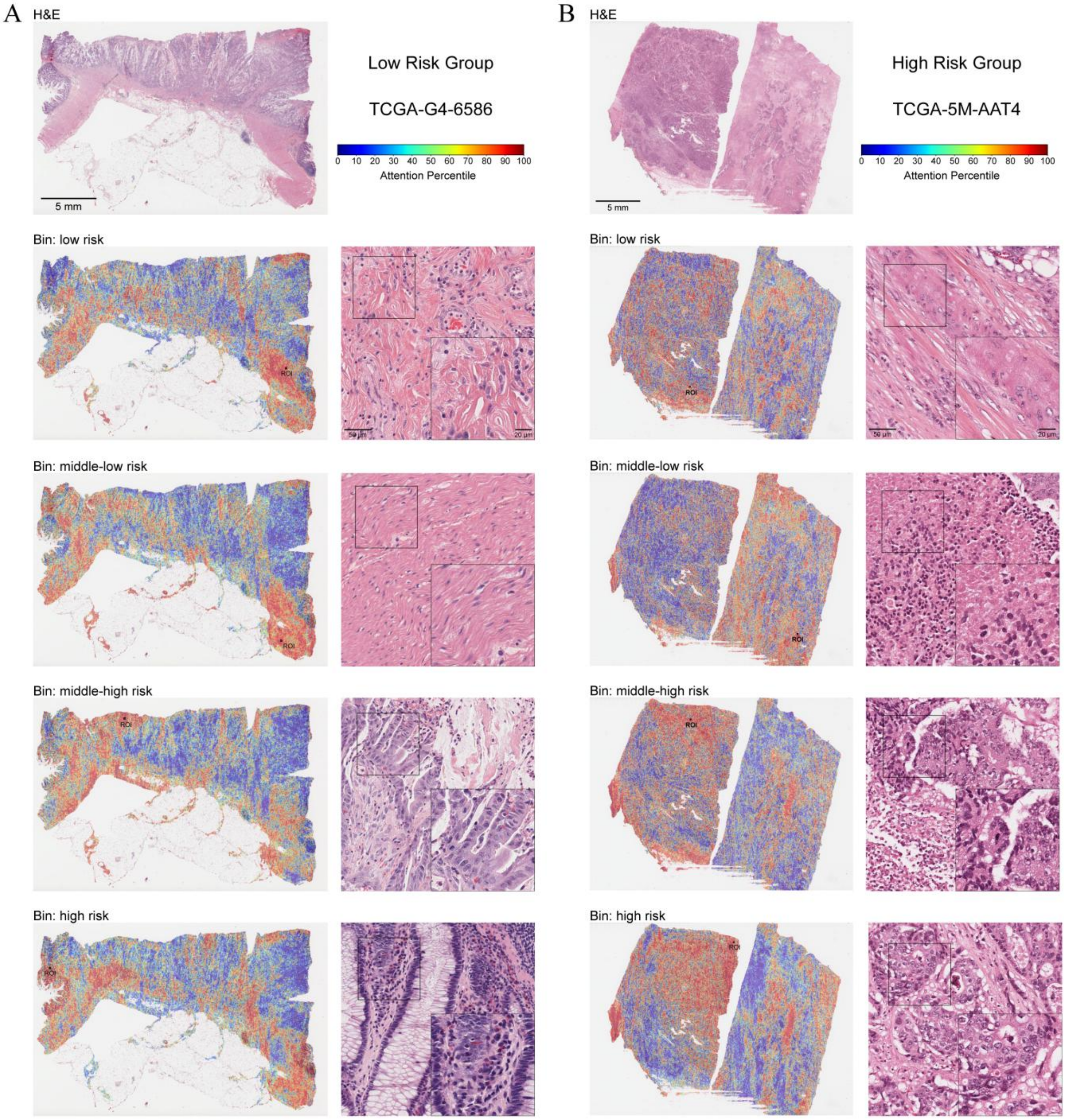

**Figure 3. Heatmap visualization of TDAM-CRC.** (A-B) H&E-stained WSIs, attention heatmaps for four survival bins, and regions of interest (ROIs) (20× and 40× magnification) of a representative sample from each of the (A) low-risk and (B) high-risk groups in the TCGA-COADREAD cohort.

**The TDAM-CRC Risk Score Independently and Accurately Predicts Clinical Outcomes**

To evaluate its clinical utility of the TDAM-CRC model, we used the median of the model's continuous risk score to stratify patients into high- and low-risk subgroups. In both the TCGA-COADREAD and ZN-CRC cohorts, high-risk status was significantly associated with worse overall survival and higher overall pathological stage **(Figure 4A-B, Supplementary Table 1-2)**. To evaluate the prognostic value of this risk stratification, we performed Kaplan-Meier (KM) survival analysis and restricted mean survival time (RMST) analysis. In both the TCGA-COADREAD and ZN-CRC cohorts, patients in the high-risk group had a significantly shorter overall survival (OS) than those in the low-risk group **(Figure 4C-D, Supplementary Table 3-4)**. This association held true within subgroups stratified by most clinical variables **(Supplementary Figure 2-3)**.

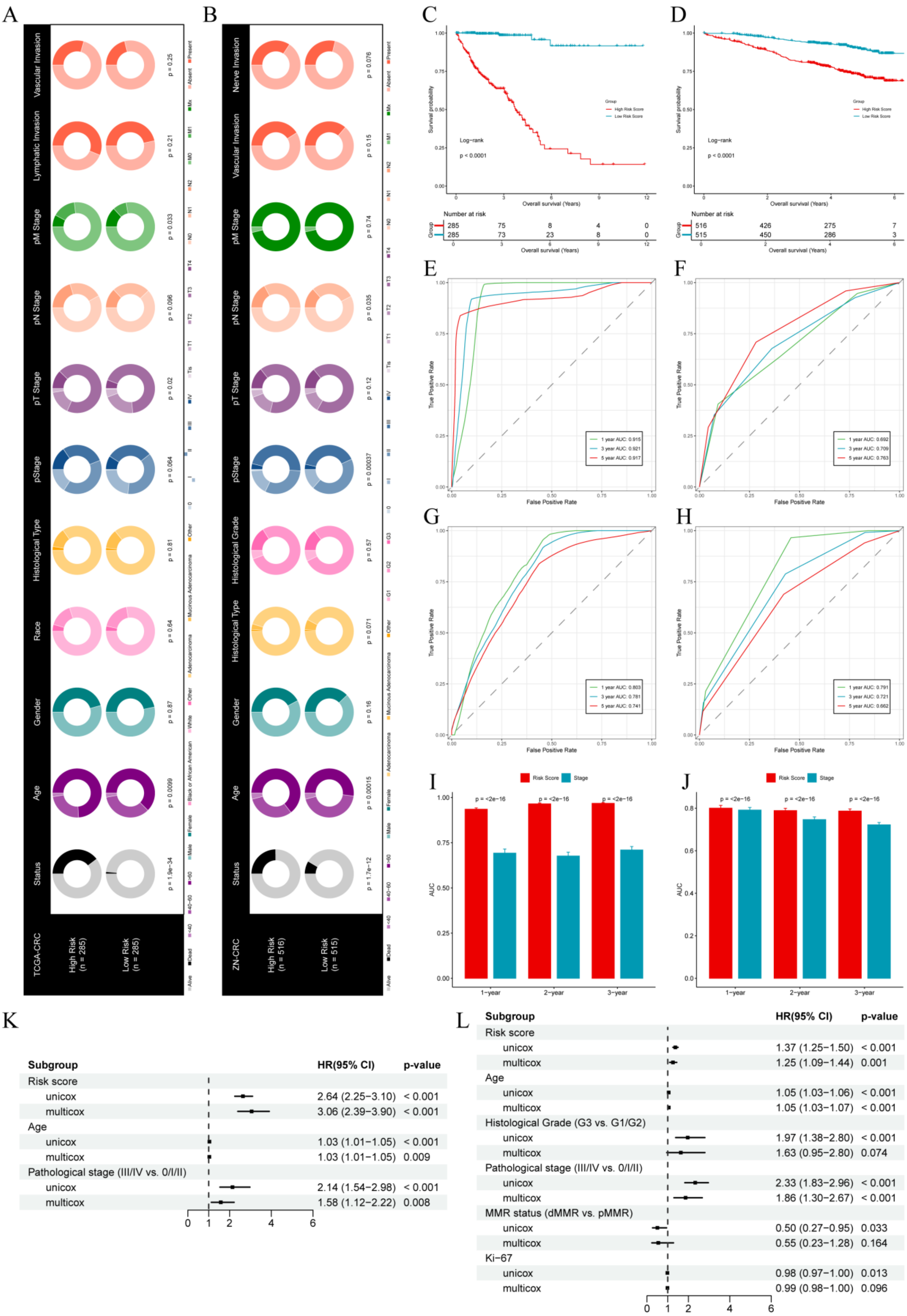
A
B
C
D
E
F
G
H
I
J
K
L
Vascular Invasion
Lymphatic Invasion
pM Stage
pN Stage
pT Stage
pStage
Histological Type
Race
Gender
Age
Status
TCGA-CRC
High Risk (n = 285)
Low Risk (n = 285)
Nerve Invasion
Histological Grade
ZN-CRC
High Risk (n = 516)
Low Risk (n = 515)
Log−rank
p < 0.0001
Survival probability
Overall survival (Years)
Number at risk
Group
High Risk Score
Low Risk Score
285 75 8 4 0
285 73 23 8 0
516 426 275 7
515 450 286 3
True Positive Rate
False Positive Rate
1 year AUC: 0.915
3 year AUC: 0.921
5 year AUC: 0.917
1 year AUC: 0.692
3 year AUC: 0.709
5 year AUC: 0.763
1 year AUC: 0.803
3 year AUC: 0.781
5 year AUC: 0.741
1 year AUC: 0.791
3 year AUC: 0.721
5 year AUC: 0.662
Risk Score
Stage
AUC
p = <2e−16
1−year
2−year
3−year
Subgroup
HR(95% CI)
p-value
Risk score
unicox 2.64 (2.25−3.10) < 0.001
multicox 3.06 (2.39−3.90) < 0.001
Age
unicox 1.03 (1.01−1.05) < 0.001
multicox 1.03 (1.01−1.05) 0.009
Pathological stage (III/IV vs. 0/I/II)
unicox 2.14 (1.54−2.98) < 0.001
multicox 1.58 (1.12−2.22) 0.008
Subgroup
HR(95% CI)
p-value
Risk score
unicox 1.37 (1.25−1.50) < 0.001
multicox 1.25 (1.09−1.44) 0.001
Age
unicox 1.05 (1.03−1.06) < 0.001
multicox 1.05 (1.03−1.07) < 0.001
Histological Grade (G3 vs. G1/G2)
unicox 1.97 (1.38−2.80) < 0.001
multicox 1.63 (0.95−2.80) 0.074
Pathological stage (III/IV vs. 0/I/II)
unicox 2.33 (1.83−2.96) < 0.001
multicox 1.86 (1.30−2.67) < 0.001
MMR status (dMMR vs. pMMR)
unicox 0.50 (0.27−0.95) 0.033
multicox 0.55 (0.23−1.28) 0.164
Ki−67
unicox 0.98 (0.97−1.00) 0.013
multicox 0.99 (0.98−1.00) 0.096

**Figure 4. Association between the TDAM-CRC risk score and clinical prognosis.** (A-B) Distribution of clinicopathological features in the high- and low-risk groups in the (A) TCGA-COADREAD and (B) ZN-CRC cohorts. (C-D) KM curves for OS in the high- and low-risk groups in the (C) TCGA-COADREAD and (D) ZN-CRC cohorts. (E-F) Time-dependent ROC curves for (E) the TDAM-CRC risk score and (F) clinicopathological staging in the TCGA-COADREAD cohort. (G-H) Time-dependent ROC curves for (G) the TDAM-CRC risk score and (H) clinicopathological staging in the ZN-CRC cohort. (I-J) Comparison of AUC values between the TDAM-CRC risk score and clinicopathological staging based on bootstrap resampling in the (I) TCGA-COADREAD and (J) ZN-CRC cohorts. (K-L) Univariable and multivariable CoxPHs evaluating the prognostic value of the TDAM-CRC risk score and other clinical variables in the (K) TCGA-COADREAD and (L) ZN-CRC cohorts.

To quantitatively assess the predictive efficacy of the TDAM-CRC, we compared it with the widely used clinicopathological staging. Time-dependent receiver operating characteristic (time-ROC) curves in the TCGA-COADREAD cohort showed that the areas under the curve (AUCs) for the TDAM-CRC risk score for predicting 1-, 3-, and 5-year survival were 0.915, 0.921, and 0.917 respectively, significantly exceeding those of pathological staging (AUCs: 0.692, 0.709, and 0.763 respectively) **(Figure 4E-F)**. Its performance was further validated in the ZN-CRC cohort, where the AUCs for the TDAM-CRC risk score in predicting 1-, 3-, and 5-year survival were 0.803, 0.781, and 0.741 respectively, consistently exceeding those of pathological staging (0.791, 0.721, and 0.662 respectively) **(Figure 4G-H)**. Bootstrap resampling validation confirmed that AUCs of the TDAM-CRC risk score were significantly better than those of pathological staging at the 1-, 2-, and 3-year time points in both cohorts **(Figure 4I-J)**.

To test the independent prognostic value of the TDAM-CRC, we employed univariable and multivariable Cox proportional hazards regression models (CoxPHs). After adjusting clinical covariates from the univariable CoxPHs (including age and pathological stage), multivariable CoxPHs revealed the risk score to be an independent predictor of poor prognosis in both the TCGA-COADREAD and ZN-CRC cohorts **(Figure 4K-L)**.

**Association of Risk Subtypes with Biological Pathways and the TME**

To delve into the molecular mechanisms underlying the TDAM-CRC-defined risk stratification, we extended our investigation beyond macroscopic pathological features to the microscopic molecular level. We first performed differential gene expression and functional enrichment analyses on the transcriptomic data from the TCGA-COADREAD cohort, after correcting batch effects in the merged TCGA-COAD and TCGA-READ datasets **(Supplementary Figure 4A-B)**. Compared with the low-risk group, 417 genes (e.g., HBZ, MAGEA12, RHAG) were significantly upregulated in the high-risk group, while 76 genes (e.g., SLC26A9, H2BC3, KRT24) were significantly downregulated **(Figure 5A & Supplementary Table 5)**. Gene Ontology (GO) and Kyoto Encyclopedia of Genes and Genomes (KEGG) analyses indicated that these differentially expressed genes (DEGs) were primarily enriched in biological processes such as lipid metabolism, oxidation-reduction processes, hormone secretion, and the cAMP signaling pathway **(Figure 5B-C & Supplementary Tables 6-7)**. Gene set enrichment analysis (GSEA) revealed that the high-risk subtype was significantly associated with pathways related to Krüppel-like factor 1 (KLF1), the IGF-IGFBP axis, gene sets related to liver and fetal tissues, regulation of blood coagulation, and heme metabolism (**Figure 5D-F**).

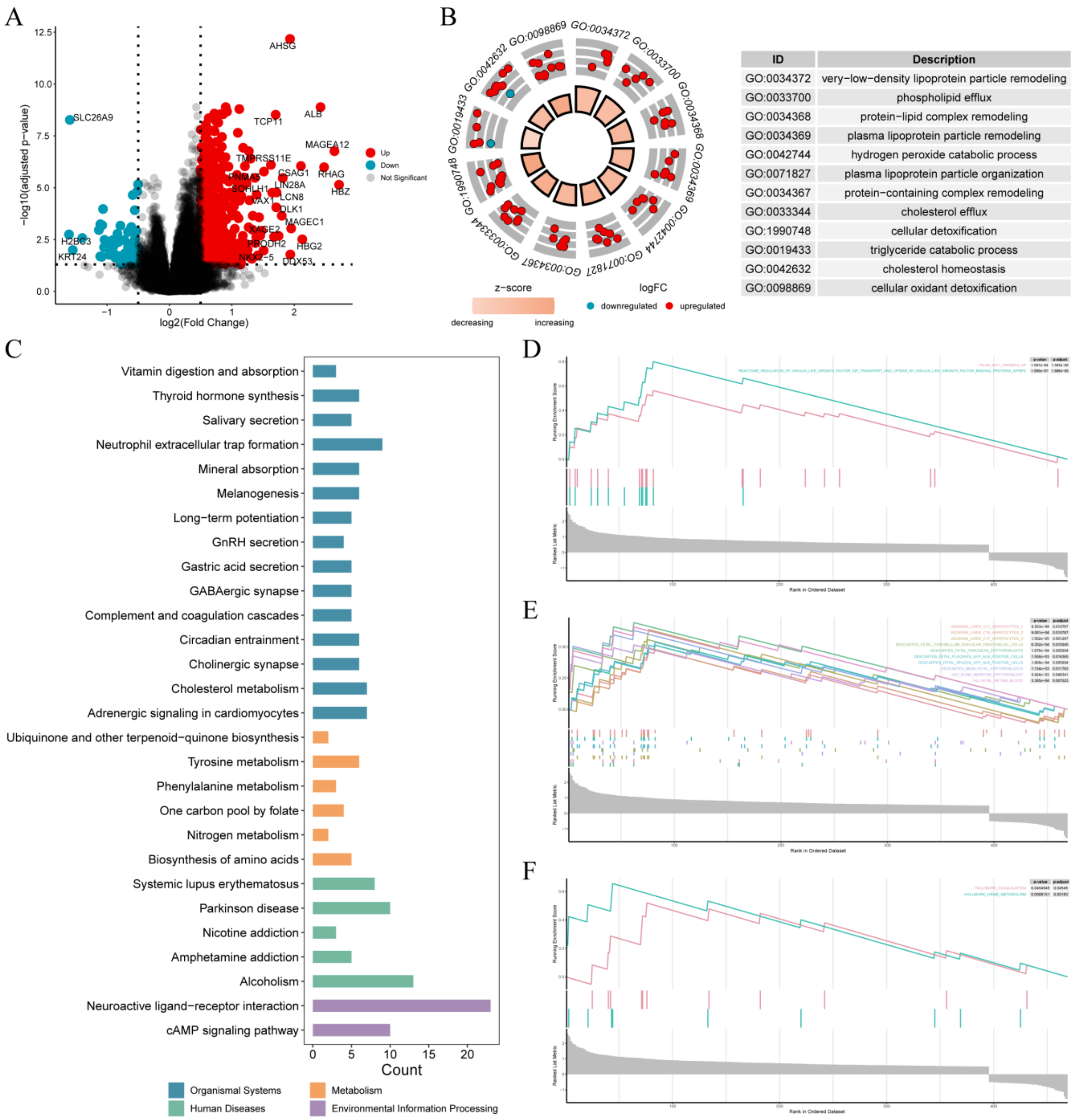

**Figure 5. Association between the TDAM-CRC risk score and biological functions.** (A) Volcano plot of DEGs between the high- and low-risk groups in the TCGA-COADREAD cohort. (B) Results of GO biological process enrichment analysis for DEGs. (C) Results of KEGG pathway enrichment analysis for DEGs. (D-F) GSEA results based on (D) the MSigDB C2 (curated gene sets) collection, (E) the MSigDB C8 (cell type signature gene sets) collection, and (F) the MSigDB H (hallmark) collection.

Given the critical role of the TME in CRC progression, we further investigated the intrinsic link between TDAM-CRC risk subtypes and the TME. Multi-omics analysis of immunomodulatory genes revealed that the high-risk group had a higher amplification frequency of genes such as CD274, PDCD1LG2, and TLR4, and a lower amplification frequency of genes like CCL5, CD70, and IFNG. Concurrently, genes including CD274, TNF, and IDO1 showed a higher deletion frequency, while genes such as CD80, IL12A, and VEGFB had a lower deletion frequency **(Figure 6A)**. To systematically compare the biological states of the different risk subtypes at a functional level, we performed single-sample GSEA (ssGSEA) on the TCGA-COADREAD cohort. The results showed that high-risk tumors exhibited significant activation of metabolic pathways (including nucleic acid, sugar, amino acid, energy, and lipid metabolism, as well as hypoxia response) and suppression of immune signaling pathways (such as BCR, TCR, and chemokine receptors) and the activity of various adaptive immune cells **(Figure 6B)**. To more precisely quantify immune cell components, we employed multiple algorithms for infiltration analysis. Quantitative analyses revealed that the high-risk group exhibited higher infiltration scores for $CD4^+$ effector memory T cells, Th1 cells, and $CD8^+$ T cells, suggesting signs of local activation of cellular immune responses. However, concomitant reductions in multiple dendritic cell subsets, along with suppressed B cell and plasma cell function, and attenuated activity other immune cell populations, collectively suggested an overall immunosuppressive state in the high-risk group **(Figure 6C)**. Finally, to link molecular features with potential clinical interventions, we performed a Connectivity Map (CMap) analysis based on the DEGs. This analysis predicted several small-molecule compounds (e.g., clofibrate, NU-1025, fasudil, MK-886, exisulind) that could potentially exert therapeutic effects by reversing the gene expression profile of the high-risk subtype, providing new leads for subsequent drug development **(Figure 6D)**.

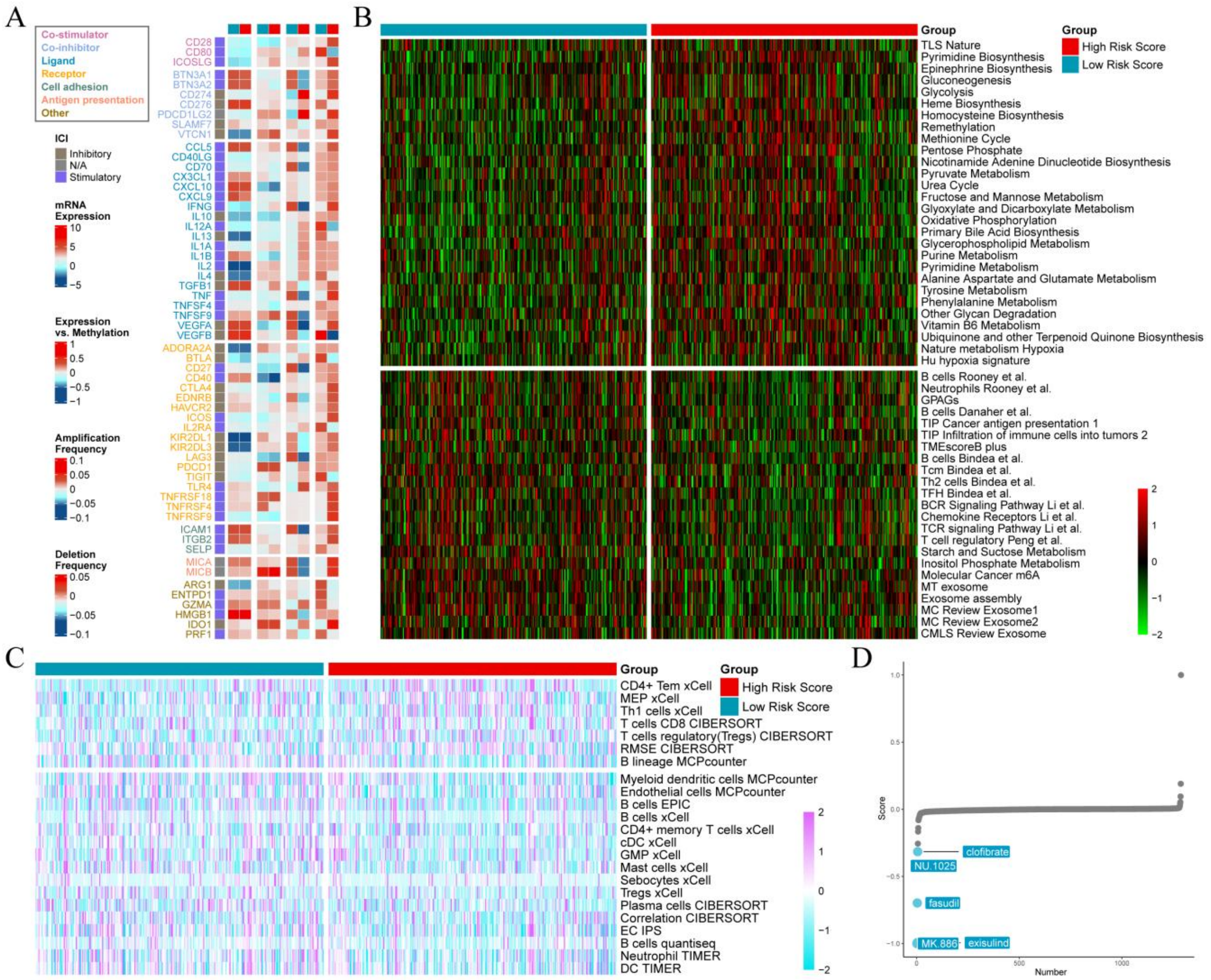

**Figure 6. Association of the TDAM-CRC risk score with the TME and drug prediction.** (A) Multi-omics analysis of key immunomodulatory genes in the TCGA-COADREAD cohort. From left to right: mRNA expression levels, correlation between gene expression and methylation, gene amplification frequency, and gene deletion frequency. (B) Heatmap of ssGSEA scores for biological pathways showing significant differences between the high- and low-risk groups in the TCGA-COADREAD cohort. (C) Heatmap of immune cell infiltration scores estimated by seven different algorithms, showing significant differences between the high- and low-risk groups in the TCGA-COADREAD cohort. (D) Candidate drugs predicted to target the high- and low-risk subtypes based on CMap analysis.

**Integrated Model-Pathology-Genomic Analysis Identifies and Validates MRPL37 as a Novel Prognostic Biomarker**

To bridge pathological morphology, deep learning features, and key molecular drivers, we employed a multi-level integrated pipeline to identify hub prognostic genes. First, using an elastic net regression model, we selected a set of 28 core pathomic features from the 512-dimensional deep learning features based on their robust association with the TDAM-CRC risk score **(Figure 7A-C)**. These features and the risk score formed a widely connected association network **(Figure 7D)**. Subsequently, we constructed an interaction network linking these core pathomic features with prognosis-related genes. By calculating eigenvector centrality, we identified MRPL37 as the top-ranking hub gene in this network **(Figure 7E & Supplementary Table 8)**, suggesting it may be a key node connecting pathological morphology with molecular prognosis.

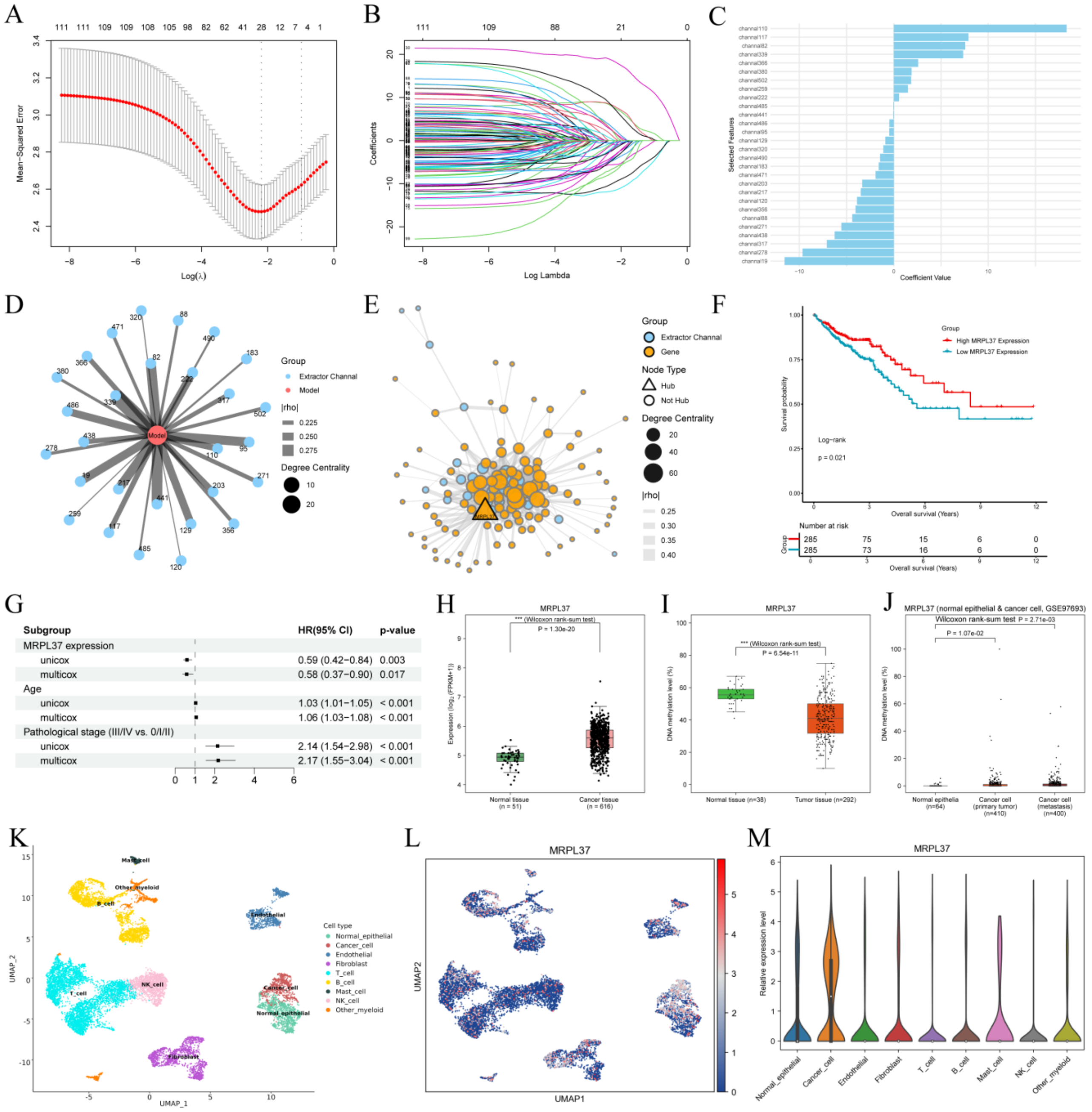

**Figure 7. Identification and validation of prognosis-related hub genes based on pathomic features.** (A) Cross-validation error curve for the elastic net regression model used to select core pathomic features. (B) Coefficient shrinkage path plot for the elastic net regression model. (C) Core pathomic features selected by the elastic net model and their regression coefficients. (D) Association network graph of core pathomic features and the TDAM-CRC risk score. (E) Association network graph of core pathomic features and prognosis-related genes, with the identified hub gene MRPL37. (F) Kaplan-Meier curves for OS in high- and low-MRPL37 expression groups in the TCGA-COADREAD cohort. (G) Univariable and multivariable CoxPHs evaluating the prognostic value of MRPL37 expression in the TCGA-COADREAD cohort. (H) Differential expression of MRPL37 in tumor versus adjacent normal tissues from CRC patients in the TCGA database. (I) Differential methylation of the MRPL37 promoter in tumor versus adjacent normal tissues from CRC patients in the TCGA database. (J) Differential methylation of the MRPL37 promoter in tumor versus normal tissues in the single-cell dataset GSE97693. (K) UMAP visualization and cell type annotation for all cells in the scRNA-seq dataset HRA000201. (L) Distribution of MRPL37 expression levels on the UMAP plot. (M) Violin plot of MRPL37 expression levels across different cell subpopulations.

To confirm the clinical value of the hub gene, we comprehensively validated the prognostic significance and biological function of MRPL37. In the TCGA-COADREAD cohort, KM curves showed that high MRPL37 expression was associated with significantly better OS **(Figure 7F)**. Univariable and multivariable CoxPHs further confirmed that MRPL37 is an independent favorable prognostic biomarker **(Figure 7G)**. To explore its regulatory mechanism, we analyzed multi-omics data and found that the high expression of MRPL37 in CRC tumor tissues was closely associated with a significant hypomethylation in its promoter region **(Figure 7H-I)**. Single-cell methylation data revealed that both primary and metastatic tumor cells had significantly higher levels of promoter methylation for MRPL37 compared with normal epithelial cells **(Figure 7J)**. Finally, to determine its cellular origin within the TME, we conducted a single-cell RNA-seq (scRNA-seq) analysis. The results showed that MRPL37 is expressed in various cells within the TME and is particularly enriched in tumor cells, suggesting it plays a key role in the intrinsic functions of tumor cells and their interactions with TME **(Figure 7K-M)**.

**A Nomogram Integrating the TDAM-CRC Risk Score Enhances Individualized Prognostic**

**Precision**

To translate our research findings into a clinically accessible tool, we constructed a nomogram by integrating independent prognostic variables in the multivariable CoxPHs. The nomogram incorporates three key prognostic factors (TDAM-CRC risk score, age, and pathological stage) to provide individualized predictions of 1- to 5-year survival probabilities **(Figure 8A)**.

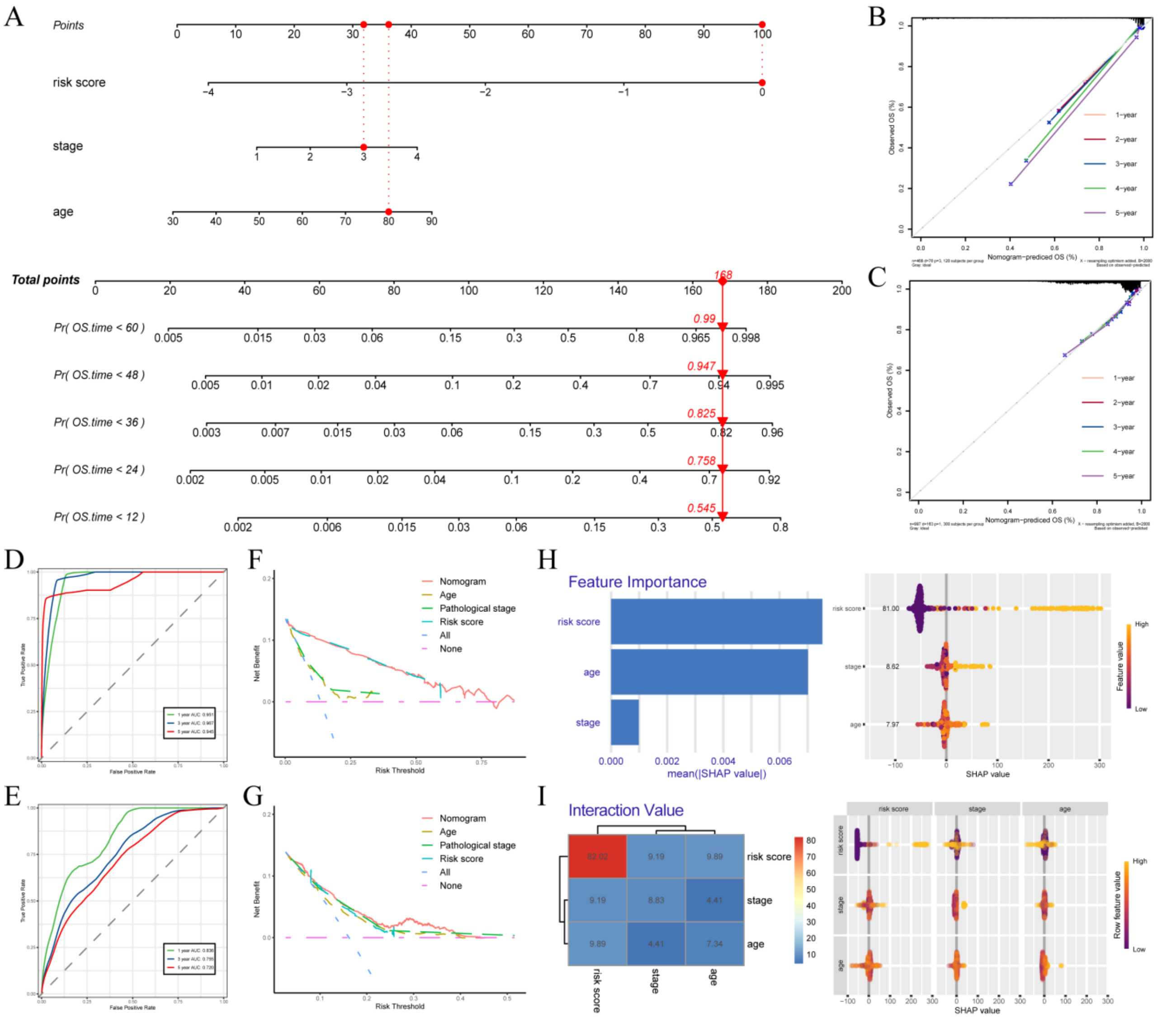

**Figure 8. Construction and evaluation of the clinical prediction nomogram.** (A) Prognostic nomogram integrating the TDAM-CRC risk score, age, and clinical stage. (B, C) Calibration curves for the nomogram in (B) the TCGA-COADREAD cohort and (C) the ZN-CRC cohort. (D, E) Time-dependent ROC curves for the nomogram predicting 1-, 3-, and 5-year survival rates in (D) the TCGA-COADREAD cohort and (E) the ZN-CRC cohort. (F, G) DCA for the nomogram in (F) the TCGA-COADREAD cohort and (G) the ZN-CRC cohort. (H) SHAP analysis revealing the global importance

of each variable in the nomogram model. (I) SHAP analysis visualizing the contribution of each variable to individual predictions and their interactions.

In both the TCGA-COADREAD training cohort and the ZN-CRC validation cohort, the calibration curves showed high consistency between the predicted and actual observed survival probabilities, confirming the nomogram's accuracy **(Figure 8B-C)**. In terms of discrimination, the nomogram demonstrated high performance, with a C-index of 0.929 in the TCGA-COADREAD cohort and AUC values for predicting 1-, 3-, and 5-year survival of 0.951, 0.967, and 0.945 respectively. In the ZN-CRC cohort, the C-index was 0.736, and the AUC values for 1-, 3-, and 5-year survival also reached 0.836, 0.755, and 0.720 respectively **(Figure 8D-E)**. Decision curve analysis (DCA) indicated that the integrated nomogram provided the highest net benefit across a wide range of clinical decision thresholds compared with any single-factor model in both cohorts, highlighting its clinical utility **(Figure 8F-G)**. Finally, to enhance its interpretability, we applied SHapley Additive exPlanations (SHAP) analysis to quantify the contribution of each variable. We confirmed that the TDAM-CRC risk score was the most influential predictor and revealed how the variables synergistically drive prognostic judgments for individual patients **(Figure 8H-I)**.

**Discussion**

In a highly heterogeneous disease such as CRC, precise prognostic stratification is important for individualized treatment(2, 19). Prognostic assessment in CRC traditionally relies on the UICC/AJCC TNM staging system. However, existing studies have shown that TNM staging is often insufficient for distinguishing clinical outcomes and can even lead to contradictory results(6, 7, 20), highlighting the clinical need for more precise risk stratification tools. This study successfully developed and validated a novel CRC risk stratification framework TDAM-CRC. Trained on H&E-stained WSIs, the model achieves precise prognostic stratification for CRC patients. The TDAM-CRC risk score demonstrated superior and robust performance in both internal and external cohorts totaling over 1,500 patients. Its predictive efficacy significantly exceeded that of the clinically used TNM staging system and SOTA models, and was confirmed as an independent prognostic factor. We further uncovered the biological mechanisms behind the risk stratification through multi-omics integration, revealing associations with specific signaling pathways and the TME, and ultimately identifying and validating MRPL37 as a novel biomarker linking deep pathomic features to patient prognosis. This study not only provides an effective new tool for clinical risk management for CRC patient but also proposes a new avenue for understanding the molecular drivers of tumor morphological heterogeneity.

One of the core contributions of this study is the architecture of the TDAM-CRC model. Traditional MIL-based computational pathology models have limited ability to capture long-range dependencies among WSI patches and efficiently aggregate global contextual information(21). To address these challenges, TDAM-CRC integrates a Transformer module, a dynamic agent attention module, and an SRMamba module. The Transformer excels at capturing long-range dependencies and has been shown to effectively integrate morphological and spatial WSI characteristics (15, 22). The dynamic agent attention can dynamically generate agents during training to weight different pathological regions, enhancing the model's ability to focus on key areas(16). The SRMamba with its linear computational complexity and powerful long-context modeling capabilities can provide an efficient solution for processing the massive patches of WSIs(17, 18). The ablation experiments and ERF analysis demonstrated that the synergy of these components gives TDAM-CRC a broad yet focused field of view to effectively aggregate key information from local, fine-grained cellular morphology to global tissue architecture. This is fully reflected in the model's performance: in both the TCGA discovery cohort and the larger external validation cohort, TDAM-CRC's predictive performance was consistently and significantly superior to that of SOTA models.

In both the TCGA and external validation cohorts, our model achieved fine prognostic stratification of CRC patient. Its predictive accuracy significantly outperformed clinical staging. This finding is consistent with existing research, which shows that pathological image features contain richer biological information than TNM staging alone(23), and deep learning can uncover prognostic signals difficult for traditional pathology to capture(24). Clinically, our findings suggest that TDAM-CRC can provide a more personalized prognostic assessment for CRC patients, helping to identify high-risk individuals and guide therapy decisions.

An ideal prognostic model should possess not only high predictive performance but also good interpretability. Through attention heatmap visualization, this study successfully linked the model's abstract decision-making process to specific histopathological features. We found that the high-risk regions identified by the model morphologically corresponded to indicators of malignant progression, such as the tumor invasive front and clusters of poorly differentiated cells, while low-risk regions corresponded to features associated with a favorable prognosis, such as higher differentiation, lymphocyte aggregates, or fibrotic stroma(25, 26). This not only provides an intuitive pathological basis for the model's predictions, enhancing its clinical credibility, but also demonstrates that a deep learning model can autonomously learn and quantify prognostic morphological patterns that are extremely complex or imperceptible to human pathologists. This data-driven pathological insight is one of the core advantages of computational pathology over traditional pathological assessment.

We further integrated the risk stratification with transcriptomic data to comprehensively uncover the biological mechanisms driving prognostic differences. Functional enrichment revealed that the high-risk subtype is closely associated with lipid metabolism reprogramming. This aligns with evidence that abnormal lipid metabolism is a hallmark of cancer, providing energy and biosynthetic precursors for rapid tumor cell proliferation, invasion, and metastasis(27). Additionally, we observed an enrichment of the cAMP signaling pathway in the high-risk group. The cAMP pathway plays a complex dual role in cell proliferation, differentiation, and apoptosis(28, 29), and its abnormal activation in CRC can promote tumor progression(30, 31). GSEA linked the high-risk status to KLF1, blood coagulation, heme metabolism, the IGF-IGFBP axis, and gene sets related to liver and fetal tissues. Each of these findings has support in literatures. For instance, Krüppel-like factors (KLFs) are transcription factors often dysregulated in cancer, modulating proliferation, apoptosis, metastasis, and the TME(32). KLF1, primarily known for its hematopoietic role(33), is reportedly upregulated in CRC(34) and may be involved in proliferation and metastasis(35). Similarly, cancer-associated coagulation abnormalities are common complications in CRC patients, closely linked to tumor progression and metastasis(36, 37). Heme oxygenase-1 (HMOX1), a key enzyme in heme metabolism, has also been reported to affect CRC growth, although its specific roles remains controversial(38, 39). The insulin-like growth factor (IGF) system is implicated in tumor development, although epidemiological findings are inconsistent(40-42). Finally, the liver- and fetal-tissue-related pathways are biologically relevant, as the liver is the most common site of CRC metastasis with different TME(43, 44), and CRC cells may enhance malignancy by reactivating embryonic gene programs(45, 46). These findings collectively suggest that high-risk CRC reshapes key metabolic and signaling pathways to gain survival and proliferative advantages, providing valuable directions for future research.

The TME is a key determinant of CRC progression and treatment response. Our analysis indicates that high-risk tumors tend to have an overall immunosuppressive state. ssGSEA showed that the signaling pathway activity of various adaptive immune cells was suppressed in the high-risk group. Results from immune cell algorithm analysis, which showed a reduction in dendritic cell subsets and suppressed B cell and plasma cell function, collectively point to weak links in the adaptive immune response chain(47, 48). Although the infiltration scores of some effector immune cells (such as $CD8^+$ T cells) were increased, this might reflect an ineffective or exhausted immune response state rather than effective anti-tumor immunity(49). This dysfunctional immune state could be a crucial reason for tumor immune evasion and continued progression(50). Concurrently, we found that key immune checkpoint molecules (such as CD274/PD-L1 and PDCD1LG2/PD-L2) had a higher amplification frequency in the high-risk group, suggesting that tumors may achieve immune escape by enhancing immunosuppressive signals. This also implies that the high-risk patient population may be more sensitive to combination therapies involving immune checkpoint inhibitors targeting the PD-1/PD-L1 axis(51, 52). Among the potential anti-CRC small-molecule drugs screened, MK-886 pretreatment has been shown to enhance sensitivity to high-hemin photodynamic therapy, inducing cell cycle arrest and promoting apoptosis in CRC cells(53); exisulind inhibits the proliferation of CRC cells and induces apoptosis(54), indicating the therapeutic potential of these small molecules.

To forge a link among pathomic features, the deep learning model, and molecular drivers, we designed an innovative multi-level integrated analysis pipeline and identified MRPL37 as the core hub gene in the pathology-gene interaction network. Mitochondrial ribosomal proteins (MRPs) are essential components for maintaining mitochondrial functional integrity, and evidence suggests their abnormal expression may be related to tumor progression(55). As a member of the MRP family, MRPL37 has been implicated in apoptosis and is upregulated in various tumor cell lines(56). However, its specific role and clinical significance in CRC were unclear. Our analysis confirms that MRPL37 may act as a tumor suppressor in CRC, with its high expression serving as an independent biomarker of favorable prognosis. Mechanistically, we found that high MRPL37 expression in tumor tissues is significantly correlated with hypomethylation of its promoter region, revealing a potential epigenetic regulatory mechanism. Single-cell level analysis further confirmed that MRPL37 is predominantly enriched in tumor cells, suggesting it may influence disease progression by regulating the mitochondrial function of the tumor cells themselves. The discovery of MRPL37 not only provides a novel, multi-omics validated prognostic biomarker for CRC, but also demonstrates how a deep learning model can be used as a probe to mine key molecular targets with critical biological functions from complex WSI data.

The ultimate goal of this research is to translate findings into clinically practical tools. We constructed and validated a nomogram that integrates the TDAM-CRC risk score, age, and clinical stage. The nomogram demonstrated high predictive accuracy and calibration in both internal and external cohorts. DCA further confirmed that the nomogram provides a higher net benefit across a wide range of clinical decision thresholds compared to any single factor. Combined with the predictive explanations provided by SHAP analysis, the nomogram serves as a robust and interpretable auxiliary tool to assess the prognosis of CRC patients and formulate personalized treatment strategies.

This study also has some limitations. First, although the model was validated in a large external cohort, this was a retrospective study and its conclusions need further confirmation from prospective clinical trials. Second, the biological pathways and the function of MRPL37 revealed in this study are primarily based on bioinformatics analyses. Their molecular mechanisms need to be elucidated through subsequent biological experiments. Finally, the TDAM-CRC is currently trained only on H&E-stained WSIs and its applicability for other types of stains or different pathological preparation protocols remains to be evaluated.

In conclusion, this study developed a novel deep learning model TDAM-CRC with both fine performance and interpretability for prognostic stratification for CRC patients. By comprehensively integrating multi-omics data, we not only revealed differences in metabolic reprogramming, signaling pathways, and the TME between risk subtypes but also identified and validated MRPL37 as a novel prognostic biomarker. The nomogram provides a useful tool for personized and precise prognostic assessment for CRC patients. Future work will focus on the prospective clinical validation of TDAM-CRC and deeper exploration of mechanisms and biomarkers of the high-risk subtype, with the aim of improving clinical outcomes for CRC patients.

## Methods

### Patient Cohorts and Data Sources

This study analyzed data from two independent cohorts of CRC patients. The TCGA-COADREAD cohort, serving as discovery cohort, integrated the colon adenocarcinoma (COAD) and rectum adenocarcinoma (READ) datasets from The Cancer Genome Atlas (TCGA)(57)'(58). After excluding samples with missing survival information, zero OS time, or poor-quality WSI, this cohort ultimately included 581 patients with H&E-stained WSIs for model training and evaluation. The clinical annotation and RNA sequencing (RNA-seq) data for this cohort were downloaded from the TCGA database via the R package TCGAbiolinks (v2.28.3) on January 31, 2025(59). The DNA methylation profiles (beta-values) and gene-level copy numbers (estimated using the GISTIC2 method) were obtained from the UCSC Xena Browser(60). To eliminate batch effects, we merged and corrected the RNA-seq data (in Count and TPM formats) from TCGA-COAD and TCGA-READ using the R package sva (v3.48.0) (61). A total of 570 samples in the TCGA-COADREAD cohort had RNA-seq data. Except for the differential gene expression analysis, which used Count data, all other transcriptomic analyses in this study were based on $\log_2$(TPM+0.1) transformed data.

To independently assess the model's robustness and generalizability, we collected data from 1031 CRC patients from Zhongnan Hospital of Wuhan University as an external validation cohort (ZN-CRC cohort). This cohort included WSIs and corresponding clinical follow-up information. The WSIs were scanned at 40× magnification using a Hamamatsu NanoZoomer scanner. This study was conducted in adherence to the Declaration of Helsinki, and all data collection and usage were approved by the Ethics Review Committee of Zhongnan Hospital of Wuhan University (No. 2025010K). The inclusion criteria were: (1) pathologically confirmed primary CRC; and (2) availability of high-quality H&E-stained surgical resection specimens. The exclusion criteria were: (1) missing key clinical or survival data; and (2) metastatic CRC.

**WSI Preprocessing and Feature Extraction**

We employed the Clustering-constrained Attention Multiple Instance Learning (CLAM) framework for standardized WSI preprocessing and feature extraction(62-64). First, automated tissue region segmentation was performed on each WSI to identify and exclude background, blurry areas, and artifacts. This segmentation process utilized the preset tcga.csv and bwh_resection.csv files from the CLAM framework for the TCGA-COADREAD and ZN-CRC cohorts respectively. Subsequently, within the segmented valid tissue regions, the images were non-overlappingly tiled into 256×256 pixel patches at an equivalent magnification of 20×.

For high-dimensional feature encoding of the patches, we utilized pretrained ResNet50(65), UNI(66), and CONCH(67) large models as feature extractors. Each patch was encoded into a 1024-dimensional (ResNet50, UNI) or 512-dimensional (CONCH) feature vector. Ultimately, each WSI was represented as a set of all its patch feature vectors, termed a bag, and served as the input for MIL models.

**TDAM-CRC Model Construction, Training, and Performance Evaluation**

We designed and developed TDAM-CRC, a novel MIL model aimed at accurately predicting patient prognosis by analyzing WSI histological features. TDAM-CRC integrates the Transformer(15, 68), the dynamic agent attention mechanism(16), and the SRMamba module(17, 18), to efficiently process vast amounts of patch information and capture long-range dependencies among key prognostic regions. The model's data flow is as follows: first, the sequence of WSI patch features is fed into a linear projection layer for dimensionality transformation. Second, to achieve computationally efficient information aggregation, we introduced the Nyström-based Transformer attention mechanism, and a set of learnable agent query vectors to adaptively capture features from the most informative regions. Subsequently, the attention-aggregated features are input into a multi-layer stack of SRMamba modules, which leverage state-space model principles to effectively model the complex spatial dependencies within the patch sequence, thereby generating a global, patient-level pathological representation. Finally, an aggregation layer fuses the information into a single patient-level vector, from which a continuous risk score is calculated to assess patient prognosis. The mathematical principles of the TDAM-CRC model are detailed below.

(1) Transformer Module

a. Input Preprocessing and Linear Mapping: Let the input features be $X \in \mathbb{R}^{B \times N \times D}$, where $B$ is the batch size, $N$ is the number of tokens in the input sequence, and $D$ is the feature dimension. First, the input features are projected to a uniform dimension $D_{model}$ through a fully connected (FC) layer, followed by a GELU activation function:

$$X_{proj} = \mathrm{GELU}(\mathrm{FC}(X))$$

b. Token Padding and Class Token Addition: To facilitate subsequent 2D positional encoding, the number of tokens $N$ is padded to $N'$, which is a perfect square. A learnable class token $X_{\mathrm{cls}}$ is prepended to the sequence:

$$X_{\mathrm{padded}} = \left[X_{\mathrm{cls}}; X_{proj}^{'}\right]$$

c. First Transformer Layer Processing: The processed sequence is input into a Transformer layer based on Nyström attention for initial feature extraction:

$$X_{\mathrm{T1}} = \mathrm{Attention}\left(\mathrm{LayerNorm}(X_{\mathrm{padded}})\right) + X_{\mathrm{padded}}$$

d. Positional Encoding: To better capture local spatial information, multi-scale convolutions are used for positional encoding enhancement:

$$X_{\text{pos}} = \text{Conv}_7(X_{\text{T1}}) + \text{Conv}_5(X_{\text{T1}}) + \text{Conv}_3(X_{\text{T1}}) + X_{\text{T1}}$$

where $\text{Conv}_\text{k}$ denotes a depthwise convolution operation with a stride of 1 and a kernel size of $k \times k$.

e. Second Transformer Layer Processing: The features, now fused with positional information, are fed into another Transformer layer for deep feature interaction:

$$X_{\text{out_T}} = \text{Attention}\left(\text{LayerNorm}(X_{\text{pos}})\right) + X_{\text{pos}}$$

(2) Dynamic Agent Attention Module

a. Agent Token Initialization: A set of trainable agent tokens $P_{\text{agent}} \in \mathbb{R}^{B \times N_p \times D_{model}}$ is introduced and reshaped to $\mathbb{R}^{B \times N_h \times N_p \times (D_{model}/N_h)}$ to match the multi-head attention mechanism, where $N_p$ and $N_h$ are the number of agent tokens and attention heads, respectively.

b. Linear Transformation to Generate query (Q), key (K), and value (V): The number of tokens from the upstream Transformer module's output, $X_{\text{out_T}}$, is adjusted to be a perfect square. Learnable parameters are used to generate the Q, K, and V:

$$Q_X = W_q \cdot X_{\text{out_T}}, \quad [K_X, V_X] = W_{kv} \cdot X_{\text{out_T}}$$

where $W_q \in \mathbb{R}^{D_{model} \times D_{model}}$, $W_{kv} \in \mathbb{R}^{D_{model} \times 2D_{model}}$, and the results are split for multi-head computation.

c. Agent-to-Patch Attention: The agent tokens serve as Q, and the patch feature sequence serves as K and V. Attention is computed to update the agent tokens, enabling them to aggregate patch information:

$$\text{Attention}_{\text{A2P}} = \text{Softmax}\left(\frac{P_{\text{agent}} \cdot K_X^\top}{\sqrt{d_k}} + B_{A2P}\right)$$

$$P_{\text{agent}}^{'} = \text{Attention}_{\text{A2P}} \cdot V_X$$

where $d_k$ is the feature dimension per head and $B_{A2P}$ is a bias term.

d. Patch-to-Agent Attention: The patch feature sequence serves as Q, and the updated agent tokens serve as K and V. Another attention interaction is performed to update the patch representations:

$$Attention_{\text{P2A}} = \text{Softmax}\left(\frac{Q_X \cdot P_{\text{agent}}^\top}{\sqrt{d_k}} + B_{P2A}\right)$$

$$X_{out_T}^{'} = Attention_{\text{P2A}} \cdot P_{\text{agent}}^{'}$$

e. Local Information Enhancement and Output Linear Mapping: A depthwise separable convolution

(DWConv) is applied to the updated features and combined with the attention output from the main path, followed by a linear projection:

$$X_{\text{out_DA}} = \text{Linear}\left(\text{DWConv}(V_X) + X'_{out_T}\right)$$

(3) SRMamba Module

a. Normalization: The input features $X_{in}^{(l)}$ for the *l*-th layer are normalized:

$$X_{norm}^{(l)} = \text{LayerNorm}\left(X_{in}^{(l)}\right)$$

b. SRMamba Module Transformation: The normalized features are input into the SRMamba module for state-space modeling, yielding an incremental feature $\Delta X$：

$$\Delta X = \text{SRMamba}\left(X_{norm}^{(l)}, \text{rate}\right)$$

where rate is a hyperparameter for controlling the scale.

c. Residual Connection: The feature representation is updated via a residual connection:

$$X_{out}^{(l)} = X_{in}^{(l)} + \Delta X$$

(4) Feature Aggregation and Risk Prediction

a. Normalization: The output feature matrix *Z* from the upstream modules is layer-normalized:

$$Z_{norm} = \text{LayerNorm}(Z)$$

b. Attention Weight Calculation: A small attention network computes the weight for each patch, and a weighted sum is performed to obtain a global feature vector $Z_{global}$:

$$A = \text{Softmax}\,(\text{Linear}(\tanh\,(\text{Linear}(Z_{norm}))))$$

$$Z_{global} = \sum_{k=1}^{N'} A_{\text{k}} \cdot Z_{norm,k}$$

c. Classification Prediction: The aggregated global features are input into a classifier to obtain logits corresponding to each prognostic risk bin:

$$\text{Logits} = \text{Classifier}(Z_{global})$$

d. Risk Score Calculation: The logits are transformed into survival risks using a Sigmoid function, which are then used to calculate cumulative survival probabilities. Finally, the patient's risk score is obtained by the negative summation of these survival probabilities.

The model was trained on the TCGA-COADREAD cohort using a 5-fold cross-validation strategy. The training employed the Adam optimizer with an initial learning rate of $2 \times 10^{-4}$ and a dropout rate of 0.25 to prevent overfitting. To process survival data, we discretized the continuous survival times into four bins (low risk, medium-low risk, medium-high risk, and high risk) and used a cross-entropy loss function designed for survival analysis(69). The maximum number of training epochs was set to 100 and an early stopping strategy based on the validation set's C-index was implemented. Specifically, if the validation C-index did not improve for 30 consecutive epochs (and the total epoch count exceeded 50) after a 5-epoch warm-up, early stopping was triggered and the best-performing model checkpoint was saved. We conducted tests of the fully trained optimal model on the independent ZN-CRC cohort and systematically compared it with SOTA models including classic pooling methods (Max-Pooling, Mean-Pooling), attention-based MIL models (ABMIL(70),, CLAM-MB, CLAM-SB(62)), a Transformer-based model (TransMIL(68)), and a state-space model-based method (MambaMIL(18)). The model was constructed and trained using Python (v3.10.13) and the PyTorch framework (v2.1.1+cu118).

To compare the dynamic predictive performance of TDAM-CRC with SOTA models, we conducted a time-dependent C-index analysis. In both the TCGA-COADREAD and ZN-CRC cohorts, we first constructed univariable CoxPHs based on the risk scores output by each model. Subsequently, we used the R package pec (v2023.04.12) to calculate and plot time-dependent C-index curves to dynamically assess and compare the prognostic discrimination ability of each model at different time points.

**Model Ablation Study and Interpretability Analysis**

To validate the necessity of each core component in the TDAM-CRC model, we designed and conducted a series of ablation studies. By systematically removing specific modules, we created three model variants: (1) removal of the Transformer module; (2) removal of the dynamic agent attention module; and (3) removal of the SRMamba module. We trained these three variants using the same strategy as the main model and compared their predictive performance with the complete TDAM-CRC model. Furthermore, to investigate the contribution of each module to information aggregation, we performed ERF analysis on the complete model and its three ablated variants(71). ERF analysis visualizes the extent of the model's output dependency on the input features by calculating the gradient of the output layer's center point with respect to the input feature map.

To investigate the pathological basis of the TDAM-CRC model's decisions and to visualize its internal decision-making process, we generated WSI attention heatmaps. The heatmap generation process followed the CLAM framework(62) and was performed as follows: first, the feature vectors of the image patches and their spatial coordinates for a given WSI were loaded. Second, in inference mode, this set of patch features was fed into the trained TDAM-CRC model for a forward pass. Finally, the model's internal attention weights corresponding to the four prognostic bins were mapped back to the original spatial coordinates of the WSI to generate the attention heatmap. In the heatmap, high-weight regions (displayed in warm colors) intuitively indicate the histological areas that the model focused on when making prognostic predictions, thereby providing a morphological basis for the model's interpretability.

**Model Risk Stratification and Prognostic Value Assessment**

Based on the risk scores output by the TDAM-CRC model, we used the median value as a cutoff to stratify patients in the TCGA-COADREAD and ZN-CRC cohorts into high-risk and low-risk groups. To explore the association between the model's risk stratification and clinicopathological factors, we used the R package gtsummary (v2.0.4) to compare the distribution of baseline characteristics, such as age, sex, histological grades, and TNM stage, between the high- and low-risk groups(72). To evaluate the clinical prognostic value of this risk stratification, we conducted KM survival analysis using the R package survminer (v0.5.0), comparing the OS between the two groups with the log-rank test. We also calculated the RMST using the R package survRM2 (v1.0-4). Additionally, we compared the OS between the two groups within subgroups of different clinical variables using KM survival analysis.

To compare the prognostic performance of the TDAM-CRC risk score with clinicopathological staging, we used the R package survivalROC (v1.0.3.1) to plot time-dependent ROC curves and calculated the AUC at 1, 3, and 5 years. For a robust performance estimate, we employed the bootstrap resampling method (500 samplings with replacement) to construct and compare the confidence intervals of the AUCs for prognostic prediction by the TDAM-CRC risk score and clinicopathological staging. To assess whether the TDAM-CRC risk score is an independent prognostic factor, we employed CoxPHs using the R package survival (v3.5-5). Univariable CoxPHs were constructed first, followed by multivariable CoxPHs that included all variables with p-value $< 0.05$ in the univariable models.

**Functional Enrichment Analysis of Risk Subtypes**

To uncover the potential biological mechanisms underlying the differences between the risk subtypes, we conducted a comprehensive bioinformatics analysis based on the RNA-seq data of the TCGA-COADREAD cohort. First, we used the R package DESeq2 (v1.40.2) to perform a DEG analysis between the high- and low-risk groups, with the screening thresholds set at an adjusted p-value < 0.05 and $|log_2(FoldChange)| > 0.5$(73). Subsequently, we performed GO(74) and KEGG(75) pathway enrichment analyses on the DEGs using the R package clusterProfiler (v4.8.2)(76), with a q-value < 0.05 as the threshold for statistical significance. Additionally, to systematically explore changes in pathway activity at a functional level, we conducted GSEA(77). The background gene sets for GSEA were sourced from the MSigDB database's C2 (curated gene sets), C8 (cell type signature gene sets), and Hallmark (h.all) collections (v2024.1.Hs) (78).

**TME Characterization and Drug Prediction**

To characterize the TME features associated with prognostic risk, we performed a series of multi-omics analyses on the TCGA-COADREAD cohort. We systematically evaluated the differences in expression levels, DNA methylation, and amplification and deletion frequencies of key immunomodulatory genes between the high- and low-risk groups(79). Using the ssGSEA algorithm in the R package IOBR (v0.99.9) (with mini_gene_count set to 5), we estimated the biological pathway activity scores for each tumor sample based on its built-in signature collection(80, 81). Concurrently, we employed seven algorithms—MCPcounter(82), EPIC(83), xCell l(84), CIBERSORT(85), IPS(86), quanTIseq(87), and TIMER(88)—to quantify the infiltration scores of various immune cells in the TME. We used the Wilcoxon rank-sum test to compare the pathway scores and immune cell scores between the high- and low-risk groups and visualized the significantly different items (p-value < 0.05) using heatmaps. To predict potential therapeutic drugs, we conducted an analysis using the eXtreme Sum (XSum) method of the CMap(89, 90). The input for this analysis was the top 150 upregulated and top 150 downregulated DEGs between the high- and low-risk groups in the TCGA-COADREAD cohort. Using the standardized CMap scores, we identified potential anti-CRC drugs that were most relevant to the gene expression profiles of the risk subtypes.

**Identification and Validation of a Prognosis-Related Hub Gene via the Deep Learning Model**

To establish a link between the deep learning risk phenotype and potential molecular drivers, we designed a pipeline to identify key hub genes. This pipeline consisted of two parallel branches: (1) Pathomic Feature Selection: We first calculated the Spearman correlation between the 512-dimensional deep learning features extracted by CONCH and the TDAM-CRC risk score, with an initial screening criterion of |correlation coefficient| ⩾ 0.2 and false discovery rate (FDR) < 0.05. Subsequently, we used the R package glmnet (v4.1-7) to input the selected features into an elastic net regression model (alpha=0.5) to further compress the features, ultimately obtaining a core set of pathomic features most robustly associated with the risk score. We used these associations to construct a pathology-model interaction network, which was visualized using the R packages tidygraph (v1.2.3) and ggraph (v2.1.0).
(2) Prognosis-Related Gene Selection: We employed univariable CoxPHs on the entire RNA-seq data of the TCGA-COADREAD cohort, defining genes with p-value < 0.01 and a hazard ratio (HR) not equal to 1 as prognosis-related genes.

Next, we integrated the results from both branches by calculating the Spearman correlation between the core pathomic feature set and the prognosis-related gene set. We selected significantly correlated feature-gene pairs using a threshold of |correlation coefficient| ⩾ 0.2 and FDR < 0.05. These associations were used to construct a pathology-gene interaction network and we employed eigenvector centrality to measure the importance of each node within the network.

We conducted a comprehensively validation of the prognostic value and biological function of the hub gene with the highest eigenvector centrality. In the TCGA-COADREAD cohort, we first assessed the difference in OS between high- and low-expression groups (split by the median) using KM survival analysis. We then included its expression level along with clinical variables in univariable and multivariable CoxPHs to test whether it was an independent prognostic factor (the multivariable model included variables with p-value < 0.05 from the univariable model). To explore its regulatory mechanisms and cellular expression profile, we used the scCancerExplorer database to analyze its expression and promoter methylation levels in tumor versus adjacent tissues from CRC patients in the TCGA database(91). Additionally, we analyzed its promoter methylation status and its expression distribution across different cell subpopulations using the single-cell datasets GSE97693(92) and HRA000201(93) respectively.

**Construction and Evaluation of a Clinical Prediction Nomogram**

To develop an individualized prognostic assessment tool that integrates multidimensional information and is easy to use in a clinical setting, we constructed a nomogram model. In the TCGA-COADREAD cohort, we included the variables confirmed as independent prognostic factors in the multivariable CoxPHs (including the TDAM-CRC risk score, age, and clinical stage) into a new CoxPH and constructed the nomogram using the R package regplot (v1.1).

We comprehensively evaluated the performance of the nomogram. Its calibration was visually assessed using calibration curves to check the consistency between predicted survival probabilities and actual observed probabilities. Discrimination was first assessed by calculating the C-index using the R package Hmisc (v5.1-0), followed by calculating the area under the time-dependent ROC curves at 1, 3, and 5 years. Clinical utility was evaluated using DCA with the R package ggDCA (v1.2) to quantify its net benefit across different risk thresholds. To validate the model's generalizability, we calculated the nomogram score for each patient in the ZN-CRC cohort using the R package nomogramFormula (v1.2.0.0) and repeated the aforementioned calibration curve, time-dependent ROC, and DCA on this external validation cohort. To enhance the model's interpretability, we applied SHAP analysis, using the R packages treeshap (v0.3.1) and shapviz (v0.9.7) to calculate and visualize the contribution of each variable to the prediction outcomes of individual patients in the TCGA-COADREAD cohort and their interactions.

**Statistical Analysis**

All statistical analyses and visualizations were performed in R software (v4.3.1). Comparisons of continuous variables between two groups were conducted using either a t-test or a Wilcoxon rank-sum test, depending on the data distribution. Comparisons of categorical variables were performed using a chi-squared test or Fisher's exact test. Survival analysis was conducted using the KM method, and differences between groups were assessed using the log-rank test. Unless otherwise specified, all statistical tests were two-sided, and a p-value $< 0.05$ was considered statistically significant.

**Ethics approval:** The construction of the ZN-CRC cohort was approved by the Institutional Review Board of Zhongnan Hospital of Wuhan University (No. 2025010K).

**Data availability**: The data used to support the findings of this study are available from the corresponding authors upon reasonable request.

**Code availability**: The source code for the TDAM-CRC model is publicly available on GitHub at https://github.com/liuxiaoping2020/TDAM-CRC.

**Declaration of interests:** The authors declare no competing interests.

**Author contributions:** Zisong Wang: Formal analysis, Software, Visualization, Writing - Original

Draft, Writing - Review & Editing; Xuanyu Wang: Software, Visualization, Writing - Review & Editing; Hang Chen: Writing - Review & Editing; Haizhou Wang: Writing - Review & Editing; Yuxin Chen: Writing - Review & Editing; Yihang Xu: Writing - Review & Editing; Yunhe Yuan: Writing - Review & Editing; Lihuan Luo: Writing - Review & Editing; Xitong Ling: Conceptualization, Writing - Review & Editing; Xiaoping Liu: Conceptualization, Supervision, Writing - Review & Editing.

**Acknowledgements:** This study was supported by the Luojia Undergraduate Innovative Research Fund of Wuhan University (104862025bk0036). We thank Zidu Wang (School of Artificial Intelligence, University of Chinese Academy of Sciences) and Shang'ao Li (School of Computer Science, Nanjing University) for valuable advice and helpful discussions regarding the deep learning model development and training. The numerical calculations in this paper have been done on the supercomputing system in the Supercomputing Center of Wuhan University.

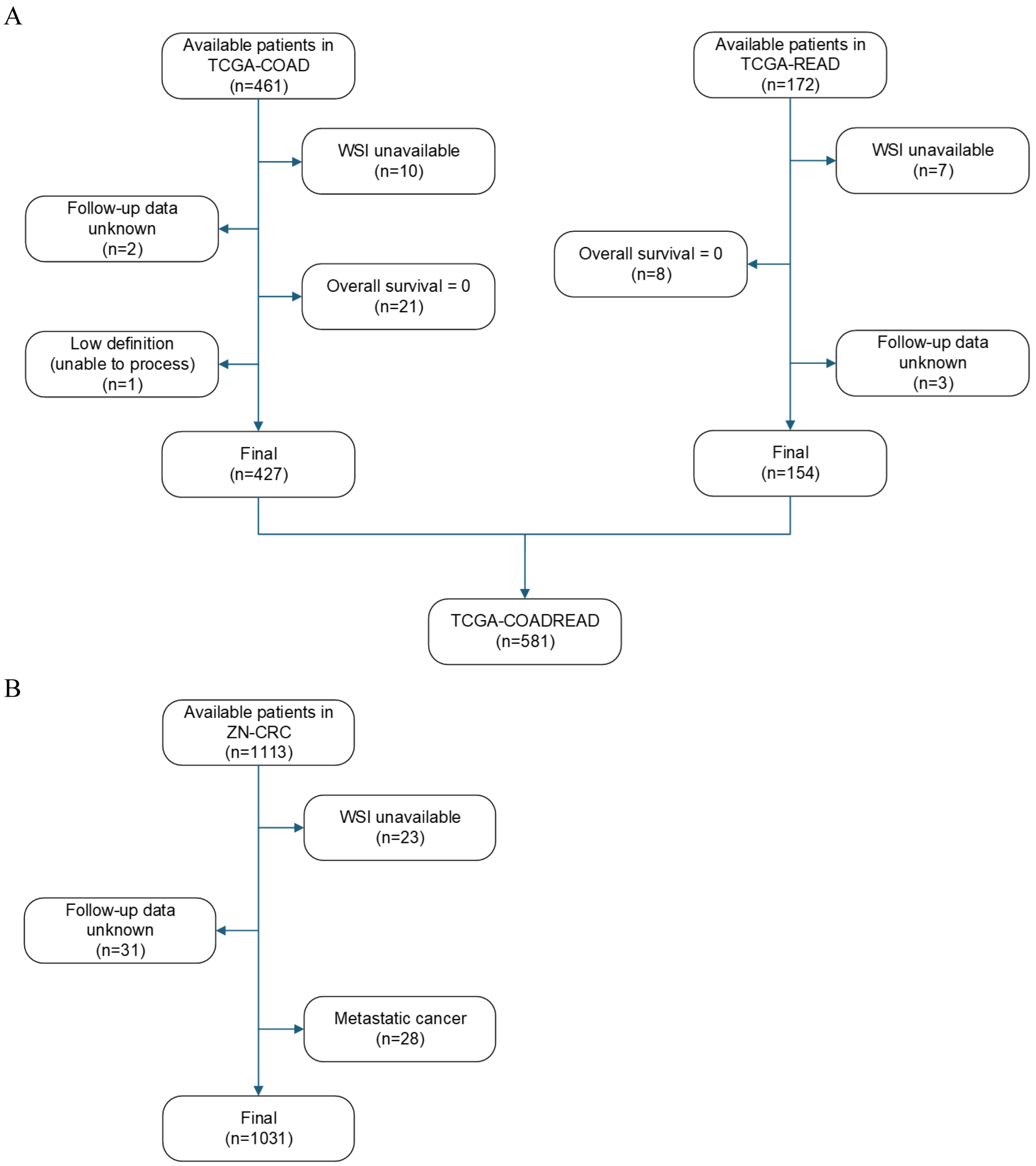

**Supplementary Figure 1.** Flow diagram of patient selection for the (A) TCGA-COADREAD and (B) ZN-CRC cohorts.

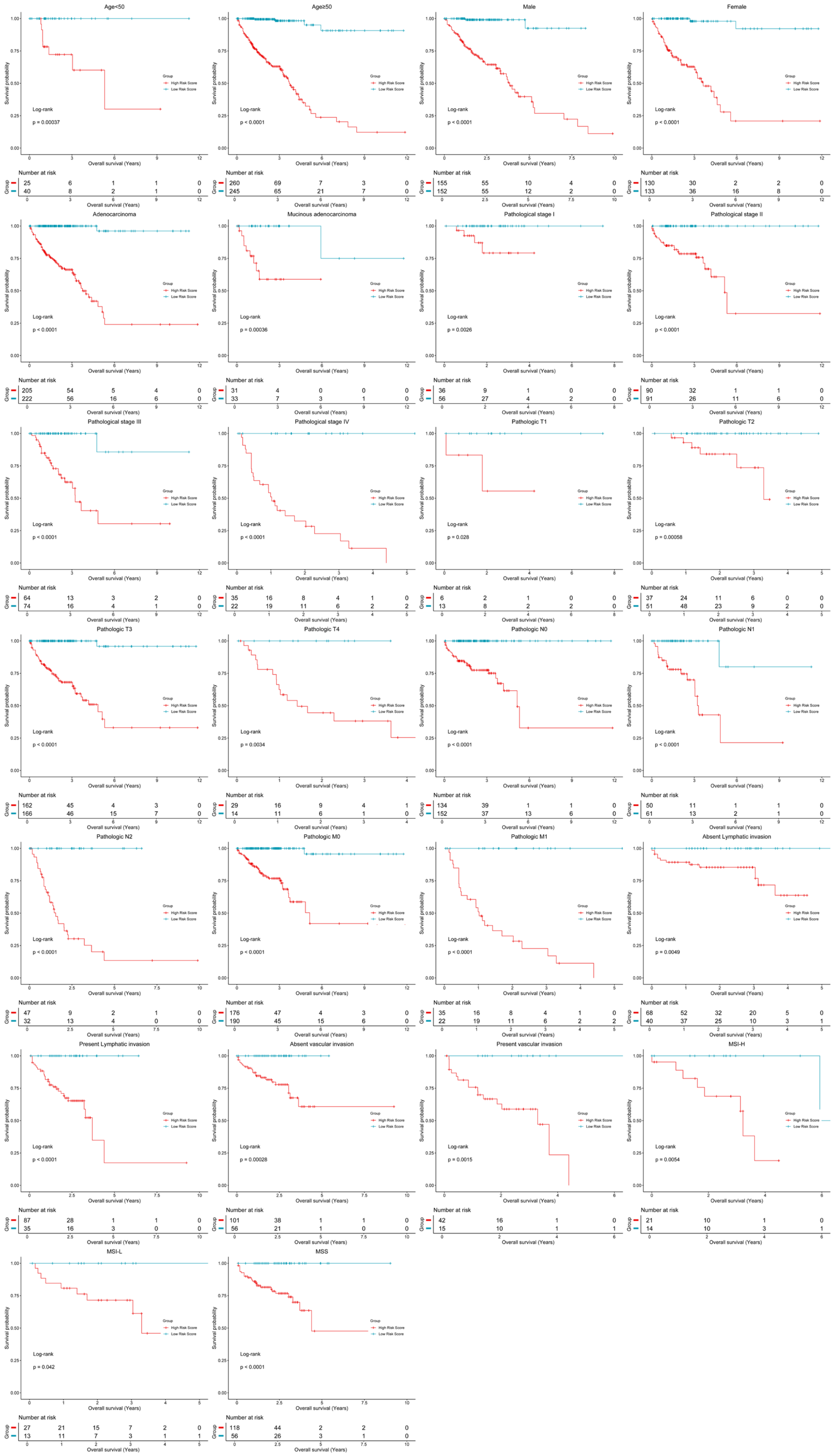

**Supplementary Figure 2.** KM survival analysis of patients in the high- and low-risk groups in the TCGA-COADREAD cohort, stratified by different clinical variables.

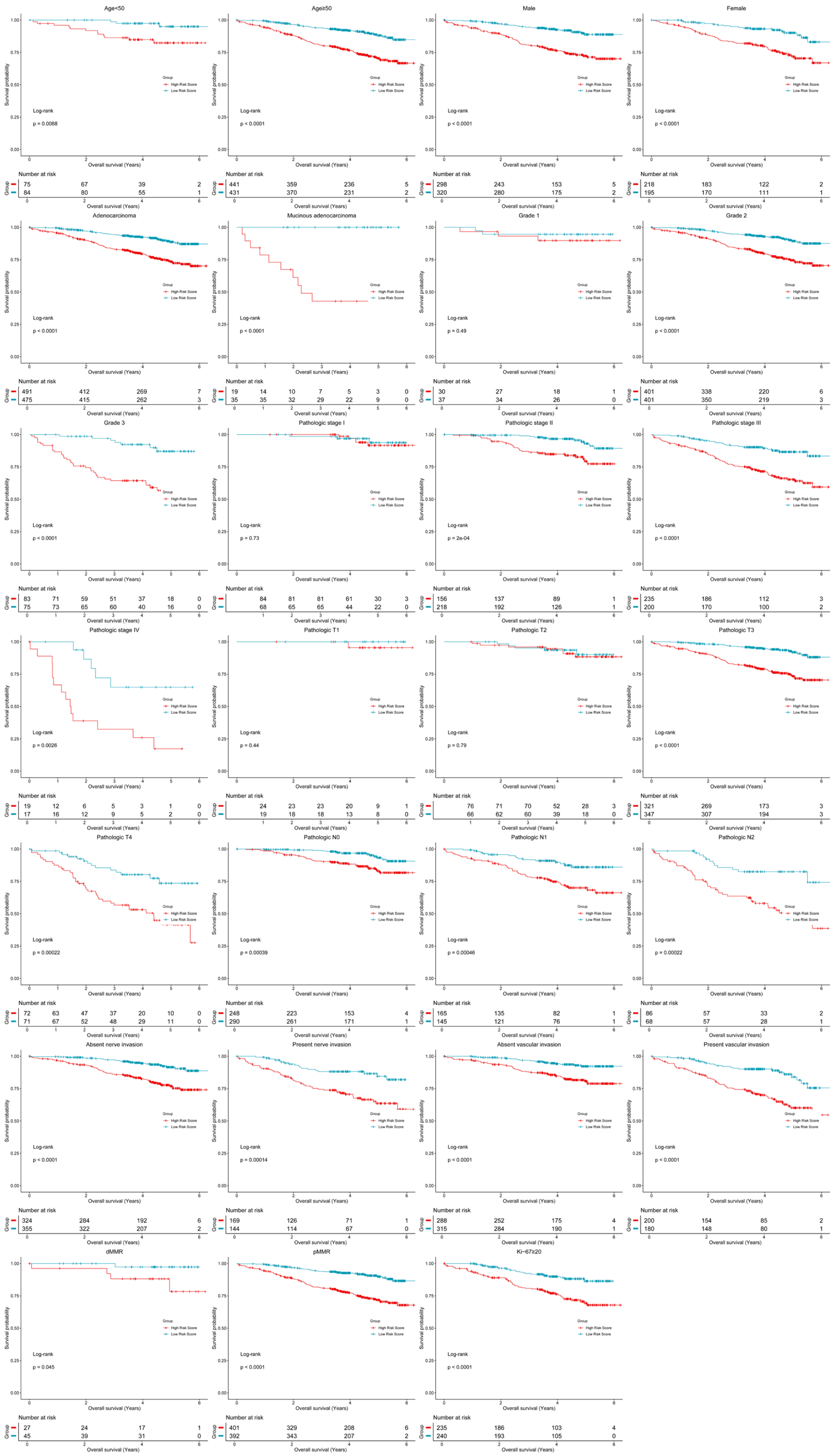

**Supplementary Figure 3.** KM survival analysis of patients in the high- and low-risk groups in the ZN-CRC cohort, stratified by different clinical variables.

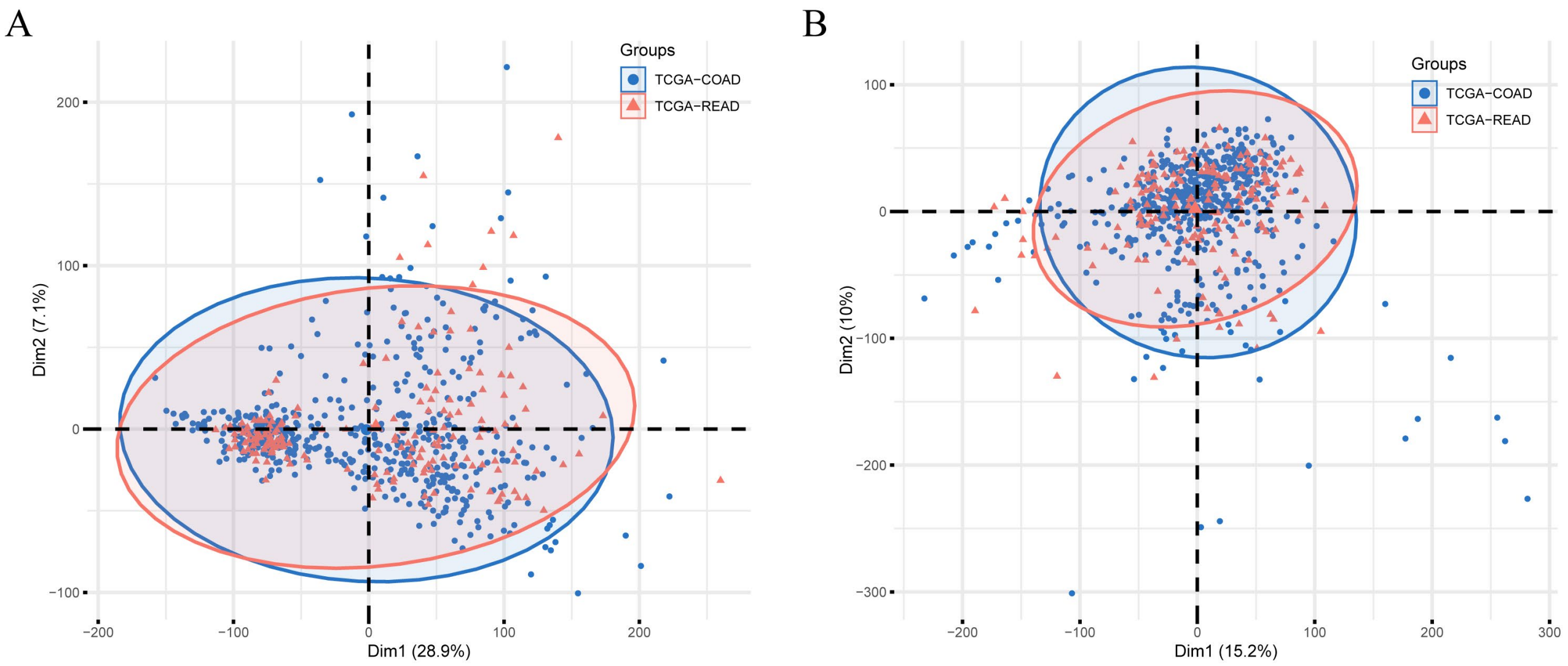

**Supplementary Figure 4.** Principal component analysis (PCA) plots of the batch-corrected, merged TCGA-COAD and TCGA-READ RNA-seq data, based on (A) Count and (B) TPM formats.

**Supplementary Table 1.** Clinicopathological characteristics of patients in the TCGA-COADREAD cohort.

| | | Risk Group | | | |
|---|---|---|---|---|---|
| **Characteristic** | **N*** | **Overall**, N = 570 (100%)† | **low**, N = 285 (50%)† | **high**, N = 285 (50%)† | **p-value**** |
| **Age** | 570 | 68.00(58.00, 75.00) | 64.00(56.00, 73.00) | 70.00(60.00, 77.00) | <0.001 |
| **Gender** | 570 | | | | 0.8 |
| *male* | | 307(54%) | 152(53%) | 155(54%) | |
| *female* | | 263(46%) | 133(47%) | 130(46%) | |
| **Race** | 345 | | | | 0.6 |
| *white* | | 271(79%) | 163(78%) | 108(80%) | |
| *black or african american* | | 61(18%) | 40(19%) | 21(16%) | |
| *other* | | 13(3.8%) | 7(3.3%) | 6(4.4%) | |
| **Histological Type** | 501 | | | | 0.8 |
| *adenocarcinoma* | | 427(85%) | 222(86%) | 205(85%) | |
| *mucinous adenocarcinoma* | | 64(13%) | 33(13%) | 31(13%) | |
| *other* | | 10(2.0%) | 4(1.5%) | 6(2.5%) | |
| **Pathological Stage** | 469 | | | | 0.064 |
| *0* | | 1(0.2%) | 1(0.4%) | 0(0%) | |
| *I* | | 92(20%) | 56(23%) | 36(16%) | |
| *II* | | 181(39%) | 91(37%) | 90(40%) | |
| *III* | | 138(29%) | 74(30%) | 64(28%) | |
| *IV* | | 57(12%) | 22(9.0%) | 35(16%) | |
| **Pathological T Stage** | 479 | | | | 0.02 |
| *Tis* | | 1(0.2%) | 1(0.4%) | 0(0%) | |
| *T1* | | 19(4.0%) | 13(5.3%) | 6(2.6%) | |
| *T2* | | 88(18%) | 51(21%) | 37(16%) | |
| *T3* | | 328(68%) | 166(68%) | 162(69%) | |
| *T4* | | 43(9.0%) | 14(5.7%) | 29(12%) | |
| **Pathological N Stage** | 476 | | | | 0.1 |
| *N0* | | 286(60%) | 152(62%) | 134(58%) | |
| *N1* | | 111(23%) | 61(25%) | 50(22%) | |
| *N2* | | 79(17%) | 32(13%) | 47(20%) | |
| **Pathological M Stage** | 473 | | | | 0.031 |
| *M0* | | 366(77%) | 190(78%) | 176(77%) | |
| *M1* | | 57(12%) | 22(9.0%) | 35(15%) | |
| *Mx* | | 50(11%) | 32(13%) | 18(7.9%) | |
| **Lymphatic Invasion** | 230 | | | | 0.2 |
| *absent* | | 108(47%) | 40(53%) | 68(44%) | |
| *present* | | 122(53%) | 35(47%) | 87(56%) | |
| **Vascular Invasion** | 214 | | | | 0.2 |
| *absent* | | 157(73%) | 56(79%) | 101(71%) | |
| *present* | | 57(27%) | 15(21%) | 42(29%) | |
| **MSI** | 249 | | | | 0.7 |
| *MSS* | | 174(70%) | 56(67%) | 118(71%) | |
| *MSI-L* | | 40(16%) | 13(16%) | 27(16%) | |
| *MSI-H* | | 35(14%) | 14(17%) | 21(13%) | |

* N Non-missing

† median (interquartile range, IQR) for continuous; n (%) for categorical

** Wilcoxon rank sum test; Pearson's Chi-squared test; Fisher's exact test

**Supplementary Table 2.** Clinicopathological characteristics of patients in the ZN-CRC cohort.

| Characteristic | N* | Risk Group | | | p-value** |
|---|---|---|---|---|---|
| | | **Overall**, N = 1031 (100%)† | **low**, N = 515 (50%)† | **high**, N = 516 (50%)† | |
| **Age** | 1,031 | 63.00(54.00, 70.00) | 61.00(53.00, 68.00) | 65.00(55.00, 71.00) | <0.001 |
| **Gender** | 1,031 | | | | 0.2 |
| *male* | | 618(60%) | 320(62%) | 298(58%) | |
| *female* | | 413(40%) | 195(38%) | 218(42%) | |
| **Histological Type** | 1,031 | | | | 0.078 |
| *adenocarcinoma* | | 966(94%) | 475(92%) | 491(95%) | |
| *mucinous adenocarcinoma* | | 54(5.2%) | 35(6.8%) | 19(3.7%) | |
| *other* | | 11(1.1%) | 5(1.0%) | 6(1.2%) | |
| **Histological Grade** | 1,027 | | | | 0.6 |
| *G1* | | 67(6.5%) | 37(7.2%) | 30(5.8%) | |
| *G2* | | 802(78%) | 401(78%) | 401(78%) | |
| *G3* | | 158(15%) | 75(15%) | 83(16%) | |
| **Pathological Stage** | 1,002 | | | | <0.001 |
| *0* | | 5(0.5%) | 0(0%) | 5(1.0%) | |
| *I* | | 152(15%) | 68(14%) | 84(17%) | |
| *II* | | 374(37%) | 218(43%) | 156(31%) | |
| *III* | | 435(43%) | 200(40%) | 235(47%) | |
| *IV* | | 36(3.6%) | 17(3.4%) | 19(3.8%) | |
| **Pathological T Stage** | 1,002 | | | | 0.12 |
| *Tis* | | 5(0.5%) | 0(0%) | 5(1.0%) | |
| *T1* | | 43(4.3%) | 19(3.8%) | 24(4.8%) | |
| *T2* | | 143(14%) | 66(13%) | 77(15%) | |
| *T3* | | 668(67%) | 347(69%) | 321(64%) | |
| *T4* | | 143(14%) | 71(14%) | 72(14%) | |
| **Pathological N Stage** | 1,002 | | | | 0.036 |
| *N0* | | 538(54%) | 290(58%) | 248(50%) | |
| *N1* | | 310(31%) | 145(29%) | 165(33%) | |
| *N2* | | 154(15%) | 68(14%) | 86(17%) | |
| **Vascular Invasion** | 983 | | | | 0.14 |
| *absent* | | 603(61%) | 315(64%) | 288(59%) | |
| *present* | | 380(39%) | 180(36%) | 200(41%) | |
| **Nerve Invasion** | 992 | | | | 0.066 |
| *absent* | | 679(68%) | 355(71%) | 324(66%) | |
| *present* | | 313(32%) | 144(29%) | 169(34%) | |
| **MMR Status** | 865 | | | | 0.034 |
| *pMMR* | | 793(92%) | 392(90%) | 401(94%) | |
| *dMMR* | | 72(8.3%) | 45(10%) | 27(6.3%) | |
| **Ki-67** | 480 | 70.00(60.00, 80.00) | 80.00(70.00, 80.00) | 70.00(60.00, 80.00) | 0.077 |
| **PD-L1** | 48 | 2.50(0.00, 5.00) | 0.50(0.00, 5.00) | 4.00(2.00, 5.00) | 0.12 |

* N Non-missing

† median (interquartile range, IQR) for continuous; n (%) for categorical

** Wilcoxon rank sum test; Pearson's Chi-squared test; Fisher's exact test

**Supplementary Table 3.** Comparison of RMST for OS between high- and low-risk groups in the TCGA-COADREAD cohort.

| Year | High Risk Group RMST | Low Risk Group RMST | Estimation | LCI | UCI | p-value |
|---|---|---|---|---|---|---|
| 1 | 10.895 | 12.000 | -1.105 | -1.435 | -0.775 | 5.16E-11 |
| 2 | 19.813 | 23.957 | -4.145 | -5.022 | -3.267 | 2.02E-20 |
| 3 | 27.621 | 35.866 | -8.244 | -9.770 | -6.718 | 3.36E-26 |
| 4 | 34.177 | 47.687 | -13.510 | -15.741 | -11.279 | 1.70E-32 |
| 5 | 38.987 | 59.423 | -20.436 | -23.457 | -17.415 | 3.93E-40 |
| 6 | 42.298 | 70.863 | -28.565 | -32.509 | -24.621 | 9.87E-46 |
| 7 | 45.214 | 81.859 | -36.645 | -41.841 | -31.449 | 1.87E-43 |
| 8 | 47.708 | 92.855 | -45.147 | -51.722 | -38.573 | 2.73E-41 |
| 9 | 49.600 | 103.851 | -54.251 | -62.169 | -46.333 | 4.10E-41 |
| 10 | 51.300 | 114.847 | -63.546 | -72.886 | -54.207 | 1.43E-40 |
| 11 | 53.001 | 125.843 | -72.842 | -83.700 | -61.983 | 1.74E-39 |

**Abbreviations:** RMST, restricted mean survival time; LCI, lower limit of 95% confidence interval; UCI, upper limit of 95% confidence interval.

**Supplementary Table 4.** Comparison of RMST for OS between high- and low-risk groups in the ZN-CRC cohort.

| Year | High Risk Group RMST | Low Risk Group RMST | Estimation | LCI | UCI | p-value |
|---|---|---|---|---|---|---|
| 1 | 11.634 | 11.971 | -0.336 | -0.494 | -0.179 | 2.87E-05 |
| 2 | 22.624 | 23.763 | -1.139 | -1.561 | -0.718 | 1.17E-07 |
| 3 | 32.792 | 35.265 | -2.473 | -3.248 | -1.698 | 4.04E-10 |
| 4 | 42.369 | 46.500 | -4.130 | -5.333 | -2.928 | 1.67E-11 |
| 5 | 51.332 | 57.471 | -6.139 | -7.816 | -4.463 | 7.13E-13 |
| 6 | 59.744 | 67.987 | -8.243 | -10.474 | -6.013 | 4.36E-13 |

**Abbreviations:** RMST, restricted mean survival time; LCI, lower limit of 95% confidence interval; UCI, upper limit of 95% confidence interval.

**Supplementary Table 5.** DEGs between the high- and low-risk groups in the TCGA-COADREAD cohort.

| gene | baseMean | log2FoldChange | lfcSE | stat | p-value | p-adjust |
|---|---|---|---|---|---|---|
| HBZ | 1.519 | 2.718 | 0.500 | 5.436 | 5.45E-08 | 7.18E-06 |
| MAGEA12 | 24.811 | 2.644 | 0.422 | 6.259 | 3.88E-10 | 1.73E-07 |
| RHAG | 1.852 | 2.478 | 0.421 | 5.886 | 3.95E-09 | 1.03E-06 |
| ALB | 41.345 | 2.421 | 0.331 | 7.305 | 2.78E-13 | 1.30E-09 |
| HBG2 | 1.102 | 2.128 | 0.561 | 3.795 | 1.48E-04 | 3.08E-03 |
| CSAG1 | 13.746 | 2.107 | 0.356 | 5.925 | 3.12E-09 | 8.98E-07 |
| MAGEC1 | 2.307 | 1.951 | 0.466 | 4.186 | 2.84E-05 | 9.28E-04 |
| DDX53 | 1.534 | 1.937 | 0.616 | 3.145 | 1.66E-03 | 1.66E-02 |
| AHSG | 17.633 | 1.932 | 0.229 | 8.426 | 3.59E-17 | 6.71E-13 |
| LIN28A | 2.574 | 1.815 | 0.323 | 5.619 | 1.92E-08 | 3.40E-06 |
| DLK1 | 3.525 | 1.799 | 0.392 | 4.590 | 4.43E-06 | 2.25E-04 |
| XAGE2 | 2.583 | 1.749 | 0.446 | 3.918 | 8.91E-05 | 2.11E-03 |
| LCN8 | 4.172 | 1.716 | 0.328 | 5.238 | 1.63E-07 | 1.69E-05 |
| VAX1 | 4.537 | 1.711 | 0.354 | 4.835 | 1.33E-06 | 8.85E-05 |
| TCP11 | 25.582 | 1.702 | 0.241 | 7.064 | 1.61E-12 | 3.02E-09 |
| PRODH2 | 1.073 | 1.675 | 0.431 | 3.886 | 1.02E-04 | 2.34E-03 |
| SOHLH1 | 3.085 | 1.647 | 0.315 | 5.228 | 1.71E-07 | 1.73E-05 |
| TMPRSS11E | 7.886 | 1.626 | 0.273 | 5.954 | 2.62E-09 | 7.91E-07 |
| PNMA5 | 12.316 | 1.515 | 0.261 | 5.792 | 6.94E-09 | 1.66E-06 |
| NKX2-5 | 1.087 | 1.509 | 0.453 | 3.330 | 8.70E-04 | 1.07E-02 |
| TYR | 2.326 | 1.479 | 0.360 | 4.109 | 3.98E-05 | 1.19E-03 |
| AL662884.4 | 0.572 | 1.455 | 0.370 | 3.929 | 8.54E-05 | 2.05E-03 |
| KLK5 | 4.764 | 1.436 | 0.316 | 4.547 | 5.45E-06 | 2.66E-04 |
| MAGEA6 | 72.039 | 1.417 | 0.456 | 3.105 | 1.90E-03 | 1.82E-02 |
| APOA2 | 19.397 | 1.397 | 0.270 | 5.181 | 2.21E-07 | 2.17E-05 |
| APOC3 | 2.109 | 1.392 | 0.359 | 3.879 | 1.05E-04 | 2.38E-03 |
| AC074143.1 | 0.520 | 1.377 | 0.377 | 3.655 | 2.58E-04 | 4.64E-03 |
| MIA | 4.175 | 1.370 | 0.243 | 5.631 | 1.79E-08 | 3.35E-06 |
| PRSS56 | 48.243 | 1.336 | 0.291 | 4.590 | 4.43E-06 | 2.25E-04 |
| ANXA8L1 | 1.317 | 1.330 | 0.321 | 4.143 | 3.43E-05 | 1.06E-03 |
| OTX2 | 0.840 | 1.322 | 0.450 | 2.935 | 3.34E-03 | 2.67E-02 |
| IGFBP1 | 25.449 | 1.302 | 0.214 | 6.087 | 1.15E-09 | 3.98E-07 |
| LCN15 | 1041.308 | 1.281 | 0.255 | 5.022 | 5.12E-07 | 4.04E-05 |
| DPPA4 | 2.605 | 1.272 | 0.203 | 6.261 | 3.83E-10 | 1.73E-07 |
| HEMGN | 1.035 | 1.269 | 0.300 | 4.226 | 2.38E-05 | 8.10E-04 |
| PASD1 | 0.729 | 1.265 | 0.414 | 3.055 | 2.25E-03 | 2.05E-02 |
| TMEFF1 | 0.470 | 1.254 | 0.372 | 3.369 | 7.54E-04 | 9.70E-03 |
| SAGE1 | 1.005 | 1.246 | 0.370 | 3.369 | 7.54E-04 | 9.70E-03 |
| TKTL1 | 9.290 | 1.235 | 0.225 | 5.479 | 4.28E-08 | 5.93E-06 |
| NKX6-3 | 13.182 | 1.233 | 0.341 | 3.611 | 3.05E-04 | 5.23E-03 |
| ADCY8 | 1.841 | 1.221 | 0.354 | 3.453 | 5.54E-04 | 7.85E-03 |
| TF | 42.137 | 1.209 | 0.204 | 5.919 | 3.24E-09 | 9.18E-07 |
| SKOR2 | 0.654 | 1.203 | 0.352 | 3.421 | 6.24E-04 | 8.49E-03 |
| DCT | 9.863 | 1.197 | 0.212 | 5.641 | 1.70E-08 | 3.20E-06 |
| HBQ1 | 5.794 | 1.189 | 0.220 | 5.410 | 6.30E-08 | 7.87E-06 |
| FTCD | 5.729 | 1.181 | 0.186 | 6.347 | 2.20E-10 | 1.18E-07 |
| SLC6A2 | 2.458 | 1.176 | 0.300 | 3.920 | 8.84E-05 | 2.10E-03 |
| C19orf81 | 5.057 | 1.151 | 0.224 | 5.143 | 2.70E-07 | 2.48E-05 |
| MT4 | 4.788 | 1.149 | 0.423 | 2.716 | 6.61E-03 | 4.27E-02 |
| OTOP1 | 0.624 | 1.140 | 0.432 | 2.637 | 8.37E-03 | 5.00E-02 |
| PROK1 | 0.955 | 1.132 | 0.367 | 3.084 | 2.04E-03 | 1.91E-02 |
| SAP25 | 0.487 | 1.126 | 0.308 | 3.657 | 2.55E-04 | 4.61E-03 |
| HP | 10.178 | 1.124 | 0.224 | 5.012 | 5.38E-07 | 4.21E-05 |
| NPIPB13 | 27.325 | 1.120 | 0.155 | 7.245 | 4.31E-13 | 1.61E-09 |
| LRRC26 | 286.141 | 1.119 | 0.212 | 5.272 | 1.35E-07 | 1.44E-05 |
| CTCFL | 2.505 | 1.117 | 0.216 | 5.168 | 2.37E-07 | 2.26E-05 |
| APOA5 | 1.679 | 1.115 | 0.310 | 3.602 | 3.16E-04 | 5.33E-03 |
| FO393400.1 | 3.918 | 1.106 | 0.245 | 4.515 | 6.32E-06 | 2.97E-04 |
| C10orf95 | 4.588 | 1.099 | 0.200 | 5.488 | 4.07E-08 | 5.84E-06 |
| AL135905.2 | 94.517 | 1.096 | 0.165 | 6.659 | 2.76E-11 | 2.25E-08 |
| SP8 | 33.073 | 1.088 | 0.205 | 5.306 | 1.12E-07 | 1.24E-05 |
| UPK1A | 7.845 | 1.082 | 0.190 | 5.702 | 1.18E-08 | 2.43E-06 |
| VPREB1 | 0.968 | 1.079 | 0.397 | 2.718 | 6.56E-03 | 4.24E-02 |
| SFTPC | 0.954 | 1.078 | 0.373 | 2.889 | 3.86E-03 | 2.93E-02 |
| OTP | 1.207 | 1.074 | 0.233 | 4.600 | 4.22E-06 | 2.18E-04 |
| HBA1 | 24.631 | 1.070 | 0.210 | 5.092 | 3.54E-07 | 3.02E-05 |
| CDC42EP2 | 2.716 | 1.068 | 0.192 | 5.554 | 2.79E-08 | 4.49E-06 |
| AC020922.1 | 0.392 | 1.068 | 0.387 | 2.763 | 5.73E-03 | 3.87E-02 |
| LHX1 | 3.009 | 1.068 | 0.320 | 3.339 | 8.40E-04 | 1.04E-02 |
| VHLL | 0.367 | 1.066 | 0.337 | 3.160 | 1.58E-03 | 1.60E-02 |
| TEX15 | 0.814 | 1.053 | 0.340 | 3.100 | 1.93E-03 | 1.85E-02 |
| PSCA | 52.417 | 1.052 | 0.203 | 5.189 | 2.11E-07 | 2.08E-05 |
| CKLF-CMTM1 | 2.220 | 1.048 | 0.209 | 5.007 | 5.53E-07 | 4.24E-05 |
| APOB | 40.904 | 1.048 | 0.251 | 4.176 | 2.97E-05 | 9.53E-04 |
| NKX6-1 | 1.828 | 1.042 | 0.287 | 3.630 | 2.84E-04 | 4.98E-03 |
| F2 | 29.510 | 1.038 | 0.204 | 5.094 | 3.50E-07 | 3.01E-05 |
| AL121845.3 | 3.689 | 1.031 | 0.212 | 4.859 | 1.18E-06 | 8.08E-05 |
| FGG | 6.727 | 1.030 | 0.330 | 3.119 | 1.82E-03 | 1.76E-02 |
| RPRML | 2.650 | 1.018 | 0.283 | 3.597 | 3.22E-04 | 5.41E-03 |
| ZAR1 | 2.511 | 1.005 | 0.219 | 4.600 | 4.22E-06 | 2.18E-04 |
| IGF2 | 755.109 | 0.998 | 0.250 | 3.998 | 6.40E-05 | 1.67E-03 |
| APOA1 | 19.020 | 0.997 | 0.191 | 5.206 | 1.93E-07 | 1.91E-05 |
| MUC16 | 37.251 | 0.995 | 0.210 | 4.736 | 2.18E-06 | 1.30E-04 |
| SOX2 | 93.592 | 0.995 | 0.254 | 3.916 | 8.98E-05 | 2.12E-03 |
| ELOA2 | 0.480 | 0.991 | 0.359 | 2.758 | 5.82E-03 | 3.91E-02 |
| VGLL1 | 9.952 | 0.986 | 0.264 | 3.737 | 1.86E-04 | 3.66E-03 |
| AC098588.2 | 0.439 | 0.982 | 0.299 | 3.287 | 1.01E-03 | 1.18E-02 |
| TRPM1 | 1.107 | 0.973 | 0.251 | 3.883 | 1.03E-04 | 2.35E-03 |
| ITIH2 | 10.185 | 0.972 | 0.186 | 5.234 | 1.66E-07 | 1.71E-05 |
| NAA11 | 0.643 | 0.960 | 0.296 | 3.247 | 1.17E-03 | 1.30E-02 |
| LKAAEAR1 | 1.623 | 0.959 | 0.249 | 3.844 | 1.21E-04 | 2.67E-03 |
| GNG10 | 22.630 | 0.953 | 0.173 | 5.521 | 3.38E-08 | 5.07E-06 |
| SEPTIN5 | 3.207 | 0.946 | 0.160 | 5.906 | 3.50E-09 | 9.63E-07 |
| TBPL2 | 0.745 | 0.943 | 0.240 | 3.926 | 8.64E-05 | 2.07E-03 |
| NOXO1 | 1.395 | 0.940 | 0.333 | 2.820 | 4.80E-03 | 3.39E-02 |
| RNF113B | 0.979 | 0.937 | 0.281 | 3.341 | 8.36E-04 | 1.04E-02 |
| BOD1L2 | 0.453 | 0.935 | 0.332 | 2.820 | 4.81E-03 | 3.39E-02 |
| CYP2W1 | 2220.772 | 0.927 | 0.181 | 5.118 | 3.09E-07 | 2.75E-05 |
| TMEM213 | 3.741 | 0.926 | 0.190 | 4.877 | 1.08E-06 | 7.62E-05 |
| CALML3 | 2.591 | 0.925 | 0.233 | 3.979 | 6.92E-05 | 1.77E-03 |
| PPY | 1.459 | 0.925 | 0.280 | 3.302 | 9.61E-04 | 1.14E-02 |
| PCBP1 | 718.806 | 0.923 | 0.128 | 7.183 | 6.80E-13 | 2.12E-09 |
| LCN6 | 0.576 | 0.916 | 0.256 | 3.576 | 3.48E-04 | 5.68E-03 |
| MDP1 | 2.363 | 0.914 | 0.204 | 4.483 | 7.37E-06 | 3.32E-04 |
| MYBPHL | 18.018 | 0.912 | 0.237 | 3.849 | 1.19E-04 | 2.63E-03 |
| MT1HL1 | 0.306 | 0.907 | 0.334 | 2.714 | 6.65E-03 | 4.28E-02 |
| CLEC2L | 9.385 | 0.906 | 0.178 | 5.087 | 3.64E-07 | 3.08E-05 |
| NMRK2 | 0.554 | 0.904 | 0.305 | 2.965 | 3.03E-03 | 2.49E-02 |
| TMED7-TICAM2 | 2.501 | 0.903 | 0.162 | 5.589 | 2.28E-08 | 3.90E-06 |
| MYG1 | 7.163 | 0.902 | 0.123 | 7.305 | 2.76E-13 | 1.30E-09 |
| NOX5 | 5.894 | 0.898 | 0.147 | 6.129 | 8.85E-10 | 3.38E-07 |
| MUC15 | 2.046 | 0.898 | 0.338 | 2.653 | 7.98E-03 | 4.83E-02 |
| HEPACAM | 1.272 | 0.888 | 0.275 | 3.234 | 1.22E-03 | 1.34E-02 |
| CCKBR | 1.859 | 0.888 | 0.282 | 3.146 | 1.66E-03 | 1.66E-02 |
| CGA | 1.770 | 0.878 | 0.233 | 3.763 | 1.68E-04 | 3.38E-03 |
| KRT74 | 1.995 | 0.875 | 0.319 | 2.741 | 6.12E-03 | 4.04E-02 |
| OBP2B | 12.603 | 0.871 | 0.236 | 3.693 | 2.22E-04 | 4.16E-03 |
| LHX3 | 1.586 | 0.870 | 0.275 | 3.160 | 1.58E-03 | 1.60E-02 |
| SPESP1 | 20.008 | 0.868 | 0.166 | 5.233 | 1.67E-07 | 1.71E-05 |
| FEZF2 | 2.492 | 0.867 | 0.320 | 2.712 | 6.68E-03 | 4.28E-02 |
| GALR3 | 1.694 | 0.862 | 0.209 | 4.123 | 3.73E-05 | 1.13E-03 |
| BCKDHA | 32.166 | 0.862 | 0.142 | 6.059 | 1.37E-09 | 4.67E-07 |
| ACY1 | 57.875 | 0.860 | 0.121 | 7.090 | 1.35E-12 | 2.86E-09 |
| EPHA8 | 6.808 | 0.858 | 0.267 | 3.212 | 1.32E-03 | 1.42E-02 |
| TMEM40 | 9.251 | 0.855 | 0.177 | 4.819 | 1.44E-06 | 9.28E-05 |
| SCO2 | 31.890 | 0.854 | 0.145 | 5.892 | 3.82E-09 | 1.01E-06 |
| IGDCC3 | 5.170 | 0.854 | 0.191 | 4.470 | 7.81E-06 | 3.44E-04 |
| GABRR1 | 14.446 | 0.853 | 0.260 | 3.280 | 1.04E-03 | 1.20E-02 |
| CYB5D1 | 4.495 | 0.853 | 0.145 | 5.899 | 3.65E-09 | 9.90E-07 |
| KASH5 | 2.602 | 0.852 | 0.238 | 3.576 | 3.49E-04 | 5.69E-03 |
| ERVH48-1 | 6.282 | 0.851 | 0.139 | 6.122 | 9.23E-10 | 3.45E-07 |
| CTXND1 | 17.928 | 0.851 | 0.197 | 4.319 | 1.57E-05 | 5.87E-04 |
| MUC19 | 1.861 | 0.849 | 0.244 | 3.481 | 5.00E-04 | 7.31E-03 |
| PCDH8 | 6.901 | 0.846 | 0.222 | 3.802 | 1.43E-04 | 3.02E-03 |
| MT3 | 62.483 | 0.845 | 0.184 | 4.588 | 4.48E-06 | 2.26E-04 |
| GCHFR | 51.365 | 0.841 | 0.128 | 6.574 | 4.91E-11 | 3.53E-08 |
| GBP7 | 3.660 | 0.841 | 0.188 | 4.466 | 7.98E-06 | 3.48E-04 |
| BORCS8-MEF2B | 0.794 | 0.840 | 0.185 | 4.539 | 5.65E-06 | 2.74E-04 |
| HSPB2 | 2.053 | 0.839 | 0.262 | 3.199 | 1.38E-03 | 1.46E-02 |
| ARL2BP | 43.400 | 0.839 | 0.132 | 6.377 | 1.81E-10 | 1.06E-07 |
| F11 | 5.729 | 0.838 | 0.233 | 3.604 | 3.14E-04 | 5.32E-03 |
| KRT84 | 1.877 | 0.835 | 0.311 | 2.684 | 7.27E-03 | 4.53E-02 |
| TXNDC5 | 118.284 | 0.832 | 0.142 | 5.851 | 4.88E-09 | 1.25E-06 |
| MGAT2 | 37.505 | 0.830 | 0.136 | 6.095 | 1.09E-09 | 3.86E-07 |
| ANXA8 | 3.021 | 0.830 | 0.244 | 3.400 | 6.73E-04 | 8.94E-03 |
| PGLYRP2 | 3.129 | 0.825 | 0.215 | 3.839 | 1.23E-04 | 2.71E-03 |
| VWA5B1 | 31.992 | 0.824 | 0.263 | 3.138 | 1.70E-03 | 1.69E-02 |
| CHST13 | 96.745 | 0.816 | 0.142 | 5.732 | 9.95E-09 | 2.14E-06 |
| NPIPB5 | 158.570 | 0.815 | 0.115 | 7.086 | 1.38E-12 | 2.86E-09 |
| CYP26A1 | 7.153 | 0.814 | 0.207 | 3.930 | 8.48E-05 | 2.05E-03 |
| RAB5IF | 691.097 | 0.814 | 0.174 | 4.670 | 3.02E-06 | 1.67E-04 |
| LRP2 | 7.193 | 0.813 | 0.226 | 3.595 | 3.24E-04 | 5.43E-03 |
| CCDC194 | 2.770 | 0.811 | 0.180 | 4.512 | 6.41E-06 | 3.00E-04 |
| PLSCR3 | 4.707 | 0.810 | 0.131 | 6.187 | 6.12E-10 | 2.55E-07 |
| PRSS21 | 100.327 | 0.809 | 0.239 | 3.389 | 7.02E-04 | 9.23E-03 |
| IFNE | 4.232 | 0.808 | 0.268 | 3.020 | 2.53E-03 | 2.19E-02 |
| LINC00514 | 115.875 | 0.804 | 0.163 | 4.931 | 8.17E-07 | 5.92E-05 |
| HRH3 | 1.160 | 0.803 | 0.294 | 2.732 | 6.29E-03 | 4.11E-02 |
| SOAT2 | 8.143 | 0.800 | 0.220 | 3.628 | 2.85E-04 | 5.01E-03 |
| AC012651.1 | 4.977 | 0.798 | 0.144 | 5.548 | 2.89E-08 | 4.62E-06 |
| UNC5A | 31.720 | 0.796 | 0.165 | 4.839 | 1.30E-06 | 8.72E-05 |
| ACP7 | 3.415 | 0.795 | 0.258 | 3.078 | 2.08E-03 | 1.94E-02 |
| SLX1A | 0.809 | 0.792 | 0.281 | 2.813 | 4.91E-03 | 3.44E-02 |
| TUBA3E | 1.490 | 0.790 | 0.218 | 3.618 | 2.97E-04 | 5.12E-03 |
| AC069503.2 | 1.448 | 0.787 | 0.146 | 5.371 | 7.85E-08 | 9.30E-06 |
| CPN2 | 4.142 | 0.785 | 0.207 | 3.797 | 1.46E-04 | 3.06E-03 |
| PGGHG | 2943.610 | 0.782 | 0.114 | 6.884 | 5.80E-12 | 7.23E-09 |
| NMUR2 | 15.827 | 0.778 | 0.260 | 2.991 | 2.78E-03 | 2.34E-02 |
| CAMK2B | 14.147 | 0.775 | 0.152 | 5.091 | 3.56E-07 | 3.02E-05 |
| H2AC16 | 8.172 | 0.773 | 0.173 | 4.482 | 7.38E-06 | 3.32E-04 |
| TSPAN32 | 20.561 | 0.773 | 0.126 | 6.149 | 7.79E-10 | 3.10E-07 |
| OPN1SW | 0.701 | 0.772 | 0.265 | 2.916 | 3.55E-03 | 2.78E-02 |
| FXYD2 | 3.891 | 0.770 | 0.163 | 4.735 | 2.19E-06 | 1.30E-04 |
| MIF | 12.565 | 0.767 | 0.256 | 2.994 | 2.76E-03 | 2.33E-02 |
| SRXN1 | 25.777 | 0.764 | 0.216 | 3.541 | 3.98E-04 | 6.27E-03 |
| AFP | 10.216 | 0.759 | 0.187 | 4.054 | 5.03E-05 | 1.39E-03 |
| MPPED1 | 0.870 | 0.758 | 0.241 | 3.137 | 1.71E-03 | 1.69E-02 |
| RPS10 | 237.091 | 0.755 | 0.132 | 5.727 | 1.02E-08 | 2.14E-06 |
| FOXL2 | 2.208 | 0.754 | 0.249 | 3.026 | 2.48E-03 | 2.17E-02 |
| TOMM5 | 206.509 | 0.754 | 0.117 | 6.467 | 9.97E-11 | 6.22E-08 |
| FOXB1 | 2.334 | 0.753 | 0.202 | 3.737 | 1.86E-04 | 3.66E-03 |
| AC011511.1 | 0.955 | 0.751 | 0.217 | 3.455 | 5.50E-04 | 7.81E-03 |
| ANP32D | 1.712 | 0.751 | 0.188 | 3.989 | 6.63E-05 | 1.71E-03 |
| KLHL35 | 247.949 | 0.749 | 0.108 | 6.916 | 4.65E-12 | 6.21E-09 |
| IRX3 | 45.353 | 0.748 | 0.169 | 4.420 | 9.87E-06 | 4.09E-04 |
| CST6 | 25.611 | 0.748 | 0.182 | 4.103 | 4.07E-05 | 1.21E-03 |
| C4orf36 | 2.477 | 0.746 | 0.171 | 4.356 | 1.33E-05 | 5.20E-04 |
| TRIM54 | 114.828 | 0.744 | 0.196 | 3.804 | 1.42E-04 | 3.01E-03 |
| NR5A1 | 2.497 | 0.743 | 0.224 | 3.321 | 8.95E-04 | 1.09E-02 |
| ADGRD2 | 4.989 | 0.740 | 0.175 | 4.221 | 2.43E-05 | 8.24E-04 |
| PLPPR3 | 4.614 | 0.738 | 0.178 | 4.146 | 3.39E-05 | 1.05E-03 |
| KRTAP10-4 | 0.860 | 0.735 | 0.250 | 2.939 | 3.30E-03 | 2.64E-02 |
| HPN | 147.038 | 0.734 | 0.199 | 3.690 | 2.24E-04 | 4.18E-03 |
| NDUFS7 | 760.879 | 0.733 | 0.118 | 6.238 | 4.43E-10 | 1.93E-07 |
| CDKN2A | 288.457 | 0.731 | 0.133 | 5.480 | 4.25E-08 | 5.93E-06 |
| IPO4 | 27.891 | 0.729 | 0.120 | 6.099 | 1.07E-09 | 3.84E-07 |
| SLC6A19 | 78.336 | 0.729 | 0.231 | 3.154 | 1.61E-03 | 1.62E-02 |
| ERAS | 1.510 | 0.728 | 0.186 | 3.913 | 9.13E-05 | 2.15E-03 |
| NME2 | 264.028 | 0.727 | 0.145 | 5.009 | 5.46E-07 | 4.24E-05 |
| PDXP | 40.196 | 0.727 | 0.153 | 4.763 | 1.90E-06 | 1.16E-04 |
| BARX1 | 14.431 | 0.727 | 0.189 | 3.841 | 1.22E-04 | 2.69E-03 |
| SHISAL2B | 0.831 | 0.724 | 0.202 | 3.587 | 3.34E-04 | 5.52E-03 |
| AC010422.3 | 0.465 | 0.723 | 0.260 | 2.780 | 5.44E-03 | 3.72E-02 |
| GPS2 | 26.987 | 0.723 | 0.108 | 6.691 | 2.22E-11 | 1.89E-08 |
| NDUFA13 | 166.125 | 0.720 | 0.122 | 5.916 | 3.29E-09 | 9.18E-07 |
| FKBP2 | 17.718 | 0.719 | 0.101 | 7.131 | 9.98E-13 | 2.67E-09 |
| XG | 25.464 | 0.719 | 0.155 | 4.652 | 3.28E-06 | 1.79E-04 |
| OTOG | 4.465 | 0.719 | 0.186 | 3.864 | 1.12E-04 | 2.50E-03 |
| PDF | 125.086 | 0.717 | 0.148 | 4.833 | 1.35E-06 | 8.85E-05 |
| BHMG1 | 2.550 | 0.714 | 0.195 | 3.666 | 2.47E-04 | 4.51E-03 |
| FTH1 | 7923.270 | 0.714 | 0.133 | 5.351 | 8.74E-08 | 9.98E-06 |
| PTGER1 | 25.998 | 0.707 | 0.135 | 5.240 | 1.60E-07 | 1.67E-05 |
| AMH | 117.302 | 0.705 | 0.162 | 4.357 | 1.32E-05 | 5.17E-04 |
| TMEM88B | 0.948 | 0.705 | 0.252 | 2.798 | 5.14E-03 | 3.55E-02 |
| UBD | 9.418 | 0.705 | 0.230 | 3.065 | 2.18E-03 | 2.00E-02 |
| SFTPB | 12.606 | 0.705 | 0.201 | 3.499 | 4.68E-04 | 6.99E-03 |
| BEX1 | 10.029 | 0.701 | 0.167 | 4.199 | 2.68E-05 | 8.88E-04 |
| KISS1 | 35.079 | 0.701 | 0.129 | 5.440 | 5.32E-08 | 7.06E-06 |
| ATF7-NPFF | 11.813 | 0.698 | 0.104 | 6.722 | 1.79E-11 | 1.76E-08 |
| CST2 | 66.299 | 0.696 | 0.148 | 4.700 | 2.60E-06 | 1.48E-04 |
| KCNQ2 | 14.209 | 0.695 | 0.211 | 3.300 | 9.66E-04 | 1.15E-02 |
| ADAMTS19 | 13.387 | 0.694 | 0.225 | 3.081 | 2.06E-03 | 1.92E-02 |
| CRLS1 | 240.310 | 0.694 | 0.135 | 5.155 | 2.54E-07 | 2.36E-05 |
| TMEM256-PLSCR3 | 0.577 | 0.693 | 0.217 | 3.196 | 1.39E-03 | 1.47E-02 |
| AC027796.3 | 0.803 | 0.691 | 0.230 | 3.010 | 2.61E-03 | 2.24E-02 |
| OXT | 3.216 | 0.690 | 0.165 | 4.175 | 2.98E-05 | 9.53E-04 |
| HES7 | 5.567 | 0.686 | 0.171 | 4.003 | 6.26E-05 | 1.65E-03 |
| TRPM5 | 74.301 | 0.686 | 0.119 | 5.742 | 9.36E-09 | 2.04E-06 |
| HAPLN2 | 7.003 | 0.685 | 0.169 | 4.066 | 4.78E-05 | 1.35E-03 |
| GABARAP | 19.236 | 0.685 | 0.102 | 6.711 | 1.93E-11 | 1.76E-08 |
| FOXN4 | 1.687 | 0.684 | 0.256 | 2.670 | 7.58E-03 | 4.66E-02 |
| TAX1BP3 | 32.901 | 0.681 | 0.152 | 4.484 | 7.34E-06 | 3.32E-04 |
| ALDOA | 1261.148 | 0.678 | 0.118 | 5.759 | 8.44E-09 | 1.88E-06 |
| ABCB6 | 47.631 | 0.676 | 0.109 | 6.197 | 5.75E-10 | 2.45E-07 |
| FMC1 | 51.625 | 0.673 | 0.122 | 5.540 | 3.02E-08 | 4.75E-06 |
| A2ML1 | 8.123 | 0.670 | 0.193 | 3.464 | 5.31E-04 | 7.62E-03 |
| AL162231.1 | 3.182 | 0.668 | 0.119 | 5.619 | 1.93E-08 | 3.40E-06 |
| PRSS33 | 297.900 | 0.668 | 0.188 | 3.551 | 3.84E-04 | 6.10E-03 |
| TTR | 26.745 | 0.667 | 0.209 | 3.192 | 1.41E-03 | 1.49E-02 |
| EIF4EBP3 | 56.492 | 0.661 | 0.119 | 5.543 | 2.97E-08 | 4.71E-06 |
| MTHFS | 51.573 | 0.657 | 0.104 | 6.346 | 2.21E-10 | 1.18E-07 |
| OR2B6 | 3.010 | 0.656 | 0.162 | 4.055 | 5.02E-05 | 1.38E-03 |
| BHMT | 5.372 | 0.653 | 0.239 | 2.734 | 6.25E-03 | 4.10E-02 |
| IFNL1 | 1.253 | 0.651 | 0.232 | 2.811 | 4.94E-03 | 3.45E-02 |
| TH | 51.792 | 0.649 | 0.144 | 4.508 | 6.55E-06 | 3.04E-04 |
| BTBD17 | 1.620 | 0.649 | 0.222 | 2.925 | 3.45E-03 | 2.73E-02 |
| AC068580.4 | 14.022 | 0.648 | 0.114 | 5.697 | 1.22E-08 | 2.49E-06 |
| UNC93A | 169.268 | 0.648 | 0.165 | 3.930 | 8.49E-05 | 2.05E-03 |
| UPK2 | 12.883 | 0.647 | 0.208 | 3.111 | 1.86E-03 | 1.80E-02 |
| ALPK3 | 716.234 | 0.644 | 0.136 | 4.726 | 2.29E-06 | 1.34E-04 |
| SPOUT1 | 7.506 | 0.643 | 0.109 | 5.897 | 3.71E-09 | 9.91E-07 |
| CALHM6 | 51.391 | 0.641 | 0.149 | 4.307 | 1.65E-05 | 6.15E-04 |
| NUDT3 | 75.592 | 0.640 | 0.124 | 5.166 | 2.39E-07 | 2.26E-05 |
| SVOPL | 13.825 | 0.640 | 0.205 | 3.121 | 1.80E-03 | 1.76E-02 |
| PRC1 | 34.485 | 0.638 | 0.106 | 6.010 | 1.86E-09 | 5.88E-07 |
| RHBDL1 | 201.312 | 0.638 | 0.101 | 6.290 | 3.17E-10 | 1.56E-07 |
| F7 | 78.063 | 0.636 | 0.181 | 3.518 | 4.35E-04 | 6.66E-03 |
| C8G | 68.823 | 0.636 | 0.126 | 5.041 | 4.64E-07 | 3.71E-05 |
| TFF2 | 228.438 | 0.632 | 0.183 | 3.457 | 5.47E-04 | 7.78E-03 |
| SELENOH | 571.745 | 0.629 | 0.090 | 6.985 | 2.86E-12 | 4.45E-09 |
| SLX1B | 0.982 | 0.629 | 0.232 | 2.709 | 6.74E-03 | 4.30E-02 |
| OR2A12 | 1.059 | 0.628 | 0.191 | 3.286 | 1.02E-03 | 1.18E-02 |
| MMP17 | 115.473 | 0.627 | 0.123 | 5.097 | 3.46E-07 | 2.99E-05 |
| SLIT1 | 29.112 | 0.627 | 0.129 | 4.861 | 1.17E-06 | 8.07E-05 |
| SMC1B | 11.317 | 0.624 | 0.130 | 4.784 | 1.72E-06 | 1.07E-04 |
| NUPR1 | 259.140 | 0.623 | 0.121 | 5.131 | 2.88E-07 | 2.59E-05 |
| NACA2 | 35.086 | 0.622 | 0.145 | 4.293 | 1.77E-05 | 6.44E-04 |
| RHBG | 5.320 | 0.622 | 0.171 | 3.646 | 2.67E-04 | 4.75E-03 |
| C1QTNF5 | 3.658 | 0.621 | 0.211 | 2.940 | 3.29E-03 | 2.64E-02 |
| H4C1 | 3.908 | 0.621 | 0.228 | 2.731 | 6.32E-03 | 4.13E-02 |
| AP001931.2 | 20.511 | 0.621 | 0.095 | 6.499 | 8.08E-11 | 5.21E-08 |
| MEF2B | 5.365 | 0.620 | 0.121 | 5.111 | 3.21E-07 | 2.83E-05 |
| MESP1 | 116.761 | 0.620 | 0.133 | 4.663 | 3.11E-06 | 1.72E-04 |
| SNRPN | 62.373 | 0.619 | 0.123 | 5.025 | 5.04E-07 | 4.01E-05 |
| TMEM160 | 580.034 | 0.613 | 0.124 | 4.932 | 8.14E-07 | 5.92E-05 |
| C20orf204 | 48.092 | 0.612 | 0.138 | 4.447 | 8.70E-06 | 3.73E-04 |
| MTRNR2L3 | 7.924 | 0.611 | 0.175 | 3.499 | 4.67E-04 | 6.99E-03 |
| PHGR1 | 7632.717 | 0.610 | 0.128 | 4.767 | 1.87E-06 | 1.14E-04 |
| KCNJ4 | 4.282 | 0.609 | 0.189 | 3.213 | 1.31E-03 | 1.42E-02 |
| CA7 | 54.718 | 0.608 | 0.148 | 4.122 | 3.75E-05 | 1.13E-03 |
| CDH20 | 1.603 | 0.606 | 0.178 | 3.412 | 6.46E-04 | 8.72E-03 |
| LCT | 8.164 | 0.606 | 0.178 | 3.395 | 6.85E-04 | 9.06E-03 |
| C11orf98 | 2.999 | 0.605 | 0.183 | 3.312 | 9.28E-04 | 1.11E-02 |
| OR13J1 | 1.346 | 0.605 | 0.218 | 2.776 | 5.51E-03 | 3.75E-02 |
| CHTF8 | 42.943 | 0.605 | 0.114 | 5.292 | 1.21E-07 | 1.31E-05 |
| POU3F2 | 3.801 | 0.604 | 0.142 | 4.260 | 2.04E-05 | 7.26E-04 |
| CNTFR | 47.700 | 0.603 | 0.193 | 3.131 | 1.74E-03 | 1.71E-02 |
| TCF23 | 5.084 | 0.603 | 0.205 | 2.942 | 3.26E-03 | 2.62E-02 |
| PCDHA7 | 4.107 | 0.602 | 0.151 | 3.991 | 6.59E-05 | 1.70E-03 |
| SMIM41 | 6.354 | 0.602 | 0.166 | 3.622 | 2.93E-04 | 5.08E-03 |
| HOXA4 | 12.538 | 0.600 | 0.111 | 5.420 | 5.96E-08 | 7.64E-06 |
| POLR2I | 441.985 | 0.599 | 0.122 | 4.932 | 8.14E-07 | 5.92E-05 |
| IL36RN | 6.528 | 0.599 | 0.214 | 2.803 | 5.06E-03 | 3.51E-02 |
| SLC23A1 | 82.820 | 0.598 | 0.116 | 5.165 | 2.40E-07 | 2.26E-05 |
| CBY3 | 2.073 | 0.598 | 0.159 | 3.768 | 1.65E-04 | 3.35E-03 |
| AL845331.1 | 0.651 | 0.596 | 0.224 | 2.667 | 7.66E-03 | 4.70E-02 |
| MRPL2 | 206.982 | 0.596 | 0.110 | 5.426 | 5.77E-08 | 7.45E-06 |
| TMDD1 | 5.034 | 0.595 | 0.138 | 4.323 | 1.54E-05 | 5.77E-04 |
| GBX2 | 11.807 | 0.592 | 0.176 | 3.356 | 7.90E-04 | 9.99E-03 |
| ENDOG | 252.112 | 0.588 | 0.127 | 4.612 | 3.99E-06 | 2.11E-04 |
| PAGE5 | 1.283 | 0.587 | 0.216 | 2.722 | 6.48E-03 | 4.20E-02 |
| MAP6D1 | 6.422 | 0.587 | 0.128 | 4.574 | 4.78E-06 | 2.38E-04 |
| PTX4 | 1.485 | 0.587 | 0.162 | 3.619 | 2.96E-04 | 5.11E-03 |
| TPGS1 | 214.300 | 0.587 | 0.123 | 4.774 | 1.81E-06 | 1.11E-04 |
| NPIPB12 | 13.101 | 0.586 | 0.103 | 5.671 | 1.42E-08 | 2.77E-06 |
| C12orf57 | 793.605 | 0.586 | 0.104 | 5.629 | 1.81E-08 | 3.35E-06 |
| DKK4 | 100.771 | 0.586 | 0.195 | 3.001 | 2.69E-03 | 2.28E-02 |
| ZNF771 | 208.614 | 0.586 | 0.086 | 6.843 | 7.78E-12 | 9.10E-09 |
| MAT1A | 148.209 | 0.585 | 0.176 | 3.318 | 9.06E-04 | 1.09E-02 |
| LRRN4 | 14.772 | 0.584 | 0.128 | 4.568 | 4.91E-06 | 2.44E-04 |
| OBP2A | 2.951 | 0.584 | 0.189 | 3.096 | 1.96E-03 | 1.86E-02 |
| PSMA6 | 325.184 | 0.584 | 0.098 | 5.965 | 2.44E-09 | 7.49E-07 |
| FBN3 | 4.326 | 0.584 | 0.203 | 2.868 | 4.13E-03 | 3.07E-02 |
| DNASE1L2 | 40.951 | 0.584 | 0.103 | 5.683 | 1.33E-08 | 2.66E-06 |
| TMEM191B | 11.646 | 0.583 | 0.125 | 4.685 | 2.80E-06 | 1.57E-04 |
| EMC6 | 33.620 | 0.583 | 0.101 | 5.800 | 6.61E-09 | 1.63E-06 |
| PMEL | 163.126 | 0.582 | 0.088 | 6.606 | 3.95E-11 | 2.96E-08 |
| MPC1L | 1.201 | 0.582 | 0.200 | 2.901 | 3.72E-03 | 2.86E-02 |
| ERFE | 85.182 | 0.581 | 0.123 | 4.716 | 2.40E-06 | 1.40E-04 |
| PANX2 | 23.998 | 0.579 | 0.133 | 4.370 | 1.24E-05 | 4.93E-04 |
| LTC4S | 39.254 | 0.579 | 0.101 | 5.747 | 9.09E-09 | 2.00E-06 |
| HES4 | 331.742 | 0.577 | 0.117 | 4.950 | 7.41E-07 | 5.48E-05 |
| C15orf61 | 187.183 | 0.575 | 0.088 | 6.554 | 5.60E-11 | 3.88E-08 |
| GEMIN7 | 23.075 | 0.575 | 0.086 | 6.708 | 1.98E-11 | 1.76E-08 |
| GNG5 | 526.163 | 0.573 | 0.105 | 5.458 | 4.81E-08 | 6.47E-06 |
| ADM5 | 38.486 | 0.573 | 0.099 | 5.815 | 6.05E-09 | 1.51E-06 |
| MRPS24 | 23.746 | 0.573 | 0.121 | 4.741 | 2.13E-06 | 1.27E-04 |
| GJC2 | 132.759 | 0.570 | 0.140 | 4.083 | 4.44E-05 | 1.28E-03 |
| DDX25 | 2.001 | 0.570 | 0.170 | 3.357 | 7.88E-04 | 9.99E-03 |
| AP001453.3 | 22.335 | 0.569 | 0.112 | 5.062 | 4.14E-07 | 3.42E-05 |
| H2AC17 | 1.217 | 0.568 | 0.180 | 3.156 | 1.60E-03 | 1.62E-02 |
| CLGN | 42.301 | 0.567 | 0.166 | 3.425 | 6.14E-04 | 8.40E-03 |
| MC1R | 78.558 | 0.567 | 0.095 | 5.942 | 2.81E-09 | 8.34E-07 |
| EGLN1 | 263.593 | 0.563 | 0.092 | 6.146 | 7.97E-10 | 3.10E-07 |
| MMP23B | 7.073 | 0.563 | 0.129 | 4.382 | 1.18E-05 | 4.72E-04 |
| AC008397.2 | 4.308 | 0.563 | 0.169 | 3.331 | 8.64E-04 | 1.06E-02 |
| FEM1A | 65.242 | 0.563 | 0.093 | 6.052 | 1.43E-09 | 4.71E-07 |
| SLC7A10 | 7.789 | 0.561 | 0.204 | 2.753 | 5.91E-03 | 3.94E-02 |
| VWCE | 54.581 | 0.560 | 0.139 | 4.041 | 5.31E-05 | 1.45E-03 |
| GRIN3B | 11.601 | 0.560 | 0.126 | 4.446 | 8.75E-06 | 3.73E-04 |
| NPIPB11 | 46.565 | 0.559 | 0.100 | 5.589 | 2.28E-08 | 3.90E-06 |
| FAM166A | 1.525 | 0.559 | 0.181 | 3.087 | 2.02E-03 | 1.90E-02 |
| TREML2 | 57.079 | 0.559 | 0.155 | 3.594 | 3.26E-04 | 5.45E-03 |
| TPPP2 | 1.877 | 0.558 | 0.169 | 3.304 | 9.54E-04 | 1.14E-02 |
| MRPL38 | 77.418 | 0.558 | 0.090 | 6.171 | 6.79E-10 | 2.76E-07 |
| MTFP1 | 56.691 | 0.557 | 0.110 | 5.056 | 4.29E-07 | 3.52E-05 |
| HOXC6 | 52.429 | 0.557 | 0.205 | 2.721 | 6.51E-03 | 4.22E-02 |
| SCN5A | 32.022 | 0.553 | 0.137 | 4.034 | 5.48E-05 | 1.48E-03 |
| PTGDR | 82.784 | 0.550 | 0.179 | 3.074 | 2.11E-03 | 1.95E-02 |
| IL17C | 14.426 | 0.550 | 0.166 | 3.320 | 9.02E-04 | 1.09E-02 |
| ADAD2 | 1.323 | 0.550 | 0.187 | 2.948 | 3.20E-03 | 2.59E-02 |
| GPR25 | 15.888 | 0.550 | 0.156 | 3.522 | 4.28E-04 | 6.56E-03 |
| RHOF | 91.707 | 0.550 | 0.114 | 4.804 | 1.55E-06 | 9.88E-05 |
| PXMP2 | 131.139 | 0.549 | 0.101 | 5.426 | 5.76E-08 | 7.45E-06 |
| CCDC189 | 18.745 | 0.548 | 0.097 | 5.682 | 1.33E-08 | 2.66E-06 |
| SH2D6 | 28.101 | 0.548 | 0.168 | 3.255 | 1.13E-03 | 1.28E-02 |
| FGFRL1 | 3328.500 | 0.547 | 0.096 | 5.671 | 1.42E-08 | 2.77E-06 |
| KCNG2 | 7.477 | 0.546 | 0.130 | 4.187 | 2.82E-05 | 9.24E-04 |
| CCDC187 | 36.324 | 0.545 | 0.147 | 3.707 | 2.10E-04 | 3.99E-03 |
| WDR38 | 2.010 | 0.545 | 0.185 | 2.951 | 3.17E-03 | 2.58E-02 |
| ROS1 | 17.997 | 0.545 | 0.179 | 3.050 | 2.29E-03 | 2.07E-02 |
| HGFAC | 4.241 | 0.543 | 0.178 | 3.053 | 2.27E-03 | 2.05E-02 |
| FOXE3 | 10.189 | 0.542 | 0.148 | 3.674 | 2.38E-04 | 4.38E-03 |
| TTC36 | 3.106 | 0.541 | 0.164 | 3.306 | 9.48E-04 | 1.13E-02 |
| CATIP | 2.236 | 0.540 | 0.148 | 3.657 | 2.55E-04 | 4.61E-03 |
| ALAS2 | 5.435 | 0.539 | 0.141 | 3.816 | 1.36E-04 | 2.90E-03 |
| ADAT3 | 228.327 | 0.535 | 0.112 | 4.794 | 1.63E-06 | 1.03E-04 |
| SERPINA4 | 48.543 | 0.535 | 0.176 | 3.034 | 2.41E-03 | 2.14E-02 |
| MGAT5B | 8.064 | 0.534 | 0.129 | 4.145 | 3.40E-05 | 1.05E-03 |
| CPZ | 11.987 | 0.533 | 0.123 | 4.331 | 1.49E-05 | 5.67E-04 |
| C16orf46 | 0.993 | 0.532 | 0.187 | 2.841 | 4.50E-03 | 3.25E-02 |
| KLK4 | 26.127 | 0.530 | 0.172 | 3.077 | 2.09E-03 | 1.94E-02 |
| FLRT1 | 34.732 | 0.524 | 0.109 | 4.823 | 1.42E-06 | 9.17E-05 |
| PCDHB8 | 19.970 | 0.524 | 0.183 | 2.859 | 4.25E-03 | 3.13E-02 |
| SUPT4H1 | 11.499 | 0.523 | 0.121 | 4.312 | 1.61E-05 | 6.02E-04 |
| ANTKMT | 713.578 | 0.521 | 0.094 | 5.561 | 2.68E-08 | 4.37E-06 |
| APOC2 | 1.098 | 0.520 | 0.190 | 2.742 | 6.11E-03 | 4.03E-02 |
| TMEM262 | 3.296 | 0.520 | 0.106 | 4.895 | 9.82E-07 | 7.01E-05 |
| MAZ | 1082.758 | 0.519 | 0.086 | 6.056 | 1.40E-09 | 4.67E-07 |
| SMIM22 | 1179.272 | 0.519 | 0.093 | 5.597 | 2.18E-08 | 3.81E-06 |
| SGSM3 | 39.346 | 0.519 | 0.097 | 5.360 | 8.32E-08 | 9.69E-06 |
| GRIN2C | 41.134 | 0.518 | 0.124 | 4.195 | 2.73E-05 | 8.99E-04 |
| ASB4 | 20.997 | 0.518 | 0.192 | 2.700 | 6.93E-03 | 4.38E-02 |
| H2BC13 | 14.461 | 0.518 | 0.128 | 4.040 | 5.34E-05 | 1.45E-03 |
| LCN12 | 165.005 | 0.518 | 0.107 | 4.831 | 1.36E-06 | 8.85E-05 |
| HCN1 | 41.458 | 0.515 | 0.186 | 2.764 | 5.71E-03 | 3.86E-02 |
| HAS1 | 25.571 | 0.515 | 0.164 | 3.135 | 1.72E-03 | 1.70E-02 |
| CFAP410 | 232.578 | 0.515 | 0.081 | 6.340 | 2.30E-10 | 1.19E-07 |
| KRT86 | 20.083 | 0.514 | 0.121 | 4.238 | 2.26E-05 | 7.80E-04 |
| PSMD9 | 202.631 | 0.514 | 0.081 | 6.368 | 1.92E-10 | 1.09E-07 |
| CSMD1 | 3.909 | 0.513 | 0.187 | 2.747 | 6.01E-03 | 3.99E-02 |
| SDHA | 538.017 | 0.512 | 0.089 | 5.728 | 1.02E-08 | 2.14E-06 |
| KCNT1 | 36.950 | 0.512 | 0.163 | 3.133 | 1.73E-03 | 1.71E-02 |
| C7orf61 | 4.656 | 0.511 | 0.111 | 4.624 | 3.77E-06 | 2.01E-04 |
| CCN5 | 31.805 | 0.510 | 0.174 | 2.933 | 3.36E-03 | 2.68E-02 |
| SHISA8 | 2.638 | 0.509 | 0.181 | 2.816 | 4.86E-03 | 3.41E-02 |
| CFAP73 | 11.161 | 0.508 | 0.139 | 3.649 | 2.63E-04 | 4.72E-03 |
| FGF8 | 3.157 | 0.508 | 0.151 | 3.364 | 7.69E-04 | 9.82E-03 |
| ACMSD | 3.337 | 0.508 | 0.173 | 2.929 | 3.40E-03 | 2.70E-02 |
| CASTOR1 | 7.587 | 0.507 | 0.110 | 4.603 | 4.16E-06 | 2.17E-04 |
| ELL3 | 17.998 | 0.506 | 0.095 | 5.312 | 1.09E-07 | 1.21E-05 |
| DLX1 | 8.791 | 0.505 | 0.184 | 2.751 | 5.94E-03 | 3.96E-02 |
| RPS27AP5 | 25.082 | 0.505 | 0.119 | 4.252 | 2.12E-05 | 7.47E-04 |
| RTN4RL2 | 41.343 | 0.505 | 0.104 | 4.845 | 1.27E-06 | 8.52E-05 |
| HDHD2 | 25.078 | 0.505 | 0.075 | 6.743 | 1.55E-11 | 1.62E-08 |
| ZNF497 | 43.946 | 0.504 | 0.087 | 5.786 | 7.20E-09 | 1.68E-06 |
| CHKB-CPT1B | 5.930 | 0.503 | 0.123 | 4.075 | 4.59E-05 | 1.31E-03 |
| C1QL1 | 30.416 | 0.502 | 0.115 | 4.365 | 1.27E-05 | 5.01E-04 |
| ECEL1 | 32.220 | 0.501 | 0.166 | 3.023 | 2.50E-03 | 2.18E-02 |
| GCGR | 3.983 | 0.501 | 0.153 | 3.280 | 1.04E-03 | 1.20E-02 |
| CTU1 | 316.608 | 0.501 | 0.099 | 5.049 | 4.43E-07 | 3.59E-05 |
| HPD | 11.356 | 0.501 | 0.145 | 3.454 | 5.51E-04 | 7.82E-03 |
| DRD4 | 35.340 | 0.501 | 0.099 | 5.046 | 4.51E-07 | 3.64E-05 |
| NPIPB4 | 125.690 | 0.500 | 0.086 | 5.845 | 5.08E-09 | 1.28E-06 |
| PPP1R1A | 22.395 | 0.500 | 0.164 | 3.050 | 2.29E-03 | 2.07E-02 |
| PLAC1 | 20.985 | 0.500 | 0.162 | 3.082 | 2.05E-03 | 1.92E-02 |
| KISS1R | 11.983 | 0.500 | 0.164 | 3.056 | 2.24E-03 | 2.04E-02 |
| CTAGE8 | 4.212 | -0.501 | 0.185 | -2.710 | 6.73E-03 | 4.30E-02 |
| CXCL8 | 3350.929 | -0.502 | 0.144 | -3.474 | 5.12E-04 | 7.43E-03 |
| UBE2L5 | 3.790 | -0.502 | 0.092 | -5.447 | 5.14E-08 | 6.86E-06 |
| FRMPD1 | 23.046 | -0.505 | 0.157 | -3.205 | 1.35E-03 | 1.44E-02 |
| EIF3CL | 14.722 | -0.506 | 0.154 | -3.292 | 9.94E-04 | 1.17E-02 |
| RPL39 | 4485.341 | -0.516 | 0.098 | -5.264 | 1.41E-07 | 1.50E-05 |
| SERTM2 | 6.319 | -0.520 | 0.192 | -2.709 | 6.75E-03 | 4.31E-02 |
| FER1L6 | 154.899 | -0.523 | 0.183 | -2.853 | 4.33E-03 | 3.17E-02 |
| NRXN1 | 9.810 | -0.526 | 0.196 | -2.681 | 7.35E-03 | 4.56E-02 |
| S100A8 | 192.625 | -0.529 | 0.129 | -4.101 | 4.11E-05 | 1.22E-03 |
| H1-5 | 32.919 | -0.534 | 0.164 | -3.255 | 1.13E-03 | 1.28E-02 |
| TAT | 3.805 | -0.545 | 0.172 | -3.162 | 1.57E-03 | 1.60E-02 |
| FMN2 | 17.303 | -0.546 | 0.190 | -2.868 | 4.14E-03 | 3.07E-02 |
| ZNF257 | 14.610 | -0.547 | 0.162 | -3.381 | 7.23E-04 | 9.44E-03 |
| BMPER | 60.271 | -0.549 | 0.157 | -3.485 | 4.92E-04 | 7.26E-03 |
| PI15 | 104.353 | -0.552 | 0.145 | -3.814 | 1.37E-04 | 2.93E-03 |
| PRIMA1 | 26.847 | -0.554 | 0.173 | -3.196 | 1.39E-03 | 1.47E-02 |
| IL1RN | 678.577 | -0.555 | 0.117 | -4.751 | 2.02E-06 | 1.21E-04 |
| ZNF578 | 15.990 | -0.556 | 0.137 | -4.063 | 4.84E-05 | 1.35E-03 |
| KCTD16 | 27.879 | -0.559 | 0.175 | -3.186 | 1.44E-03 | 1.51E-02 |
| GPR22 | 3.031 | -0.564 | 0.184 | -3.073 | 2.12E-03 | 1.96E-02 |
| SCEL | 65.013 | -0.567 | 0.173 | -3.285 | 1.02E-03 | 1.19E-02 |
| CLEC1B | 1.235 | -0.567 | 0.202 | -2.808 | 4.98E-03 | 3.47E-02 |
| ZMAT4 | 1.459 | -0.567 | 0.199 | -2.845 | 4.44E-03 | 3.22E-02 |
| NPY1R | 50.734 | -0.576 | 0.156 | -3.683 | 2.30E-04 | 4.28E-03 |
| KCNJ16 | 5.632 | -0.581 | 0.191 | -3.038 | 2.38E-03 | 2.13E-02 |
| ANO3 | 3.231 | -0.586 | 0.203 | -2.889 | 3.87E-03 | 2.94E-02 |
| PTH2R | 3.359 | -0.586 | 0.222 | -2.639 | 8.33E-03 | 4.97E-02 |
| PCDHB3 | 15.988 | -0.588 | 0.114 | -5.167 | 2.37E-07 | 2.26E-05 |
| H4C3 | 19.325 | -0.597 | 0.191 | -3.127 | 1.77E-03 | 1.73E-02 |
| NEFL | 20.732 | -0.598 | 0.189 | -3.164 | 1.56E-03 | 1.59E-02 |
| CMA1 | 5.989 | -0.598 | 0.222 | -2.700 | 6.93E-03 | 4.38E-02 |
| SIK1 | 48.538 | -0.602 | 0.187 | -3.223 | 1.27E-03 | 1.38E-02 |
| TDRD1 | 5.226 | -0.610 | 0.144 | -4.236 | 2.28E-05 | 7.86E-04 |
| AQP9 | 172.931 | -0.614 | 0.146 | -4.216 | 2.48E-05 | 8.39E-04 |
| MMP8 | 17.009 | -0.618 | 0.189 | -3.264 | 1.10E-03 | 1.25E-02 |
| ERBB4 | 1.543 | -0.632 | 0.223 | -2.838 | 4.54E-03 | 3.27E-02 |
| SCN7A | 29.487 | -0.634 | 0.193 | -3.282 | 1.03E-03 | 1.20E-02 |
| L1TD1 | 1066.237 | -0.635 | 0.222 | -2.858 | 4.26E-03 | 3.14E-02 |
| TMEM132C | 25.104 | -0.643 | 0.197 | -3.265 | 1.09E-03 | 1.25E-02 |
| TBX4 | 17.523 | -0.663 | 0.189 | -3.500 | 4.66E-04 | 6.98E-03 |
| CLEC4D | 6.500 | -0.664 | 0.163 | -4.060 | 4.91E-05 | 1.37E-03 |
| FDCSP | 47.412 | -0.668 | 0.209 | -3.202 | 1.36E-03 | 1.45E-02 |
| KRT6A | 119.896 | -0.668 | 0.205 | -3.260 | 1.12E-03 | 1.26E-02 |
| NPY5R | 3.284 | -0.670 | 0.238 | -2.815 | 4.87E-03 | 3.42E-02 |
| SMPX | 26.491 | -0.682 | 0.216 | -3.165 | 1.55E-03 | 1.59E-02 |
| CAV3 | 0.751 | -0.706 | 0.220 | -3.217 | 1.30E-03 | 1.41E-02 |
| CPS1 | 277.292 | -0.722 | 0.191 | -3.770 | 1.64E-04 | 3.33E-03 |
| NTS | 54.329 | -0.722 | 0.239 | -3.024 | 2.49E-03 | 2.18E-02 |
| CLDN10 | 18.901 | -0.727 | 0.260 | -2.803 | 5.06E-03 | 3.51E-02 |
| AL049839.2 | 0.780 | -0.746 | 0.279 | -2.671 | 7.55E-03 | 4.65E-02 |
| RTP3 | 0.817 | -0.771 | 0.291 | -2.647 | 8.11E-03 | 4.89E-02 |
| CYP4A11 | 1.417 | -0.774 | 0.251 | -3.081 | 2.07E-03 | 1.93E-02 |
| NLRP4 | 1.968 | -0.782 | 0.292 | -2.674 | 7.49E-03 | 4.62E-02 |
| COL6A5 | 6.318 | -0.786 | 0.184 | -4.280 | 1.87E-05 | 6.75E-04 |
| LEP | 4.731 | -0.816 | 0.221 | -3.692 | 2.22E-04 | 4.16E-03 |
| MEI4 | 0.950 | -0.824 | 0.278 | -2.960 | 3.08E-03 | 2.52E-02 |
| PADI3 | 92.052 | -0.833 | 0.208 | -3.997 | 6.40E-05 | 1.67E-03 |
| H2AP | 1.160 | -0.873 | 0.291 | -3.003 | 2.67E-03 | 2.27E-02 |
| SERPINA9 | 1.822 | -0.879 | 0.278 | -3.161 | 1.57E-03 | 1.60E-02 |
| H3C14 | 2.423 | -0.895 | 0.254 | -3.523 | 4.26E-04 | 6.56E-03 |
| HSPB3 | 15.866 | -0.912 | 0.242 | -3.761 | 1.69E-04 | 3.40E-03 |
| AC010255.2 | 2.169 | -0.929 | 0.266 | -3.494 | 4.76E-04 | 7.09E-03 |
| UGT2B4 | 6.396 | -0.965 | 0.309 | -3.128 | 1.76E-03 | 1.73E-02 |
| WIF1 | 160.711 | -1.041 | 0.275 | -3.777 | 1.59E-04 | 3.25E-03 |
| SOX21 | 1.714 | -1.047 | 0.302 | -3.468 | 5.25E-04 | 7.54E-03 |
| PPBP | 266.893 | -1.064 | 0.222 | -4.787 | 1.69E-06 | 1.06E-04 |
| ORM1 | 23.745 | -1.077 | 0.251 | -4.294 | 1.76E-05 | 6.44E-04 |
| TTC29 | 8.471 | -1.086 | 0.358 | -3.029 | 2.45E-03 | 2.16E-02 |
| REG1B | 1135.733 | -1.107 | 0.256 | -4.324 | 1.54E-05 | 5.77E-04 |
| AMY2A | 0.482 | -1.109 | 0.363 | -3.052 | 2.27E-03 | 2.05E-02 |
| CLDN18 | 270.648 | -1.131 | 0.274 | -4.132 | 3.60E-05 | 1.10E-03 |
| ACSM2A | 0.811 | -1.396 | 0.363 | -3.845 | 1.20E-04 | 2.66E-03 |
| KRT24 | 1.251 | -1.548 | 0.462 | -3.350 | 8.09E-04 | 1.02E-02 |
| SLC26A9 | 34.546 | -1.596 | 0.230 | -6.947 | 3.73E-12 | 5.36E-09 |
| H2BC3 | 2.197 | -1.609 | 0.405 | -3.974 | 7.07E-05 | 1.79E-03 |

**Supplementary Table 6.** GO enrichment analysis of DEGs between the high- and low-risk groups in the TCGA-COADREAD cohort.

| ONTOLOGY | ID | Description | GeneRatio | BgRatio | p-value | p-adjust | q-value | geneID | Count |
|---|---|---|---|---|---|---|---|---|---|
| BP | GO:0010901 | regulation of very-low-density lipoprotein particle remodeling | 5/410 | 6/18614 | 2.98E-08 | 1.28E-04 | 1.14E-04 | APOA1/APOA2/APOC3/APOA5/APOC2 | 5 |
| BP | GO:0006570 | tyrosine metabolic process | 6/410 | 17/18614 | 1.11E-06 | 2.38E-03 | 2.13E-03 | DCT/TH/TYR/HPD/TTC36/TAT | 6 |
| BP | GO:0034370 | triglyceride-rich lipoprotein particle remodeling | 5/410 | 12/18614 | 3.53E-06 | 3.78E-03 | 3.38E-03 | APOA1/APOA2/APOC3/APOA5/APOC2 | 5 |
| BP | GO:0034372 | very-low-density lipoprotein particle remodeling | 5/410 | 12/18614 | 3.53E-06 | 3.78E-03 | 3.38E-03 | APOA1/APOA2/APOC3/APOA5/APOC2 | 5 |
| BP | GO:0009072 | aromatic amino acid metabolic process | 7/410 | 32/18614 | 5.00E-06 | 4.29E-03 | 3.84E-03 | DCT/TH/TYR/HPD/TTC36/TAT/ACMSD | 7 |
| BP | GO:0033700 | phospholipid efflux | 5/410 | 14/18614 | 8.60E-06 | 6.14E-03 | 5.50E-03 | APOA1/APOA2/APOC3/APOA5/APOC2 | 5 |
| BP | GO:0007631 | feeding behavior | 11/410 | 109/18614 | 2.87E-05 | 1.45E-02 | 1.29E-02 | MC1R/MMP17/TH/OXT/GALR3/LEP/NPY1R/PPY/CNTFR/NMUR2/NPY5R | 11 |
| BP | GO:0010903 | negative regulation of very-low-density lipoprotein particle remodeling | 3/410 | 4/18614 | 4.17E-05 | 1.45E-02 | 1.29E-02 | APOA1/APOA2/APOC3 | 3 |
| BP | GO:0033686 | positive regulation of luteinizing hormone secretion | 3/410 | 4/18614 | 4.17E-05 | 1.45E-02 | 1.29E-02 | KISS1/LEP/FOXL2 | 3 |
| BP | GO:0034368 | protein-lipid complex remodeling | 6/410 | 30/18614 | 4.18E-05 | 1.45E-02 | 1.29E-02 | APOA1/APOA2/APOB/APOC3/APOA5/APOC2 | 6 |
| BP | GO:0034369 | plasma lipoprotein particle remodeling | 6/410 | 30/18614 | 4.18E-05 | 1.45E-02 | 1.29E-02 | APOA1/APOA2/APOB/APOC3/APOA5/APOC2 | 6 |
| BP | GO:0042744 | hydrogen peroxide catabolic process | 6/410 | 30/18614 | 4.18E-05 | 1.45E-02 | 1.29E-02 | HBZ/HBQ1/HBA1/HP/MT3/HBG2 | 6 |
| BP | GO:0032275 | luteinizing hormone secretion | 4/410 | 10/18614 | 4.38E-05 | 1.45E-02 | 1.29E-02 | KISS1/CGA/LEP/FOXL2 | 4 |
| BP | GO:0071827 | plasma lipoprotein particle organization | 7/410 | 46/18614 | 6.09E-05 | 1.75E-02 | 1.57E-02 | APOA1/APOA2/APOB/APOC3/SOAT2/APOA5/APOC2 | 7 |
| BP | GO:0034367 | protein-containing complex remodeling | 6/410 | 32/18614 | 6.14E-05 | 1.75E-02 | 1.57E-02 | APOA1/APOA2/APOB/APOC3/APOA5/APOC2 | 6 |
| BP | GO:0071825 | protein-lipid complex subunit organization | 7/410 | 49/18614 | 9.24E-05 | 2.25E-02 | 2.02E-02 | APOA1/APOA2/APOB/APOC3/SOAT2/APOA5/APOC2 | 7 |
| BP | GO:0015669 | gas transport | 5/410 | 22/18614 | 9.78E-05 | 2.25E-02 | 2.02E-02 | RHAG/HBZ/HBQ1/HBA1/HBG2 | 5 |
| BP | GO:0009636 | response to toxic substance | 16/410 | 242/18614 | 9.82E-05 | 2.25E-02 | 2.02E-02 | ALB/ABCB6/LTC4S/HBZ/HBQ1/SLC23A1/NUPR1/HBA1/HP/MT3/TH/GJC2/HBG2/CPS1/SRXN1/NEFL | 16 |
| BP | GO:0009111 | vitamin catabolic process | 4/410 | 12/18614 | 9.98E-05 | 2.25E-02 | 2.02E-02 | MTHFS/CYP2W1/PDXP/CYP26A1 | 4 |
| BP | GO:0022602 | ovulation cycle process | 7/410 | 52/18614 | 1.36E-04 | 2.80E-02 | 2.50E-02 | KISS1/AMH/AFP/CGA/LEP/NR5A1/NPY5R | 7 |
| BP | GO:0046461 | neutral lipid catabolic process | 6/410 | 37/18614 | 1.44E-04 | 2.80E-02 | 2.50E-02 | APOA2/APOB/APOC3/CPS1/APOA5/APOC | 6 |
| BP | GO:0046464 | acylglycerol catabolic process | 6/410 | 37/18614 | 1.44E-04 | 2.80E-02 | 2.50E-02 | APOA2/APOB/APOC3/CPS1/APOA5/APOC | 6 |
| BP | GO:0033344 | cholesterol efflux | 8/410 | 71/18614 | 1.64E-04 | 3.05E-02 | 2.73E-02 | GPS2/APOA1/APOA2/APOB/APOC3/SOAT2/APOA5/APOC2 | 8 |
| BP | GO:0015670 | carbon dioxide transport | 4/410 | 14/18614 | 1.95E-04 | 3.33E-02 | 2.98E-02 | RHAG/HBZ/HBA1/HBG2 | 4 |
| BP | GO:0042743 | hydrogen peroxide metabolic process | 7/410 | 55/18614 | 1.95E-04 | 3.33E-02 | 2.98E-02 | HBZ/HBQ1/HBA1/HP/MT3/HBG2/NOXO1 | 7 |
| BP | GO:0033684 | regulation of luteinizing hormone secretion | 3/410 | 6/18614 | 2.02E-04 | 3.33E-02 | 2.98E-02 | KISS1/LEP/FOXL2 | 3 |
| BP | GO:0034377 | plasma lipoprotein particle assembly | 5/410 | 26/18614 | 2.27E-04 | 3.52E-02 | 3.15E-02 | APOA1/APOA2/APOB/APOC3/SOAT2 | 5 |
| BP | GO:1990748 | cellular detoxification | 10/410 | 115/18614 | 2.30E-04 | 3.52E-02 | 3.15E-02 | ALB/ABCB6/LTC4S/HBZ/HBQ1/HBA1/HP/MT3/HBG2/SRXN1 | 10 |
| BP | GO:0019433 | triglyceride catabolic process | 5/410 | 27/18614 | 2.74E-04 | 4.05E-02 | 3.62E-02 | APOB/APOC3/CPS1/APOA5/APOC2 | 5 |
| BP | GO:0042632 | cholesterol homeostasis | 9/410 | 99/18614 | 3.37E-04 | 4.26E-02 | 3.81E-02 | PLSCR3/APOA1/APOA2/APOB/APOC3/SOAT2/APOA5/CAV3/APOC2 | 9 |
| BP | GO:0098869 | cellular oxidant detoxification | 9/410 | 99/18614 | 3.37E-04 | 4.26E-02 | 3.81E-02 | ALB/LTC4S/HBZ/HBQ1/HBA1/HP/MT3/HBG2/SRXN1 | 9 |
| BP | GO:0019748 | secondary metabolic process | 7/410 | 60/18614 | 3.38E-04 | 4.26E-02 | 3.81E-02 | PMEL/MC1R/DCT/CYP2W1/TH/TYR/ACM | 7 |
| BP | GO:0015671 | oxygen transport | 4/410 | 16/18614 | 3.42E-04 | 4.26E-02 | 3.81E-02 | HBZ/HBQ1/HBA1/HBG2 | 4 |
| BP | GO:0061101 | neuroendocrine cell differentiation | 4/410 | 16/18614 | 3.42E-04 | 4.26E-02 | 3.81E-02 | OTP/POU3F2/FGF8/LHX3 | 4 |
| BP | GO:0032278 | positive regulation of gonadotropin secretion | 3/410 | 7/18614 | 3.48E-04 | 4.26E-02 | 3.81E-02 | KISS1/LEP/FOXL2 | 3 |
| BP | GO:0055092 | sterol homeostasis | 9/410 | 100/18614 | 3.63E-04 | 4.29E-02 | 3.84E-02 | PLSCR3/APOA1/APOA2/APOB/APOC3/SOAT2/APOA5/CAV3/APOC2 | 9 |
| BP | GO:0046660 | female sex differentiation | 10/410 | 122/18614 | 3.71E-04 | 4.29E-02 | 3.84E-02 | NUPR1/AMH/AFP/CGA/LEP/LRP2/LHX1/NR5A1/FOXL2/CSMD1 | 10 |
| BP | GO:0065005 | protein-lipid complex assembly | 5/410 | 29/18614 | 3.88E-04 | 4.38E-02 | 3.92E-02 | APOA1/APOA2/APOB/APOC3/SOAT2 | 5 |
| BP | GO:0097237 | cellular response to toxic substance | 10/410 | 124/18614 | 4.22E-04 | 4.41E-02 | 3.95E-02 | ALB/ABCB6/LTC4S/HBZ/HBQ1/HBA1/HP/MT3/HBG2/SRXN1 | 10 |
| BP | GO:0032274 | gonadotropin secretion | 4/410 | 17/18614 | 4.40E-04 | 4.41E-02 | 3.95E-02 | KISS1/CGA/LEP/FOXL2 | 4 |
| BP | GO:0060192 | negative regulation of lipase activity | 4/410 | 17/18614 | 4.40E-04 | 4.41E-02 | 3.95E-02 | APOA2/ANXA8L1/APOC3/ANXA8 | 4 |
| BP | GO:0010902 | positive regulation of very-low-density lipoprotein particle remodeling | 2/410 | 2/18614 | 4.84E-04 | 4.41E-02 | 3.95E-02 | APOA5/APOC2 | 2 |
| BP | GO:0021985 | neurohypophysis development | 2/410 | 2/18614 | 4.84E-04 | 4.41E-02 | 3.95E-02 | OTP/POU3F2 | 2 |
| BP | GO:0042365 | water-soluble vitamin catabolic process | 2/410 | 2/18614 | 4.84E-04 | 4.41E-02 | 3.95E-02 | MTHFS/PDXP | 2 |
| BP | GO:0060112 | generation of ovulation cycle rhythm | 2/410 | 2/18614 | 4.84E-04 | 4.41E-02 | 3.95E-02 | KISS1/NPY5R | 2 |
| BP | GO:0060621 | negative regulation of cholesterol import | 2/410 | 2/18614 | 4.84E-04 | 4.41E-02 | 3.95E-02 | APOA2/APOC3 | 2 |
| BP | GO:2000296 | negative regulation of hydrogen peroxide catabolic | 2/410 | 2/18614 | 4.84E-04 | 4.41E-02 | 3.95E-02 | HP/MT3 | 2 |
| BP | GO:0061351 | neural precursor cell proliferation | 11/410 | 150/18614 | 4.99E-04 | 4.46E-02 | 3.99E-02 | DCT/ELL3/VAX1/OTP/POU3F2/GJC2/LRP2/SKOR2/FGF8/GBX2/LHX1 | 11 |
| BP | GO:0046884 | follicle-stimulating hormone secretion | 3/410 | 8/18614 | 5.47E-04 | 4.67E-02 | 4.18E-02 | CGA/LEP/FOXL2 | 3 |
| BP | GO:0030299 | intestinal cholesterol absorption | 4/410 | 18/18614 | 5.56E-04 | 4.67E-02 | 4.18E-02 | APOA1/APOA2/LEP/SOAT2 | 4 |
| BP | GO:0034384 | high-density lipoprotein particle clearance | 4/410 | 18/18614 | 5.56E-04 | 4.67E-02 | 4.18E-02 | APOA1/APOA2/APOC3/APOC2 | 4 |
| BP | GO:0021953 | central nervous system neuron differentiation | 12/410 | 177/18614 | 5.72E-04 | 4.72E-02 | 4.22E-02 | C12orf57/OTP/LEP/NKX6-1/SKOR2/GBX2/LHX1/TTC36/LHX3/DLX1/FEZF2/FOXN4 | 12 |
| BP | GO:0032309 | icosanoid secretion | 6/410 | 48/18614 | 6.18E-04 | 5.00E-02 | 4.47E-02 | DRD4/OXT/LEP/CYP4A11/MIF/NMUR2 | 6 |
| CC | GO:0031838 | haptoglobin-hemoglobin complex | 5/439 | 11/19518 | 2.32E-06 | 4.31E-04 | 3.89E-04 | HBZ/HBQ1/HBA1/HP/HBG2 | 5 |
| CC | GO:0034361 | very-low-density lipoprotein particle | 6/439 | 20/19518 | 3.71E-06 | 4.31E-04 | 3.89E-04 | APOA1/APOA2/APOB/APOC3/APOA5/APOC2 | 6 |
| CC | GO:0034385 | triglyceride-rich plasma lipoprotein particle | 6/439 | 20/19518 | 3.71E-06 | 4.31E-04 | 3.89E-04 | APOA1/APOA2/APOB/APOC3/APOA5/APOC2 | 6 |
| CC | GO:0042627 | chylomicron | 5/439 | 12/19518 | 3.91E-06 | 4.31E-04 | 3.89E-04 | APOA2/APOB/APOC3/APOA5/APOC2 | 5 |
| CC | GO:0072562 | blood microparticle | 14/439 | 144/19518 | 4.75E-06 | 4.31E-04 | 3.89E-04 | AHSG/ALB/TF/ITIH2/APOA1/APOA2/F2/HBA1/C8G/HP/ORM1/CPN2/HBG2/FGG | 14 |
| CC | GO:0034366 | spherical high-density lipoprotein particle | 4/439 | 8/19518 | 1.65E-05 | 1.24E-03 | 1.12E-03 | APOA1/APOA2/APOC3/APOC2 | 4 |
| CC | GO:0034364 | high-density lipoprotein particle | 6/439 | 27/19518 | 2.48E-05 | 1.61E-03 | 1.45E-03 | APOA1/APOA2/APOB/APOC3/APOA5/APOC2 | 6 |
| CC | GO:0034363 | intermediate-density lipoprotein particle | 3/439 | 4/19518 | 4.45E-05 | 2.52E-03 | 2.28E-03 | APOB/APOC3/APOC2 | 3 |
| CC | GO:0033162 | melanosome membrane | 5/439 | 21/19518 | 8.50E-05 | 3.51E-03 | 3.16E-03 | PMEL/ABCB6/DCT/TH/TYR | 5 |
| CC | GO:0045009 | chitosome | 5/439 | 21/19518 | 8.50E-05 | 3.51E-03 | 3.16E-03 | PMEL/ABCB6/DCT/TH/TYR | 5 |
| CC | GO:0090741 | pigment granule membrane | 5/439 | 21/19518 | 8.50E-05 | 3.51E-03 | 3.16E-03 | PMEL/ABCB6/DCT/TH/TYR | 5 |
| CC | GO:0005833 | hemoglobin complex | 4/439 | 12/19518 | 1.08E-04 | 4.10E-03 | 3.69E-03 | HBZ/HBQ1/HBA1/HBG2 | 4 |
| CC | GO:0034358 | plasma lipoprotein particle | 6/439 | 36/19518 | 1.38E-04 | 4.46E-03 | 4.02E-03 | APOA1/APOA2/APOB/APOC3/APOA5/APOC2 | 6 |
| CC | GO:1990777 | lipoprotein particle | 6/439 | 36/19518 | 1.38E-04 | 4.46E-03 | 4.02E-03 | APOA1/APOA2/APOB/APOC3/APOA5/APOC2 | 6 |
| CC | GO:0032994 | protein-lipid complex | 6/439 | 39/19518 | 2.18E-04 | 6.59E-03 | 5.94E-03 | APOA1/APOA2/APOB/APOC3/APOA5/APOC2 | 6 |
| CC | GO:0031983 | vesicle lumen | 18/439 | 327/19518 | 4.71E-04 | 1.34E-02 | 1.20E-02 | AHSG/ALB/TF/TXNDC5/ALDOA/APOA1/HP/NME2/PPBP/ORM1/APOB/S100A8/IGF2/MMP8/TTR/FGG/SERPINA4/MIF | 18 |
| CC | GO:0031093 | platelet alpha granule lumen | 7/439 | 67/19518 | 7.55E-04 | 2.02E-02 | 1.82E-02 | AHSG/ALB/ALDOA/PPBP/ORM1/IGF2/FGG | 7 |
| CC | GO:0034774 | secretory granule lumen | 17/439 | 322/19518 | 1.06E-03 | 2.68E-02 | 2.42E-02 | AHSG/ALB/TF/TXNDC5/ALDOA/APOA1/HP/NME2/PPBP/ORM1/S100A8/IGF2/MMP8/TTR/FGG/SERPINA4/MIF | 17 |
| CC | GO:0060205 | cytoplasmic vesicle lumen | 17/439 | 325/19518 | 1.17E-03 | 2.81E-02 | 2.53E-02 | AHSG/ALB/TF/TXNDC5/ALDOA/APOA1/HP/NME2/PPBP/ORM1/S100A8/IGF2/MMP8/TTR/FGG/SERPINA4/MIF | 17 |
| CC | GO:1904724 | tertiary granule lumen | 6/439 | 55/19518 | 1.43E-03 | 3.22E-02 | 2.91E-02 | ALDOA/FTH1/HP/PPBP/ORM1/MMP8 | 6 |
| CC | GO:0033557 | Slx1-Slx4 complex | 2/439 | 3/19518 | 1.49E-03 | 3.22E-02 | 2.91E-02 | SLX1A/SLX1B | 2 |
| CC | GO:0071682 | endocytic vesicle lumen | 4/439 | 23/19518 | 1.59E-03 | 3.29E-02 | 2.96E-02 | APOA1/HBA1/HP/APOB | 4 |
| CC | GO:0000800 | lateral element | 3/439 | 12/19518 | 2.14E-03 | 4.22E-02 | 3.80E-02 | SMC1B/KASH5/MEI4 | 3 |
| MF | GO:0043177 | organic acid binding | 18/410 | 212/18369 | 1.38E-06 | 1.16E-03 | 9.17E-04 | ALB/GCHFR/MTHFS/FTCD/EGLN1/HBZ/HBQ1/CYP2W1/HBA1/LCN12/CASTOR1/TH/GRIN3B/S100A8/CYP26A1/HBG2/CPS1/TA | 18 |
| MF | GO:0004252 | serine-type endopeptidase activity | 15/410 | 170/18369 | 6.56E-06 | 2.74E-03 | 2.17E-03 | RHBDL1/TMPRSS11E/F2/HP/PRSS56/KLK5/HPN/F11/PRSS33/F7/PRSS21/MMP8/KLK4/HGFAC/CMA1 | 15 |
| MF | GO:0070653 | high-density lipoprotein particle receptor binding | 3/410 | 3/18369 | 1.10E-05 | 3.08E-03 | 2.44E-03 | APOA1/APOA2/APOC3 | 3 |
| MF | GO:0019825 | oxygen binding | 7/410 | 39/18369 | 2.18E-05 | 4.16E-03 | 3.30E-03 | ALB/HBZ/HBQ1/HBA1/TH/CYP26A1/HBG2 | 7 |
| MF | GO:0008236 | serine-type peptidase activity | 15/410 | 190/18369 | 2.49E-05 | 4.16E-03 | 3.30E-03 | RHBDL1/TMPRSS11E/F2/HP/PRSS56/KLK5/HPN/F11/PRSS33/F7/PRSS21/MMP8/KLK4/HGFAC/CMA1 | 15 |
| MF | GO:0017171 | serine hydrolase activity | 15/410 | 194/18369 | 3.17E-05 | 4.42E-03 | 3.51E-03 | RHBDL1/TMPRSS11E/F2/HP/PRSS56/KLK5/HPN/F11/PRSS33/F7/PRSS21/MMP8/KLK4/HGFAC/CMA1 | 15 |
| MF | GO:0031720 | haptoglobin binding | 4/410 | 10/18369 | 4.62E-05 | 5.51E-03 | 4.37E-03 | HBZ/HBQ1/HBA1/HBG2 | 4 |
| MF | GO:0016209 | antioxidant activity | 9/410 | 82/18369 | 8.75E-05 | 9.15E-03 | 7.26E-03 | ALB/LTC4S/HBZ/HBQ1/HBA1/HP/MT3/HBG2/SRXN1 | 9 |
| MF | GO:0004175 | endopeptidase activity | 23/410 | 428/18369 | 1.01E-04 | 9.42E-03 | 7.47E-03 | PSMD9/RHBDL1/PSMA6/TMPRSS11E/MMP17/F2/HP/PRSS56/KLK5/TAX1BP3/MMP23B/CPS1/HPN/F11/PRSS33/F7/PRSS21/MMP8/ADAMTS19/KLK4/HGFAC/ECEL1/CMA1 | 23 |
| MF | GO:0031406 | carboxylic acid binding | 14/410 | 200/18369 | 1.67E-04 | 1.25E-02 | 9.95E-03 | ALB/GCHFR/MTHFS/FTCD/EGLN1/CYP2W1/LCN12/CASTOR1/TH/GRIN3B/S100A8/CYP26A1/CPS1/TAT | 14 |
| MF | GO:0015267 | channel activity | 24/410 | 478/18369 | 2.02E-04 | 1.25E-02 | 9.95E-03 | SLC26A9/NOX5/RHAG/TRPM5/LRRC26/GRIN3B/CALHM6/AQP9/GRIN2C/KCNG2/GJC2/SCN5A/TRPM1/RHBG/KCNQ2/SCN7A/GABRR1/KCNJ4/KCNT1/KCNJ16/NMUR2/ANO3/HCN1/OTOP1 | 24 |
| MF | GO:0005344 | oxygen carrier activity | 4/410 | 14/18369 | 2.05E-04 | 1.25E-02 | 9.95E-03 | HBZ/HBQ1/HBA1/HBG2 | 4 |
| MF | GO:0022803 | passive transmembrane transporter activity | 24/410 | 479/18369 | 2.08E-04 | 1.25E-02 | 9.95E-03 | SLC26A9/NOX5/RHAG/TRPM5/LRRC26/GRIN3B/CALHM6/AQP9/GRIN2C/KCNG2/GJC2/SCN5A/TRPM1/RHBG/KCNQ2/SCN7A/GABRR1/KCNJ4/KCNT1/KCNJ16/NMUR2/ANO3/HCN1/OTOP1 | 24 |
| MF | GO:0060228 | phosphatidylcholine-sterol O-acyltransferase activator activity | 3/410 | 6/18369 | 2.10E-04 | 1.25E-02 | 9.95E-03 | APOA1/APOA2/APOA5 | 3 |
| MF | GO:0070325 | lipoprotein particle receptor binding | 5/410 | 29/18369 | 4.13E-04 | 2.30E-02 | 1.82E-02 | APOA1/APOA2/APOB/APOC3/APOA5 | 5 |
| MF | GO:0030506 | ankyrin binding | 4/410 | 19/18369 | 7.27E-04 | 3.80E-02 | 3.01E-02 | RHAG/SCN5A/RHBG/KCNQ2 | 4 |
| MF | GO:0140104 | molecular carrier activity | 8/410 | 88/18369 | 7.79E-04 | 3.83E-02 | 3.04E-02 | IPO4/TF/SCO2/EMC6/HBZ/HBQ1/HBA1/HBG2 | 8 |
| MF | GO:0048018 | receptor ligand activity | 23/410 | 497/18369 | 8.41E-04 | 3.91E-02 | 3.10E-02 | MIA/CDC42EP2/APOA1/F2/PPBP/IL1RN/ERFE/AMH/OXT/IGF2/CGA/LEP/CXCL8/FGF8/IL17C/PPY/TTR/PROK1/NTS/IFNE/MIF/IFNL1/IL36RN | 23 |
| MF | GO:0005244 | voltage-gated monoatomic ion channel activity | 12/410 | 189/18369 | 1.14E-03 | 4.74E-02 | 3.76E-02 | TRPM5/LRRC26/GRIN3B/GRIN2C/KCNG2/SCN5A/KCNQ2/SCN7A/KCNJ4/KCNT1/KCNJ16/HCN1 | 12 |
| MF | GO:0022832 | voltage-gated channel activity | 12/410 | 189/18369 | 1.14E-03 | 4.74E-02 | 3.76E-02 | TRPM5/LRRC26/GRIN3B/GRIN2C/KCNG2/SCN5A/KCNQ2/SCN7A/KCNJ4/KCNT1/KCNJ16/HCN1 | 12 |
| MF | GO:0016597 | amino acid binding | 6/410 | 55/18369 | 1.37E-03 | 4.74E-02 | 3.76E-02 | GCHFR/CASTOR1/TH/GRIN3B/CPS1/TAT | 6 |
| MF | GO:0001601 | peptide YY receptor activity | 2/410 | 3/18369 | 1.47E-03 | 4.74E-02 | 3.76E-02 | NPY1R/NPY5R | 2 |
| MF | GO:0001602 | pancreatic polypeptide receptor activity | 2/410 | 3/18369 | 1.47E-03 | 4.74E-02 | 3.76E-02 | NPY1R/NPY5R | 2 |
| MF | GO:0004167 | dopachrome isomerase activity | 2/410 | 3/18369 | 1.47E-03 | 4.74E-02 | 3.76E-02 | DCT/MIF | 2 |
| MF | GO:0005152 | interleukin-1 receptor antagonist activity | 2/410 | 3/18369 | 1.47E-03 | 4.74E-02 | 3.76E-02 | IL1RN/IL36RN | 2 |
| MF | GO:0030527 | structural constituent of chromatin | 8/410 | 97/18369 | 1.47E-03 | 4.74E-02 | 3.76E-02 | H2AC16/H2BC13/H2BC3/H3C14/H1-5/H2AC17/H4C3/H4C1 | 8 |
| MF | GO:0008199 | ferric iron binding | 3/410 | 11/18369 | 1.59E-03 | 4.94E-02 | 3.91E-02 | TF/FTH1/TH | 3 |
| MF | GO:0005179 | hormone activity | 9/410 | 122/18369 | 1.67E-03 | 4.98E-02 | 3.95E-02 | ERFE/AMH/OXT/IGF2/CGA/LEP/PPY/TTR/NTS | 9 |

**Supplementary Table 7.** KEGG enrichment analysis of DEGs between the high- and low-risk groups in the TCGA-COADREAD cohort.

| ID | **Description** | **GeneRatio** | **BgRatio** | **p-value** | **p-adjust** | **q-value** | **geneID** | **Count** |
|---|---|---|---|---|---|---|---|---|
| hsa04080 | Neuroactive ligand-receptor interaction | 23/191 | 370/9434 | 1.52E-06 | 3.76E-04 | 3.46E-04 | 4157/3814/5731/2147/1815/116444/2905/5020/8484/1081/3952/4886/553 9/2642/2569/887/5729/84634/4922/56923/4889/11255/5746 | 23 |
| hsa04979 | Cholesterol metabolism | 7/191 | 51/9434 | 6.81E-05 | 6.18E-03 | 5.69E-03 | 335/336/338/345/8435/4036/344 | 7 |
| hsa00350 | Tyrosine metabolism | 6/191 | 36/9434 | 7.48E-05 | 6.18E-03 | 5.69E-03 | 1638/7054/7299/3242/6898/4282 | 6 |
| hsa05034 | Alcoholism | 13/191 | 191/9434 | 1.30E-04 | 8.08E-03 | 7.44E-03 | 2790/2787/7054/8332/116444/2905/8340/810/3018/126961/8336/8364/83 | 13 |
| hsa04978 | Mineral absorption | 6/191 | 61/9434 | 1.40E-03 | 6.74E-02 | 6.20E-02 | 115019/7018/2495/486/340024/645745 | 6 |
| hsa04081 | Hormone signaling | 12/191 | 219/9434 | 1.63E-03 | 6.74E-02 | 6.20E-02 | 4157/2790/2787/1815/268/5020/3481/1081/3952/114/2642/887 | 12 |
| hsa00360 | Phenylalanine metabolism | 3/191 | 16/9434 | 3.77E-03 | 1.25E-01 | 1.15E-01 | 3242/6898/4282 | 3 |
| hsa04918 | Thyroid hormone synthesis | 6/191 | 75/9434 | 4.03E-03 | 1.25E-01 | 1.15E-01 | 213/486/1081/4036/114/7276 | 6 |
| hsa00670 | One carbon pool by folate | 4/191 | 39/9434 | 7.70E-03 | 2.12E-01 | 1.95E-01 | 10588/10841/4143/635 | 4 |
| hsa05322 | Systemic lupus erythematosus | 8/191 | 144/9434 | 8.83E-03 | 2.19E-01 | 2.02E-01 | 733/8332/8340/3018/126961/8336/8364/8359 | 8 |
| hsa04720 | Long-term potentiation | 5/191 | 67/9434 | 1.13E-02 | 2.44E-01 | 2.25E-01 | 816/2905/810/114/5502 | 5 |
| hsa05031 | Amphetamine addiction | 5/191 | 69/9434 | 1.27E-02 | 2.44E-01 | 2.25E-01 | 816/7054/116444/2905/810 | 5 |
| hsa04713 | Circadian entrainment | 6/191 | 97/9434 | 1.37E-02 | 2.44E-01 | 2.25E-01 | 2790/2787/816/2905/810/114 | 6 |
| hsa04977 | Vitamin digestion and absorption | 3/191 | 26/9434 | 1.51E-02 | 2.44E-01 | 2.25E-01 | 335/9963/338 | 3 |
| hsa04916 | Melanogenesis | 6/191 | 101/9434 | 1.65E-02 | 2.44E-01 | 2.25E-01 | 4157/1638/816/7299/810/114 | 6 |
| hsa04024 | cAMP signaling pathway | 10/191 | 226/9434 | 1.65E-02 | 2.44E-01 | 2.25E-01 | 816/486/116444/268/2905/5020/810/1081/4886/114 | 10 |
| hsa01230 | Biosynthesis of amino acids | 5/191 | 75/9434 | 1.78E-02 | 2.44E-01 | 2.25E-01 | 95/226/8277/1373/4143 | 5 |
| hsa04613 | Neutrophil extracellular trap formation | 9/191 | 196/9434 | 1.81E-02 | 2.44E-01 | 2.25E-01 | 8332/366/8340/3018/126961/8336/8364/2266/8359 | 9 |
| hsa04971 | Gastric acid secretion | 5/191 | 76/9434 | 1.87E-02 | 2.44E-01 | 2.25E-01 | 816/810/114/887/3773 | 5 |
| hsa04082 | Neuroactive ligand signaling | 9/191 | 199/9434 | 1.98E-02 | 2.45E-01 | 2.26E-01 | 2790/2787/1815/116444/2905/6530/114/2569/11255 | 9 |
| hsa00130 | Ubiquinone and other terpenoid-quinone biosynthesis | 2/191 | 12/9434 | 2.36E-02 | 2.78E-01 | 2.56E-01 | 3242/6898 | 2 |
| hsa04725 | Cholinergic synapse | 6/191 | 116/9434 | 3.02E-02 | 3.39E-01 | 3.12E-01 | 2790/2787/816/114/3785/3761 | 6 |
| hsa04382 | Cornified envelope formation | 9/191 | 217/9434 | 3.22E-02 | 3.39E-01 | 3.12E-01 | 3892/51208/6279/51702/192666/3853/9071/121391/3890 | 9 |
| hsa04610 | Complement and coagulation cascades | 5/191 | 88/9434 | 3.28E-02 | 3.39E-01 | 3.12E-01 | 2147/733/2160/2155/2266 | 5 |
| hsa04727 | GABAergic synapse | 5/191 | 89/9434 | 3.42E-02 | 3.39E-01 | 3.12E-01 | 11337/2790/2787/114/2569 | 5 |
| hsa04261 | Adrenergic signaling in cardiomyocytes | 7/191 | 154/9434 | 3.68E-02 | 3.51E-01 | 3.24E-01 | 816/486/6331/810/114/6332/5502 | 7 |
| hsa04929 | GnRH secretion | 4/191 | 65/9434 | 4.24E-02 | 3.89E-01 | 3.58E-01 | 3814/1081/84634/348980 | 4 |
| hsa00910 | Nitrogen metabolism | 2/191 | 17/9434 | 4.54E-02 | 3.96E-01 | 3.65E-01 | 766/1373 | 2 |
| hsa04970 | Salivary secretion | 5/191 | 97/9434 | 4.67E-02 | 3.96E-01 | 3.65E-01 | 486/1470/810/114/279 | 5 |
| hsa05012 | Parkinson disease | 10/191 | 271/9434 | 4.87E-02 | 3.96E-01 | 3.65E-01 | 5715/374291/5687/51079/5413/6389/816/7054/810/112714 | 10 |
| hsa05033 | Nicotine addiction | 3/191 | 41/9434 | 4.97E-02 | 3.96E-01 | 3.65E-01 | 116444/2905/2569 | 3 |

**Supplementary Table 8.** Correlation between core deep learning-derived pathomics features and prognosis-related genes.

| Term | Group | Degree | Eigenvector Centrality |
|---|---|---|---|
| MRPL37 | Gene | 67 | 1.000 |
| NALCN | Gene | 55 | 0.977 |
| TNIP3 | Gene | 39 | 0.814 |
| CLDN23 | Gene | 38 | 0.781 |
| SYPL2 | Gene | 39 | 0.762 |
| MPP2 | Gene | 39 | 0.731 |
| 339 | Extractor Channal | 33 | 0.724 |
| NGLY1 | Gene | 39 | 0.710 |
| NSG1 | Gene | 28 | 0.590 |
| EPHB2 | Gene | 19 | 0.565 |
| 217 | Extractor Channal | 20 | 0.560 |
| 88 | Extractor Channal | 17 | 0.544 |
| 278 | Extractor Channal | 17 | 0.536 |
| 441 | Extractor Channal | 16 | 0.531 |
| ARL8A | Gene | 14 | 0.513 |
| OSR1 | Gene | 25 | 0.501 |
| 320 | Extractor Channal | 13 | 0.484 |
| DNASE1L3 | Gene | 22 | 0.465 |
| FLT1 | Gene | 12 | 0.448 |
| 380 | Extractor Channal | 11 | 0.445 |
| 356 | Extractor Channal | 14 | 0.439 |
| NRG1 | Gene | 11 | 0.434 |
| 502 | Extractor Channal | 10 | 0.432 |
| UBE2H | Gene | 10 | 0.432 |
| GNAT2 | Gene | 11 | 0.421 |
| WDR86 | Gene | 10 | 0.419 |
| EMP2 | Gene | 12 | 0.418 |
| TRAP1 | Gene | 10 | 0.418 |
| H2BE1 | Gene | 10 | 0.406 |
| ANKRD6 | Gene | 18 | 0.402 |
| 471 | Extractor Channal | 9 | 0.392 |
| MARK4 | Gene | 12 | 0.370 |
| CPT2 | Gene | 8 | 0.370 |
| 485 | Extractor Channal | 9 | 0.366 |
| SLC35D1 | Gene | 8 | 0.360 |
| ZG16 | Gene | 9 | 0.352 |
| COX11 | Gene | 8 | 0.321 |
| 271 | Extractor Channal | 10 | 0.315 |
| 95 | Extractor Channal | 7 | 0.311 |
| 490 | Extractor Channal | 7 | 0.309 |
| 366 | Extractor Channal | 7 | 0.302 |
| VAV2 | Gene | 13 | 0.298 |
| C1QTNF4 | Gene | 7 | 0.298 |
| NPIPB11 | Gene | 7 | 0.292 |
| TPP1 | Gene | 6 | 0.288 |
| ITLN1 | Gene | 9 | 0.288 |
| CTDSP1 | Gene | 6 | 0.270 |
| 317 | Extractor Channal | 8 | 0.266 |
| RIMKLB | Gene | 10 | 0.261 |
| C5orf46 | Gene | 10 | 0.254 |
| TIGD6 | Gene | 6 | 0.253 |
| MYH3 | Gene | 5 | 0.243 |
| MAPK9 | Gene | 5 | 0.241 |
| HEYL | Gene | 13 | 0.241 |
| ACOT11 | Gene | 10 | 0.239 |
| PHF1 | Gene | 5 | 0.238 |
| DNAJB2 | Gene | 5 | 0.234 |
| TSPYL2 | Gene | 5 | 0.219 |
| KCNMB3 | Gene | 6 | 0.208 |
| PRKRIP1 | Gene | 5 | 0.205 |
| UBQLNL | Gene | 4 | 0.200 |
| LARS2 | Gene | 4 | 0.200 |
| RASA4B | Gene | 4 | 0.179 |
| MYH2 | Gene | 5 | 0.171 |
| 19 | Extractor Channal | 5 | 0.171 |
| CTNNA1 | Gene | 4 | 0.162 |
| 438 | Extractor Channal | 4 | 0.160 |
| ZCWPW1 | Gene | 3 | 0.156 |
| ZNF98 | Gene | 3 | 0.143 |
| LEP | Gene | 3 | 0.128 |
| MAGED4B | Gene | 3 | 0.126 |
| RBP7 | Gene | 3 | 0.126 |
| SPINK4 | Gene | 3 | 0.120 |
| 486 | Extractor Channal | 3 | 0.116 |
| PCDH9 | Gene | 2 | 0.113 |
| MUCL3 | Gene | 2 | 0.113 |
| 222 | Extractor Channal | 5 | 0.099 |
| NLGN1 | Gene | 2 | 0.099 |
| SIX4 | Gene | 2 | 0.099 |
| KCNIP2 | Gene | 2 | 0.099 |
| SEMA4C | Gene | 2 | 0.099 |
| ASPG | Gene | 3 | 0.095 |
| MID2 | Gene | 2 | 0.085 |
| TCF7L1 | Gene | 2 | 0.078 |
| TRPA1 | Gene | 3 | 0.075 |
| GADD45B | Gene | 6 | 0.072 |
| FAM167B | Gene | 2 | 0.070 |
| 259 | Extractor Channal | 2 | 0.063 |
| RNF112 | Gene | 3 | 0.061 |
| INSC | Gene | 1 | 0.057 |
| ANXA8 | Gene | 1 | 0.057 |
| MRPS7 | Gene | 1 | 0.057 |
| ACADL | Gene | 1 | 0.057 |
| CLCA1 | Gene | 1 | 0.057 |
| MRPL58 | Gene | 1 | 0.057 |
| DBX2 | Gene | 1 | 0.057 |
| ELP6 | Gene | 1 | 0.057 |
| MRPL22 | Gene | 1 | 0.057 |
| TBX19 | Gene | 1 | 0.057 |
| IL13RA2 | Gene | 1 | 0.057 |
| NDUFAB1 | Gene | 1 | 0.056 |
| PTPN14 | Gene | 1 | 0.056 |
| UCN | Gene | 1 | 0.056 |
| GPSM2 | Gene | 5 | 0.051 |
| 129 | Extractor Channal | 4 | 0.049 |
| TMEM233 | Gene | 1 | 0.043 |
| ZNF676 | Gene | 1 | 0.043 |
| NGF | Gene | 1 | 0.042 |
| NUDT6 | Gene | 1 | 0.040 |
| GLP2R | Gene | 1 | 0.040 |
| PMM2 | Gene | 1 | 0.040 |
| SALL1 | Gene | 1 | 0.040 |
| CST11 | Gene | 1 | 0.040 |
| TPSG1 | Gene | 1 | 0.034 |
| FGF5 | Gene | 2 | 0.031 |
| RFPL4B | Gene | 1 | 0.029 |
| PRG4 | Gene | 1 | 0.029 |
| ACTR8 | Gene | 3 | 0.028 |
| CYP19A1 | Gene | 1 | 0.026 |
| MEF2B | Gene | 2 | 0.018 |
| HOMER3 | Gene | 1 | 0.014 |
| HLX | Gene | 1 | 0.014 |
| 203 | Extractor Channal | 1 | 0.010 |
| HTR2B | Gene | 1 | 0.004 |
| RYR2 | Gene | 1 | 0.003 |
| B3GNT6 | Gene | 1 | 0.003 |
| PPARGC1A | Gene | 1 | 0.003 |
| 117 | Extractor Channal | 3 | 0.003 |
| DNAI4 | Gene | 1 | 0.002 |
| 120 | Extractor Channal | 1 | 0.000 |
| 110 | Extractor Channal | 1 | 0.000 |